\documentclass[11pt]{article}

\usepackage[final]{acl}

\usepackage{times}
\usepackage{latexsym}
\usepackage{float}

\usepackage[T1]{fontenc}

\usepackage[utf8]{inputenc}

\usepackage{microtype}

\usepackage{inconsolata}

\usepackage{booktabs}
\usepackage{multirow}
\usepackage{arydshln}
\usepackage{pdflscape}
 
\usepackage{booktabs}
\usepackage{multirow}
\usepackage{arydshln}
\usepackage{adjustbox}
 
\usepackage{graphicx}
\usepackage{longtable}
\usepackage{hyperref}
\usepackage{url}
\usepackage{array}
\usepackage{booktabs}
\usepackage{arydshln}
\usepackage[table]{xcolor}
\usepackage{colortbl}
\usepackage{subcaption}
\usepackage{listings}
\usepackage[most]{tcolorbox}
\tcbuselibrary{breakable, skins}
\usepackage{booktabs}
\usepackage{multirow}
\definecolor{darklime}{RGB}{50,160,60} 
 
\definecolor{promptdark}{RGB}{74,99,73}
\definecolor{promptlight}{RGB}{243,247,239}
\newtcolorbox{cmechpromptbox}[1]{%
  enhanced,
  breakable,
  width=\linewidth,
  colback=promptlight,
  colframe=promptdark,
  colbacktitle=promptdark,
  coltitle=white,
  title={#1},
  fonttitle=\bfseries,
  right=8pt,
  top=8pt,
  bottom=8pt,
  toptitle=6pt,
  bottomtitle=6pt,
}
 
\newif\ifdraft
\drafttrue   
 
\definecolor{highlightblue}{RGB}{0,90,200}

\definecolor{correction}{RGB}{0,90,200}
\definecolor{previous}{RGB}{255,140,0}

\newcommand{\inlinepdfcrop}[3][1.35em]{%
  \raisebox{-.35\height}{\includegraphics[height=#1,page=1,trim=#2,clip]{#3}}%
}
\usepackage{amssymb}
 
\title{Evaluating Cross-lingual Knowledge Consistency in Code-Mixed vis-a-vis Indian Languages using IndicKLAR
\raisebox{-0.5ex}{\includegraphics[height=1.5em]{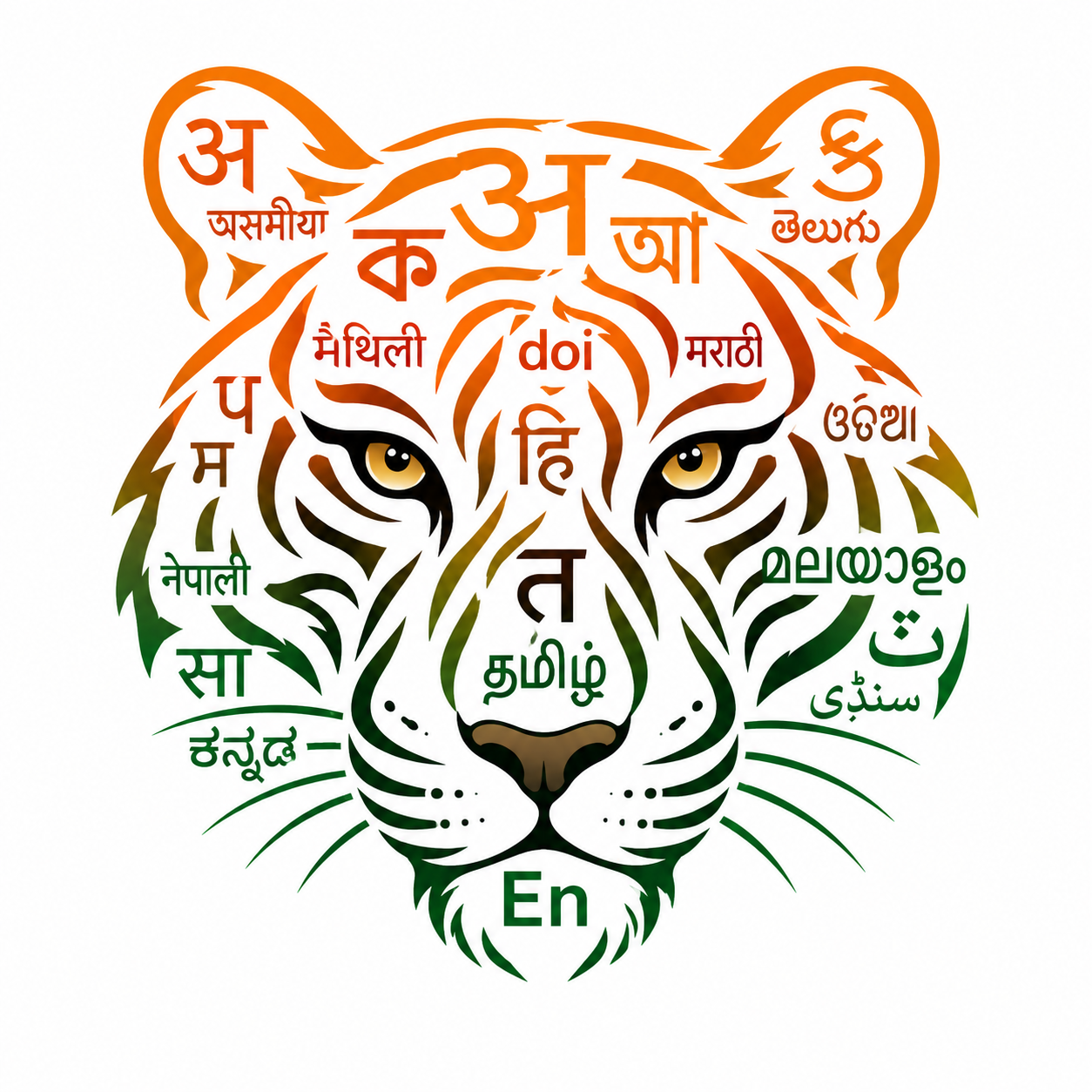}} }

\author{\normalfont
Debajyoti Mazumder$^1$, Divyansh Pathak$^1$, Prashant Kodali$^2$, \\
Aditya Joshi$^3$, Akshay Agarwal$^1$, Jasabanta Patro$^1$ \\ 
$^1$Indian Institute of Science Education and Research, Bhopal, India \\
$^2$Microsoft Research India \hspace{1cm} $^3$UNSW Sydney, Australia \\
\texttt{\{debajyoti22,divyansh22,akagarwal,jpatro\}@iiserb.ac.in}\\
\texttt{kodali.prashant@gmail.com, aditya.joshi@unsw.edu.au} \vspace{0.3em}
\\
\raisebox{-0.4ex}{\includegraphics[height=1em]{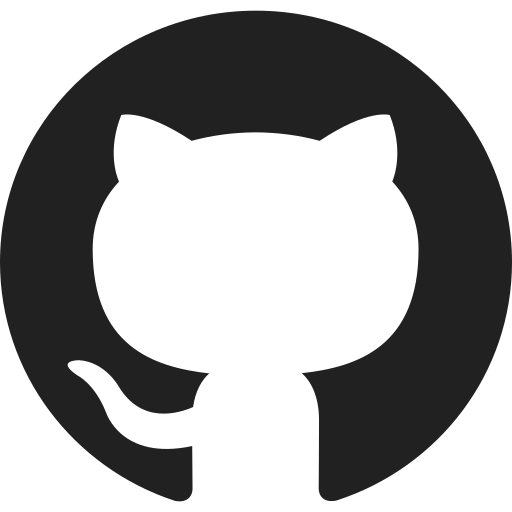}}\hspace{0.3em}\href{https://github.com/debajyotimaz/IndicKLAR_EMNLP_2026}{GitHub}\hspace{0.8em}
\raisebox{-0.4ex}{\includegraphics[height=1em]{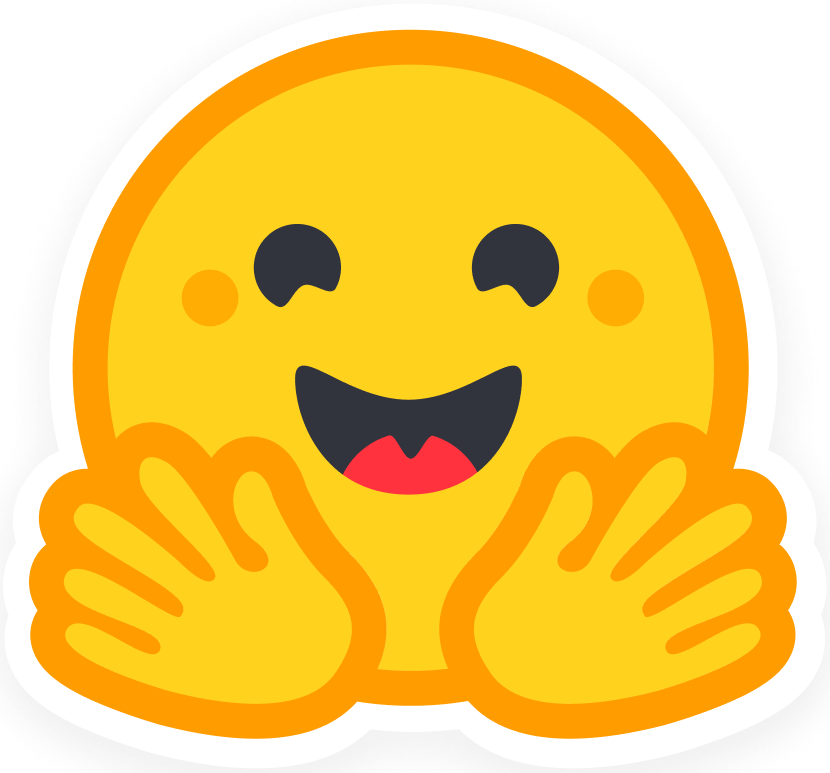}}\hspace{0.3em}\href{https://huggingface.co/datasets/debajyotimaz/IndicKLAR}{Dataset}
}

\begin{document}
\maketitle
\begin{abstract}

Large language models often exhibit a substantial gap between their performance in English and in lower-resourced languages on equivalent knowledge queries---a cross-lingual consistency issue that remains underexplored for Indian languages and their code-mixed counterparts. To study this gap, we introduce \textbf{\textsc{IndicKLAR}}, an Indic extension of the KLAR-CLC benchmark covering 18 of the 22 scheduled Indian languages. For 11 widely used language pairs, we additionally provide code-mixed variants. Both monolingual and code-mixed inputs verified by native speakers. This three-way alignment enables us to examine how knowledge recall consistency varies across English, code-mixed, and native Indian language inputs. Across nine open-weight models, we find that the accuracy gap between native-language and English inputs can reach $\sim$0.50, while code-mixed inputs substantially reduce this gap, bringing performance within $\sim$0.05 of English without any model-level intervention. Motivated by this finding, we evaluate several prompting strategies that differ in how explicitly language conversion is exposed: a two-stage translate-then-answer setup, a one-stage joint translation-and-answer prompt, and \textbf{\textit{Translate-in-Thought (TinT)}}---a single-step strategy in which the model internally converts the input and outputs only the final answer. Across the native $\rightarrow$ code-mixed $\rightarrow$ English performance trajectory, we observe a consistent \textit{flip point}---the transition from incorrect to correct prediction---between the native and code-mixed settings. Notably, this pattern holds both when the code-mixed representation is explicitly provided as input or when the model is prompted to convert internally using TinT.

\end{abstract}

\section{Introduction}

\begin{figure*}
    \centering
    \includegraphics[width=0.85\linewidth]{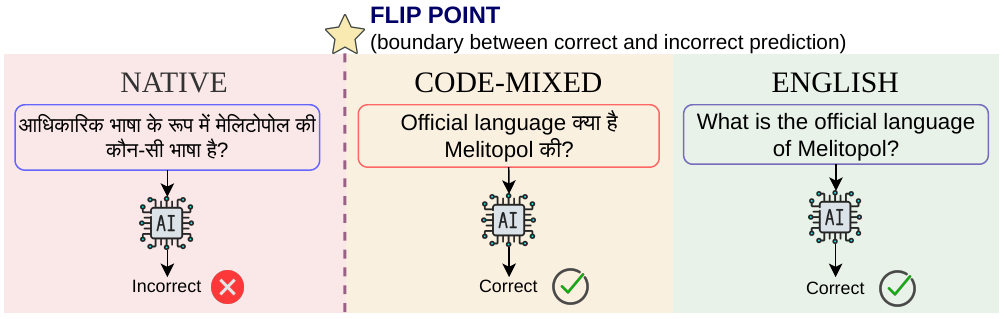}
    \caption{Illustration of the \textit{flip point}: the boundary between incorrect and correct prediction, lies between the native Indian languages and code-mixed settings.}
    \label{fig:}
\end{figure*}

Large language models (LLMs) are increasingly deployed in multilingual contexts, yet their capabilities remain unevenly distributed across languages \cite{Choudhury_Deshpande_2021,khanuja-etal-2023-evaluating,fu-liu-2025-reliable}. Performance is consistently highest for high-resource languages like English, while significantly degrading for low-resource languages---a disparity that poses a critical barrier for equitable AI deployment in linguistically diverse regions \cite{wang-etal-2025-linguistic}. A crucial step toward truly equitable multilingual systems is improving \emph{crosslingual knowledge consistency}: the ability of models to recall knowledge consistently across language variants of the same query.

This work addresses crosslingual consistency in the Indian language ecosystem, a setting of significant practical importance affecting over 1.4 billion people. India recognizes 22 scheduled languages\footnote{\url{https://en.wikipedia.org/wiki/Languages_with_official_recognition_in_India}} spanning multiple language families, yet most LLMs remain substantially undertrained on them ~\cite{niyogi-etal-2026-paramanu,khan-etal-2024-indicllmsuite}. Moreover, code-mixing between English and Indian languages is common in everyday digital communication---a phenomenon largely absent from existing crosslingual benchmarks, which overwhelmingly focus on clean monolingual inputs.

To fill this gap, we introduce \textbf{\textsc{IndicKLAR}}, an Indic extension of KLAR-CLC~\cite{wang-etal-2025-lost-multilinguality}—a crosslingual factual consistency benchmark built around structured knowledge-base triples—covering 18 of the 22 scheduled Indian languages alongside 11 code-mixed variants constructed with English. \textsc{IndicKLAR} is the first benchmark to pair monolingual Indic factual queries with code-mixed counterparts under native-speaker verification of both, at this scale. This three-way alignment enables aligned evaluation across English, native Indian languages, and naturally motivated code-mixed inputs. The code-mixed variants are constructed by interleaving English content words into the grammatical structure of each Indian language, reflecting real-world multilingual communication patterns. Both the monolingual and code-mixed variants undergo human quality verification by native or near-native speakers across all 11 language settings.

Our evaluation on \textsc{IndicKLAR} reveals a pronounced consistency gap between native and English inputs---one that, surprisingly, code-mixed variants largely close without any model-level intervention.

Drawing inspiration from bilingual cognition---where speakers responding in a non-native language often process internally in their dominant one \cite{grosjean2008studying, kroll1994category, green2013language}---we ask whether LLMs possess a latent translation capability that can be elicited through prompting alone.
This motivates Translate-in-Thought (TinT), a simple single-pass prompting strategy in which the model is instructed to perform language conversion internally and emit only the final answer (no intermediate text is generated).
Alongside TinT, we evaluate two-step and one-step joint prompting strategies, systematically varying the visibility of the intermediate conversion.

We summarize our contributions as follows:
\begin{itemize}
\item We introduce \textsc{IndicKLAR}, an Indic extension of KLAR-CLC covering 18 of the 22 scheduled Indian languages and code-mixed variants for 11 widely used language pairs---human-verified across both monolingual and code-mixed forms---and conduct a systematic consistency analysis showing that code-mixed inputs substantially reduce native-language performance gaps.
\item 
We evaluate \textit{Translate-in-Thought (TinT)}, a novel single-pass prompting strategy that instructs the model to perform language conversion internally without generating intermediate text, and systematically compare it against explicit translate-then-answer and code-mixed baselines. Our findings reveal that TinT substantially improves cross-lingual recall without requiring explicit translation.
\item Our study reveals that the \textit{flip point}---the boundary between incorrect and correct prediction---consistently falls between the native low-resource and code-mixed settings, whether viewed as a property of the input surface form or the model's internal thought process.
\end{itemize}

\section{Related Work}

\paragraph{Crosslingual Consistency Benchmarks.} Existing benchmarks for crosslingual factual consistency predominantly focus on high-resource and globally distributed language pairs, leaving the Indian language ecosystem underrepresented \citet{qi-etal-2023-cross, wang-etal-2025-lost-multilinguality}. These benchmarks cover few Indian languages and rarely include code-mixed variants that reflect real-world communication patterns. While \citet{han2025mubench} and \citet{liu2025entity} include limited evaluation on code-switched inputs, their code-switching is constructed systematically via direct placeholder substitution, which does not reflect naturalistic code-switching observed in real-world multilingual communication---and may partly explain why \citet{han2025mubench} report inconsistent gains on code-switched variants compared to monolingual baselines. More broadly, existing evaluation frameworks such as \citet{gupta-etal-2025-found} rely on translating model outputs back into English via large external systems such as NLLB-54B before comparison against gold labels---a setup that can obscure whether the low-resource output was itself faithful, and makes the evaluation pipeline comparatively expensive. \textsc{IndicKLAR} addresses these gaps by providing native-language evaluation across 18 of the 22 scheduled Indian languages with native-speaker verification of monolingual and code-mixed variants for 11 of them, without dependence on external translation systems.

\paragraph{Prompting for Multilingual Reasoning.} On the methods side, prior work has largely approached low-resource multilingual performance through training-time interventions. \citet{ai-etal-2025-knowledge} find that code-switching training and cross-lingual word alignment objectives yield the most consistent improvements across languages. However, they do not examine how inference-time prompting strategies affect model behavior in code-mixed settings. \citet{kim-etal-2025-llms} argue that activating the right internal mechanisms---including better neuron-level utilization and internal translation behavior---can substantially improve how existing multilingual knowledge is used. This suggests that prompting strategies alone may unlock stronger crosslingual consistency without retraining. The approach of \citet{wang-etal-2025-lost-multilinguality} requires extensive hyperparameter search across language-specific configurations, making it costly to adapt to new models and languages at scale. They further argue that translation-based methods are ineffective for crosslingual factual prediction. Our results directly challenge this claim, showing that translation used as an intermediate substrate substantially improves over direct prompting.


Prior multilingual prompting approaches have explored cross-lingual conversion as an explicit intermediate step. \citet{etxaniz-etal-2024-multilingual} introduce \textit{self-translate}, where the model first generates an English translation before performing inference, while \citet{huang-etal-2023-languages} use cross-lingual-thought prompting (XLT) to reformulate non-English inputs before task solving. In contrast, TinT prompts the model to reason through the language conversion without explicitly generating the intermediate translation, and emits only the final answer. We evaluate TinT alongside explicit translation and code-mixing strategies to systematically study their effects on cross-lingual factual consistency. Complementing this, \citet{gao-etal-2025-understanding} similarly find that LLMs internally pivot through an English-centric representation when retrieving cross-lingual context. Their setting, however, is context-grounded retrieval rather than closed-book factual recall, and does not consider code-mixed input.

These gaps in both data and methodology jointly motivate our work. \textsc{IndicKLAR} provides broad coverage of Indian languages, including human-verified code-mixed variants. Building on this benchmark, our prompting study---including Translate-in-Thought (TinT)---examines crosslingual consistency as an intrinsic model capability rather than one dependent on external linguistic support.

\section{IndicKLAR Benchmark}
\label{sec:dataset}

We extend \textsc{KLAR-CLC}~\cite{wang-etal-2025-lost-multilinguality} to 18 of the 22 scheduled Indian languages, creating language-aligned instances for consistent evaluation across multiple language settings.

\paragraph{Monolingual Indian Language Variants.}
We translate English \textsc{KLAR-CLC} into 18 Indian languages using the Google Translator API\footnote{\url{https://cloud.google.com/translate?hl=en}}, yielding 18 Indian-language variants aligned with the original English source : Assamese (as), Bengali (bn), Dogri (doi), Gujarati (gu), Hindi (hi), Kannada (kn), Konkani (gom), Maithili (mai), Malayalam (ml), Marathi (mr), Nepali (ne), Odia (or), Punjabi (pa), Sanskrit (sa), Sindhi (sd), Tamil (ta), Telugu (te), and Urdu (ur). The languages with their ISO codes are presented in Table~\ref{tab:languages_iso}.

\paragraph{Code-Mixed Variants.}
For 11 out of the 18 Indian languages---those for which we had linguist coverage or native speaker (Assamese, Bengali, Gujarati, Hindi, Malayalam, Marathi, Odia, Punjabi, Sanskrit, Tamil, and Telugu)---we manually constructed code-mixed variants using GPT-5.3 Instant\footnote{\label{gpt_API}\url{https://chatgpt.com/}}. 
The code-mixed variants are generated from the original English \textsc{KLAR-CLC} queries, using the corresponding Indic-language templates as grammatical scaffolds. This preserves the underlying fact triples, candidate sets, and gold answers across the three variants by construction. The GPT-based template generation procedure and prompt are provided in Appendix~\ref{app:cm_generation}. Both the monolingual and code-mixed variants for these 11 languages were manually verified by native speakers for fluency and naturalness of English--Indic mixing; annotator demographics are presented in \S\ref{para:human_veri_demographics}.

\paragraph{Dataset Statistics.}
\textsc{IndicKLAR} provides three-way parallel variants (monolingual native Indian, code-mixed, and English) for 11 languages, and two-way parallel variants (monolingual native Indian and English) for the remaining 7. Following \textsc{KLAR-CLC}~\cite{wang-etal-2025-lost-multilinguality}, each language setting contains 2,619 instances spanning 20 fact categories. Number of samples, average query lengths, and average answer lengths are reported in Table~\ref{tab:dataset_stats} in the Appendix.


\section{Experiments}

\subsection{Baseline and Trajectory}

We evaluate model robustness across three input settings: (i) \textbf{English (EN)}---the high-resource upper bound, (ii) \textbf{Native (NA)}---the query in original script, and (iii) \textbf{Code-Mixed (CM)}---the query in code-mixed form, retaining the native Indic script for the source-language structure while interleaving English content words in Latin script. This comparison is designed to test whether code-mixing enables models to leverage stronger English reasoning while preserving low-resource semantics.

\subsection{Prompting Strategies}

Beyond direct prompting, we investigate several conversion-based strategies designed to bridge the crosslingual gap, categorized into three groups.

\textbf{2-Step Strategies:} These decouple the transformation and answering tasks into two separate inference calls. \textit{2Step-CM} first converts the query into a code-mixed form, while \textit{2Step-EN} translates it fully into English, after which the final answer is generated from the reformulated input. \textit{2Step-Translit} transliterates the query into Roman script without translation, isolating the effect of script conversion alone.

\textbf{1-Step Strategies:} These request both the transformation and the final answer in a single prompt. \textit{1Step-CM+Ans} generates a code-mixed conversion alongside the answer, and \textit{1Step-EN+Ans} does the same with an English translation.

\textbf{Translate-in-Thought (TinT):} The TinT strategy instructs the model, via a single user-prompt directive, to internally reason through an English (\textit{TinT-EN}) or code-mixed (\textit{TinT-CM}) representation and emit only the final answer---unlike explicit chain-of-thought, no intermediate text is generated. This tests the model's ability to utilize latent multilingual reasoning while maintaining inference efficiency.

\noindent Detailed prompts are presented in Appendix~\ref{app:prompting_strategies}.

\begin{table}[t]
\centering
\footnotesize
\renewcommand{\arraystretch}{1.08}
\setlength{\tabcolsep}{2pt}
\resizebox{\columnwidth}{!}{
\begin{tabular}{@{}>{\raggedright\arraybackslash}p{0.18\columnwidth}>{\raggedright\arraybackslash}p{0.26\columnwidth}>{\raggedright\arraybackslash}p{0.46\columnwidth}@{}}
\toprule
\textbf{Strategy} & \textbf{Transformation} & \textbf{Procedure} \\
\midrule
Base (EN/NA/CM) & None & Direct prompting on the source query in the given language. \\
2Step-CM & Code-mix conversion & Convert to code-mixed form, then answer in a second call. \\
2Step-EN & English translation & Translate to English, then answer in a second call. \\
2Step-Translit & Roman transliteration & Transliterate to Roman script without translation, then answer in a second call. \\
1Step-CM+Ans & Code-mix conversion & Emit conversion and answer in a single pass. \\
1Step-EN+Ans & English translation & Emit translation and answer in a single pass. \\
TinT-CM & Latent code-mixed reasoning & Reason internally in code-mix; output only the answer. \\
TinT-EN & Latent English reasoning & Reason internally in English; output only the answer. \\
\bottomrule
\end{tabular}
}
\caption{Prompting strategies evaluated in our experiments. All start from the source query.}
\label{tab:prompt-configs}
\end{table}

\subsection{Experimental Setup}
\textbf{Models:} We evaluate open-weight language models from multiple families: \textit{Llama 3.1}, \textit{Llama 3.2}, \textit{Gemma 3}, and \textit{Qwen 2.5}, spanning roughly 1B--14B parameters. Per-model details are provided in Appendix Table~\ref{tab:appendix-model-details}.

\noindent\textbf{Metrics:} We report two primary metrics: (i) Accuracy---the percentage of correct predictions, and (ii) Cross-Lingual Consistency (CLC)---a measure of prediction overlap across language pairs. These metrics are defined in Appendix \ref{app:eval_metrics}.

\noindent\textbf{Reproducibility:} All models are evaluated with greedy decoding (temperature$=$0, top-$p$$=$1.0, \texttt{max\_new\_tokens}$=$10) using each model's official chat template, with 3-shot in-context examples drawn from the English source split. Experiments use a single fixed seed (12345) and run on single NVIDIA A100 GPU (80GB).

\begin{figure*}[!htbp]
  \centering
  \includegraphics[width=\textwidth]{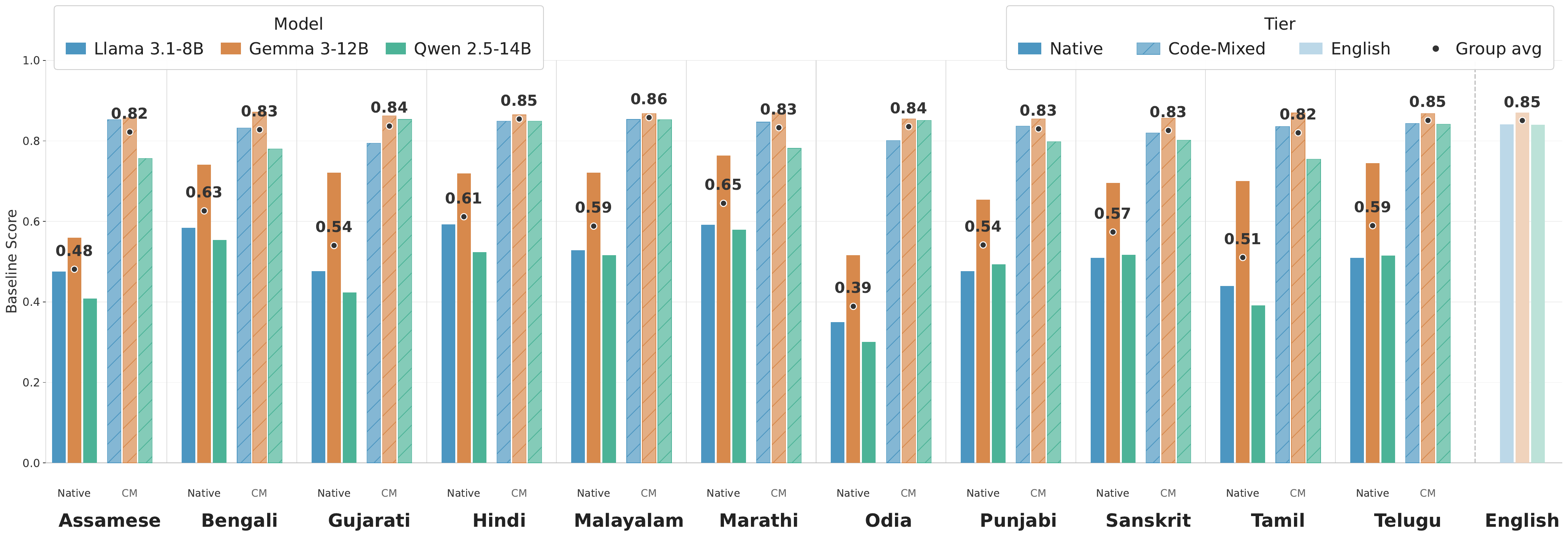}
  \caption{Baseline accuracy scores across Native (solid), Code-Mixed (hatched), and English (faded) input tiers for eleven Indic languages and English (En) using LLaMA-3.1-8B, Gemma-3-12B, and Qwen-2.5-14B. Black dots denote the average accuracy across the three models for each input tier.}
  \label{fig:eng_native_cm_comparison}
\end{figure*}

\section{Results}
\label{sec:results}

We investigate how model performance varies across three input conditions---native script, code-mixed (CM), and English---across nine models spanning three families, and organize the discussion by the question each subsection answers: the English--Native gap (\S\ref{sec:res_gap}), prompting strategies (\S\ref{sec:res_prompting}), model-size scaling (\S\ref{sec:res_scaling}), language-family and category effects (\S\ref{sec:res_lang_cat}), pairwise CLC (\S\ref{sec:res_pairwise}), romanization and comparison with prior work (\S\ref{sec:res_roman_litcomp}), and generalization and the location of \textit{flip point} (\S\ref{sec:res_gen_flip}).

\subsection{The English--Native Gap}
\label{sec:res_gap}

Figure~\ref{fig:eng_native_cm_comparison} reveals two consistent patterns. 
First, the Native tier exhibits severe degradation compared to English, particularly for smaller models, where accuracy scores approach 0.10 on Assamese and Telugu. This degradation is strongly associated with longer tokenization in native-script inputs: across the 18 native Indic languages, average tokenization length is negatively correlated with language-wise accuracy for all three evaluated models (Pearson $r=-0.717$ to $-0.570$, all $p<0.05$; see Appendix~\ref{app:tokenization}).
Second, CM scores consistently approach English-level performance across all languages and model scales, the average gap across the three model remaining within ${\sim}0.05$, compared to a Native--English gap of up to ${\sim}0.50$. The observed gains suggest that code-mixing can substantially reduce the lower-resource performance gap through input formulation alone, without modifying the model.

\subsection{Prompting Strategies}
\label{sec:res_prompting}

\begin{figure*}[h]
    \centering
    \includegraphics[width=\textwidth]{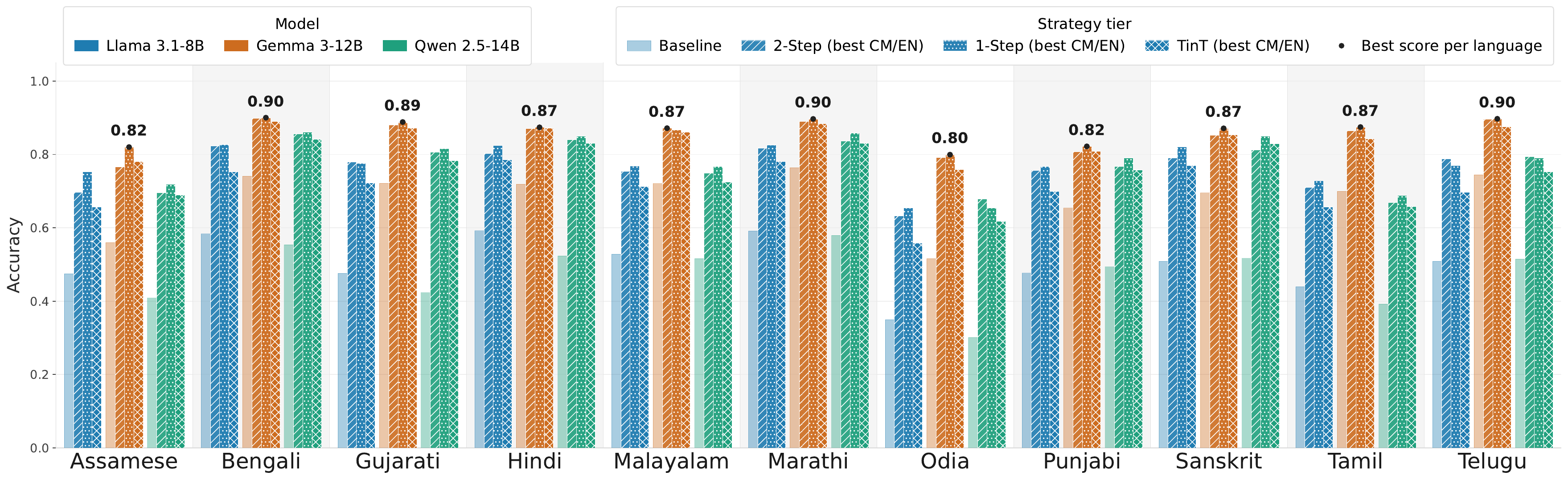}
    \includegraphics[width=\textwidth]{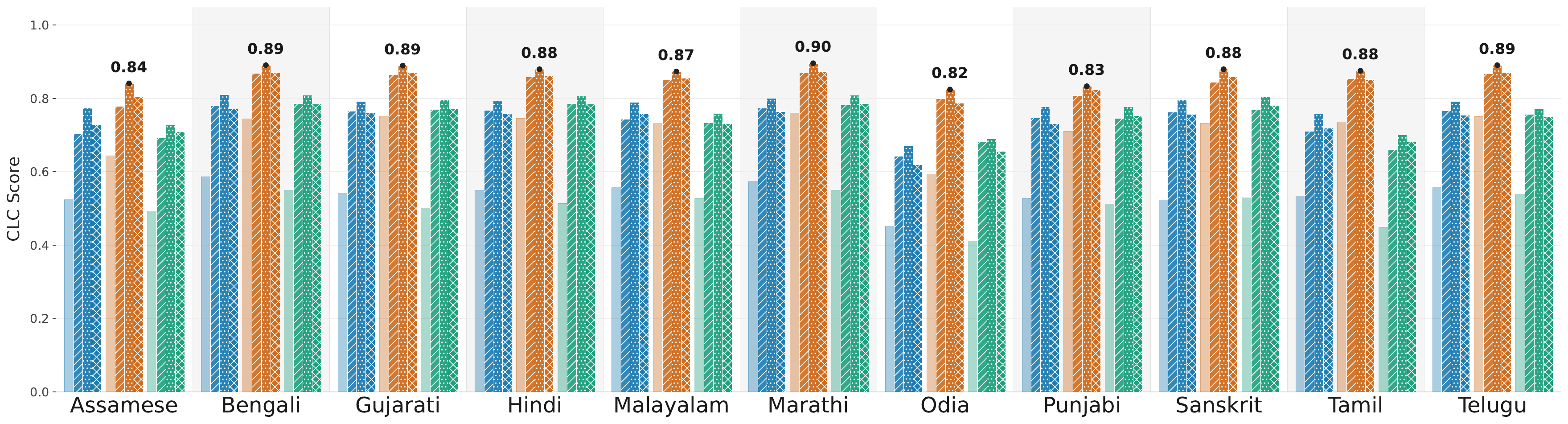}
    \caption{Performance of all prompting strategies across eleven Indian languages on two evaluation metrics: accuracy (top row) and CLC score (bottom row). The numerical scores for each model are reported in Appendix~\ref{app:acc_clc_main_exp_results}.}
    \label{fig:prompting_results_acc_clc}
\end{figure*}

\begin{figure}[h]
    \centering
    \includegraphics[width=\linewidth]{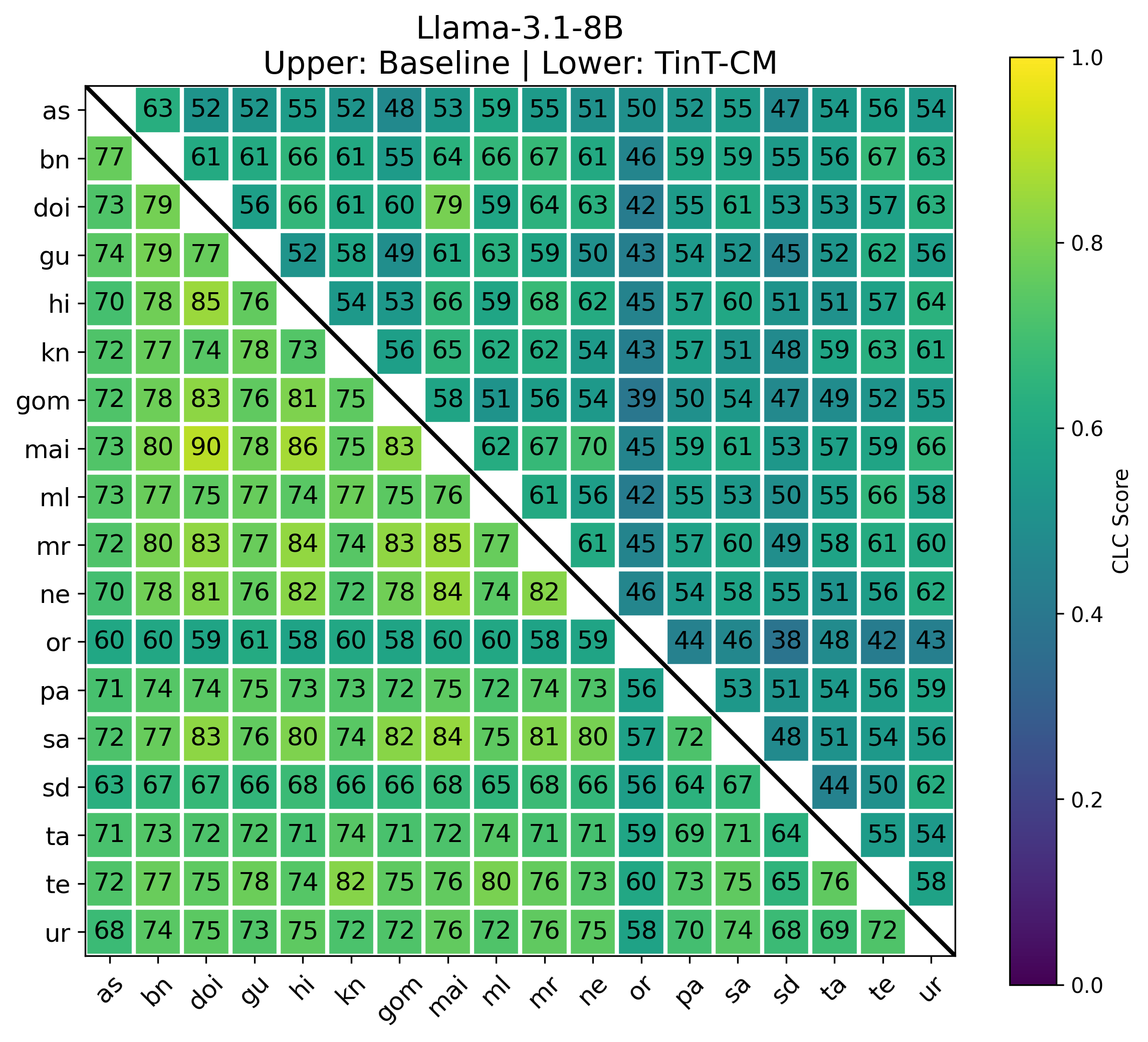}
    \caption{
    Pairwise cross-lingual consistency (CLC) comparison for \textbf{Llama-3.1-8B-Instruct}. The upper triangular region shows the baseline CLC scores, while the lower triangular region corresponds to \textbf{TinT-CM}. 
    }
    \label{fig:triangular_heatmap_llama31_8b_titcm}
\end{figure}

Figure~\ref{fig:prompting_results_acc_clc} consolidates the performance scores across 11 language pairs. 
Across families, selected translation-based and code-mixed strategies reduce the gap to English, often yielding stronger performance than direct low-resource prompting. Our proposed \textbf{TinT} prompting shows statistically significant gains over the baseline (see Appendix~\ref{app:Statistical_significance}) while also offering the practical benefit of lower inference time compared to other prompting strategies (Appendix~\ref{app:inference_time_analysis}).

\subsection{Model-Size Scaling}
\label{sec:res_scaling}

\begin{figure}[h]
    \centering
    \includegraphics[width=\linewidth]{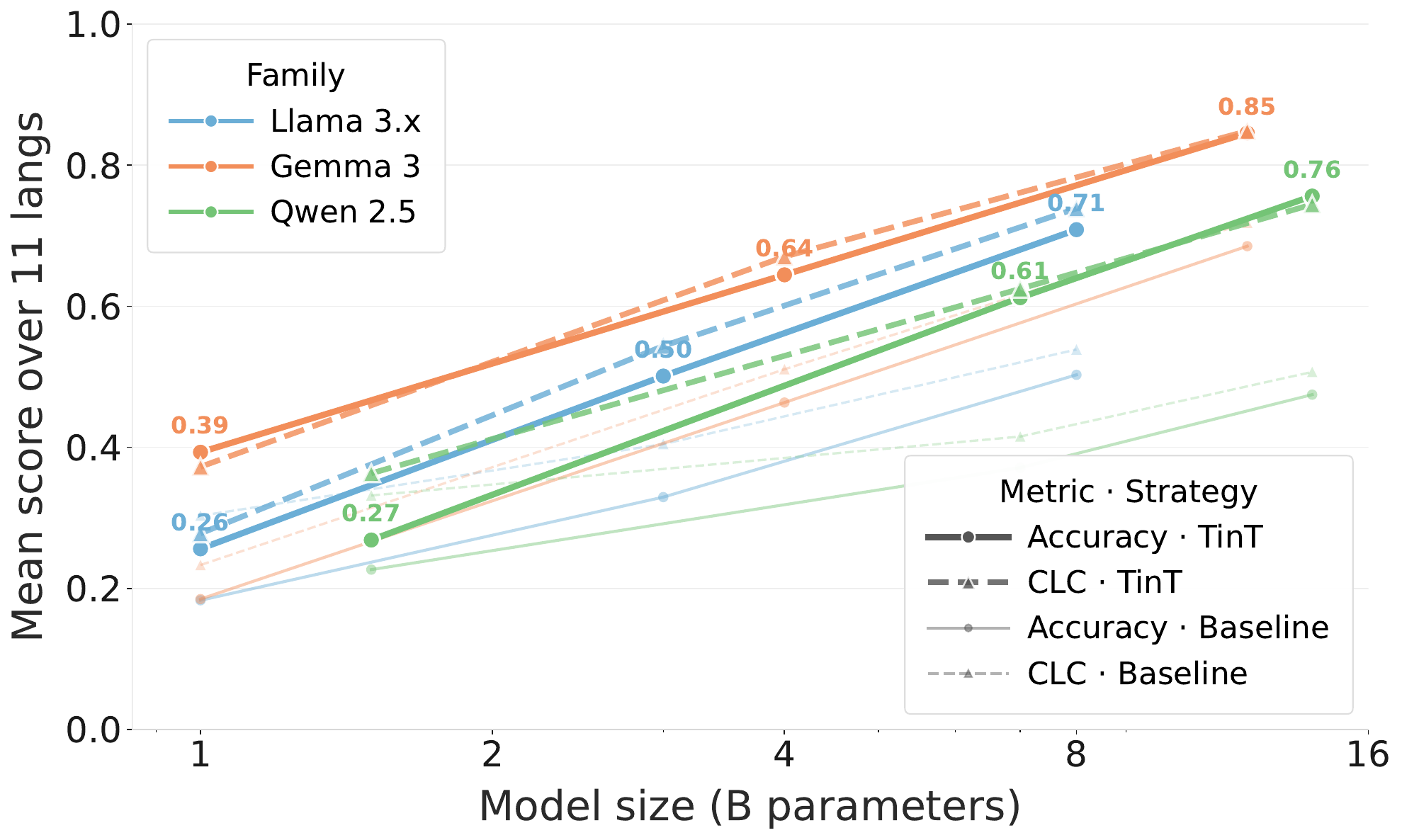}
    \caption{Mean Accuracy and CLC of TinT and Baseline strategies across 11 Indic languages.
    }
    \label{fig:scaling_model_size}
\end{figure}

Figure~\ref{fig:scaling_model_size} reveals a consistent scaling trend across all three model families: as model size increases, both accuracy and CLC improve monotonically for TinT, with TinT uniformly outperforming the baseline at every scale---mean accuracy rises from 0.26 at 1B to 0.71 at 8B for Llama, and from 0.39 to 0.85 for Gemma over the same range. The gap between TinT and the baseline widens with model size, suggesting that larger models possess stronger latent translation ability. Gemma leads across all sizes, followed by Qwen and Llama. At the lower end, prompting strategies occasionally underperform the baseline for very small models ($<$1.5B), indicating a minimum capacity threshold below which reformulation is difficult. This degradation is not universal: while \texttt{Llama-3.2-1B} and \texttt{Qwen2.5-1.5B} show drops under several strategies, \texttt{Gemma-3-1B} consistently improves across all strategies and languages, possibly owing to stronger multilingual representations induced by its pretraining data composition. More details are provided in Appendix \ref{app:scaling_tint_analysis}.

\begin{figure*}[h]
\centering


\begin{subfigure}[h]{0.30\textwidth}
    \centering
    \includegraphics[width=\linewidth]{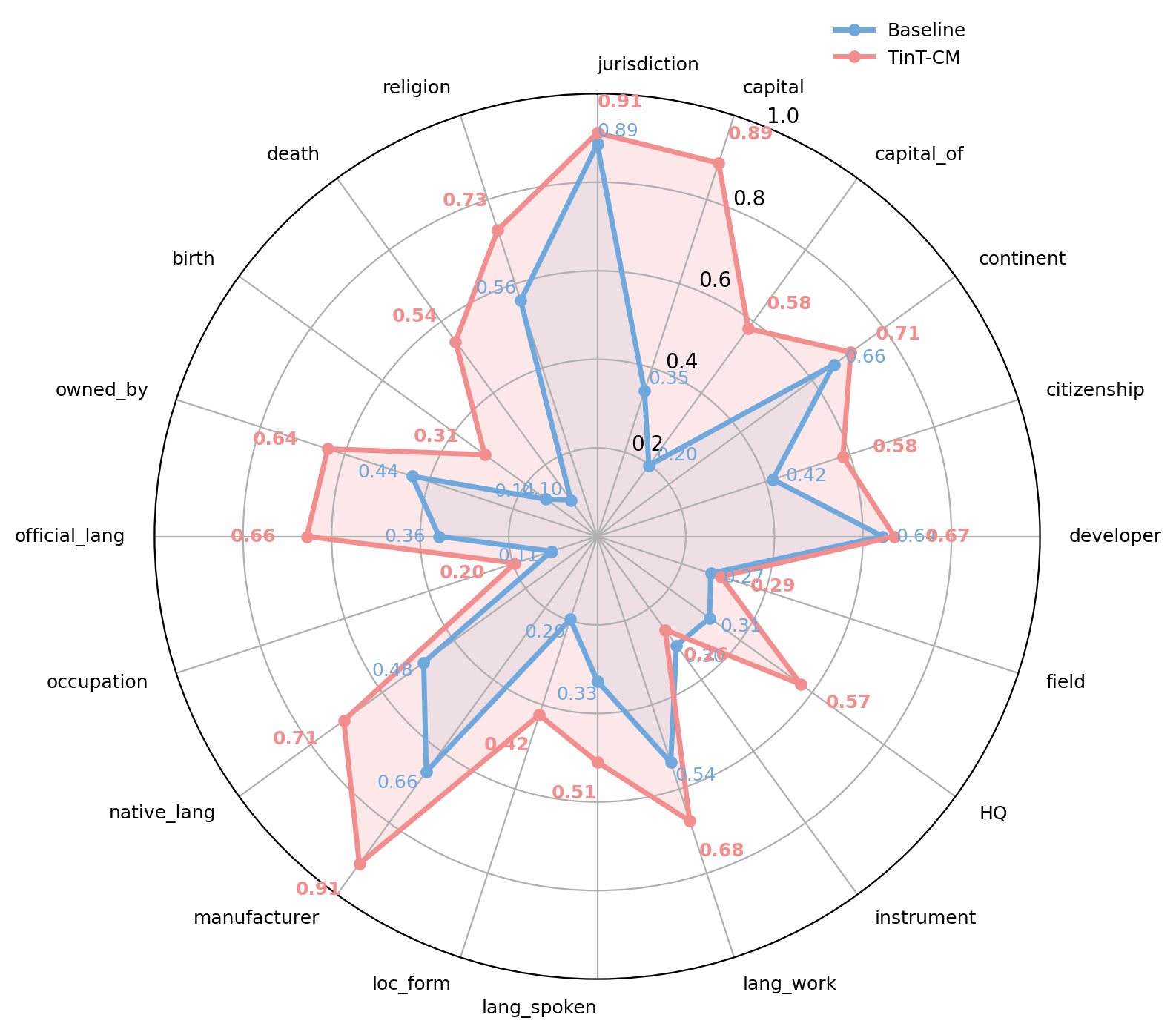}
    \caption{Assamese}
\end{subfigure}
\hfill
\begin{subfigure}[h]{0.30\textwidth}
    \centering
    \includegraphics[width=\linewidth]{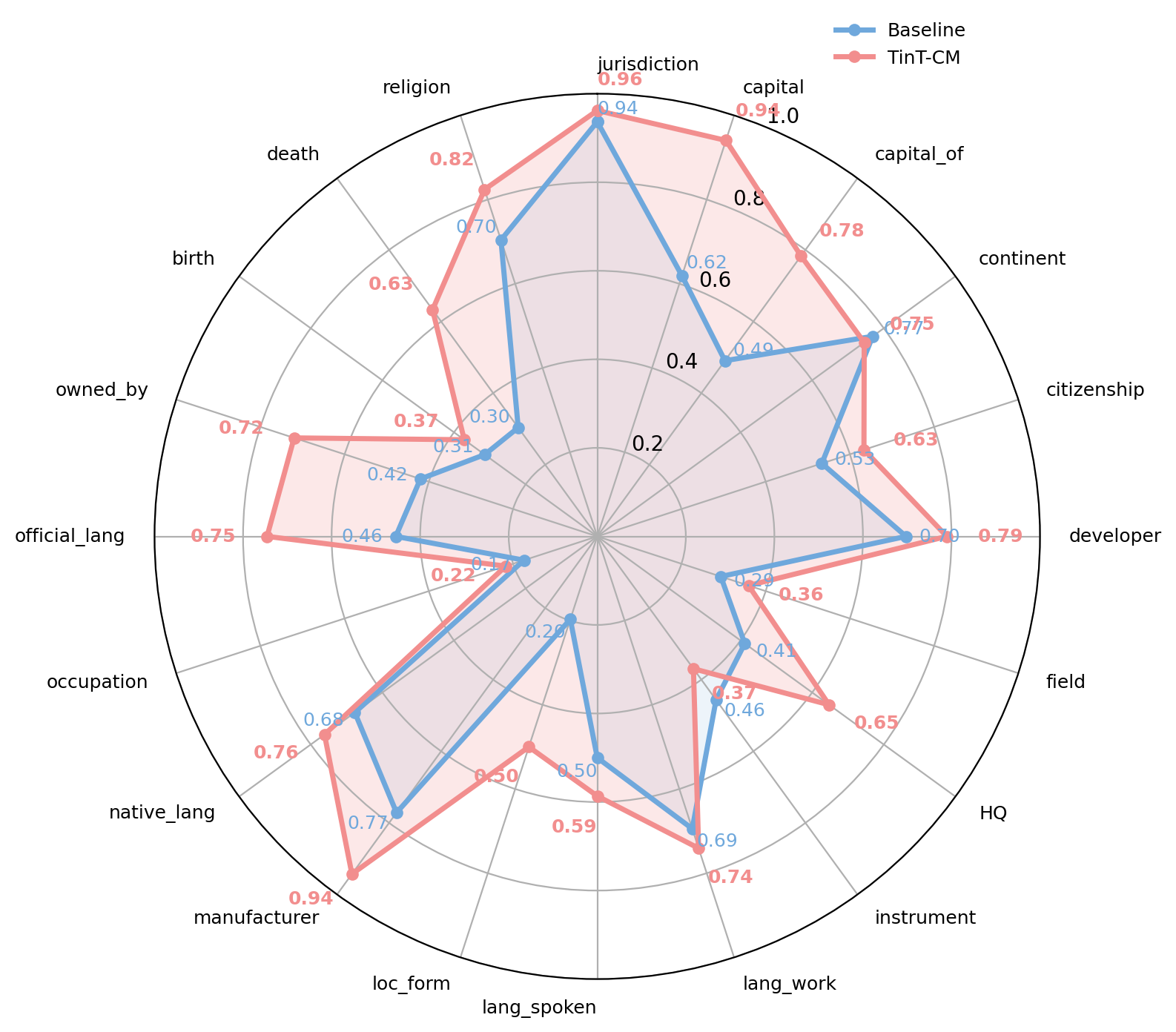}
    \caption{Bengali}
\end{subfigure}
\hfill
\begin{subfigure}[h]{0.30\textwidth}
    \centering
    \includegraphics[width=\linewidth]{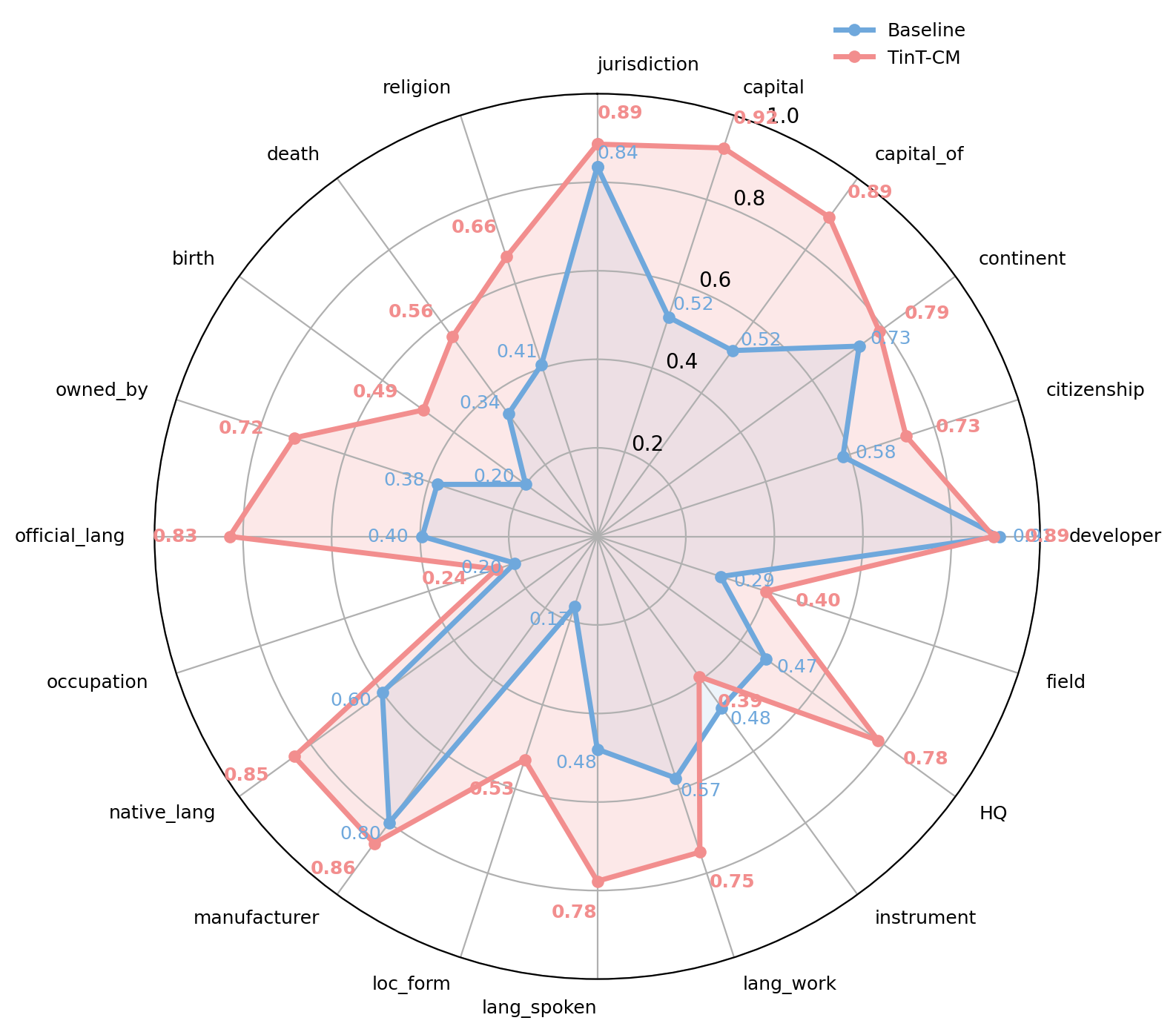}
    \caption{Hindi}
\end{subfigure}

\vspace{0.5em}


\begin{subfigure}[h]{0.30\textwidth}
    \centering
    \includegraphics[width=\linewidth]{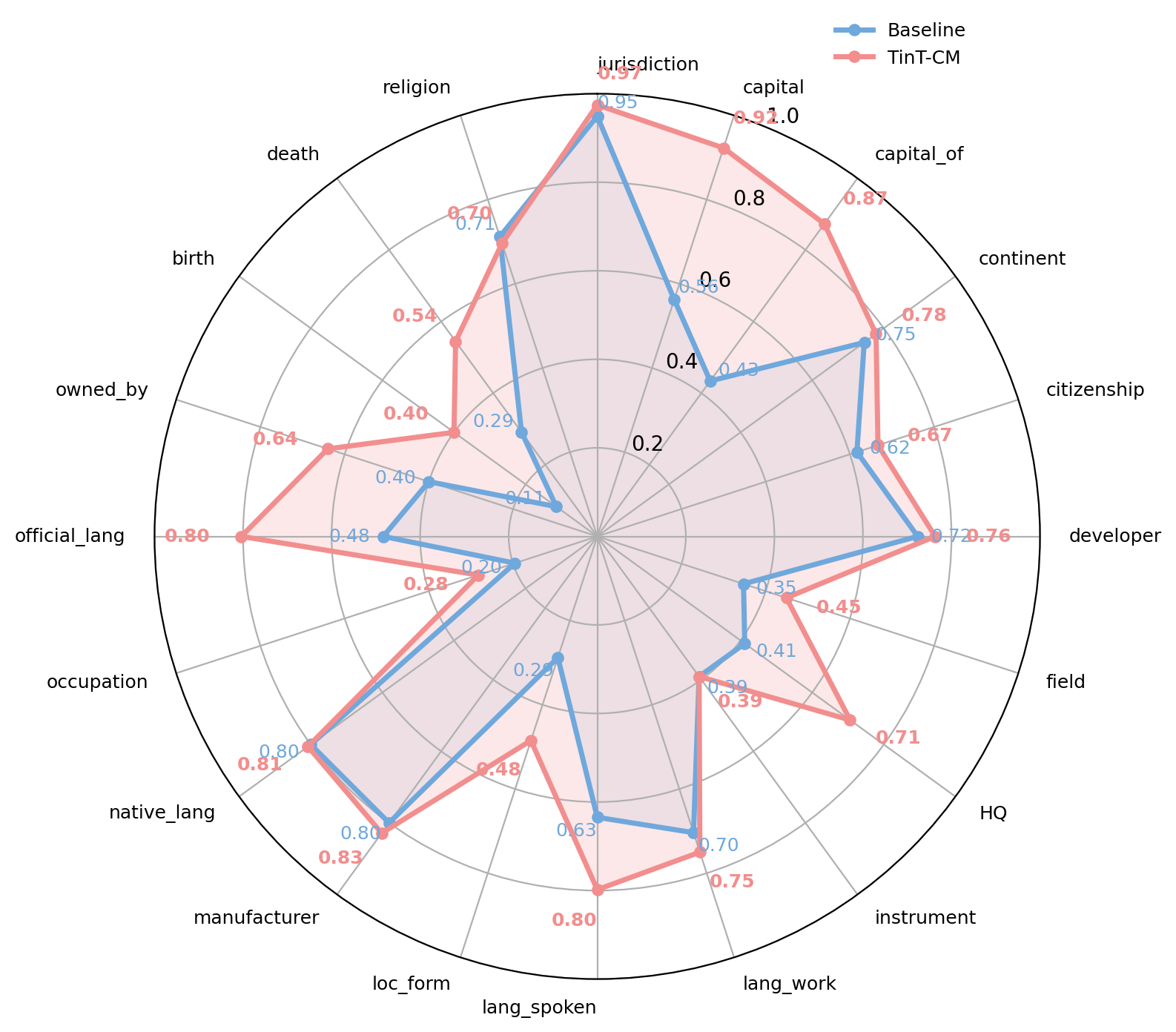}
    \caption{Marathi}
\end{subfigure}
\hspace{0.08\textwidth}
\begin{subfigure}[h]{0.30\textwidth}
    \centering
    \includegraphics[width=\linewidth]{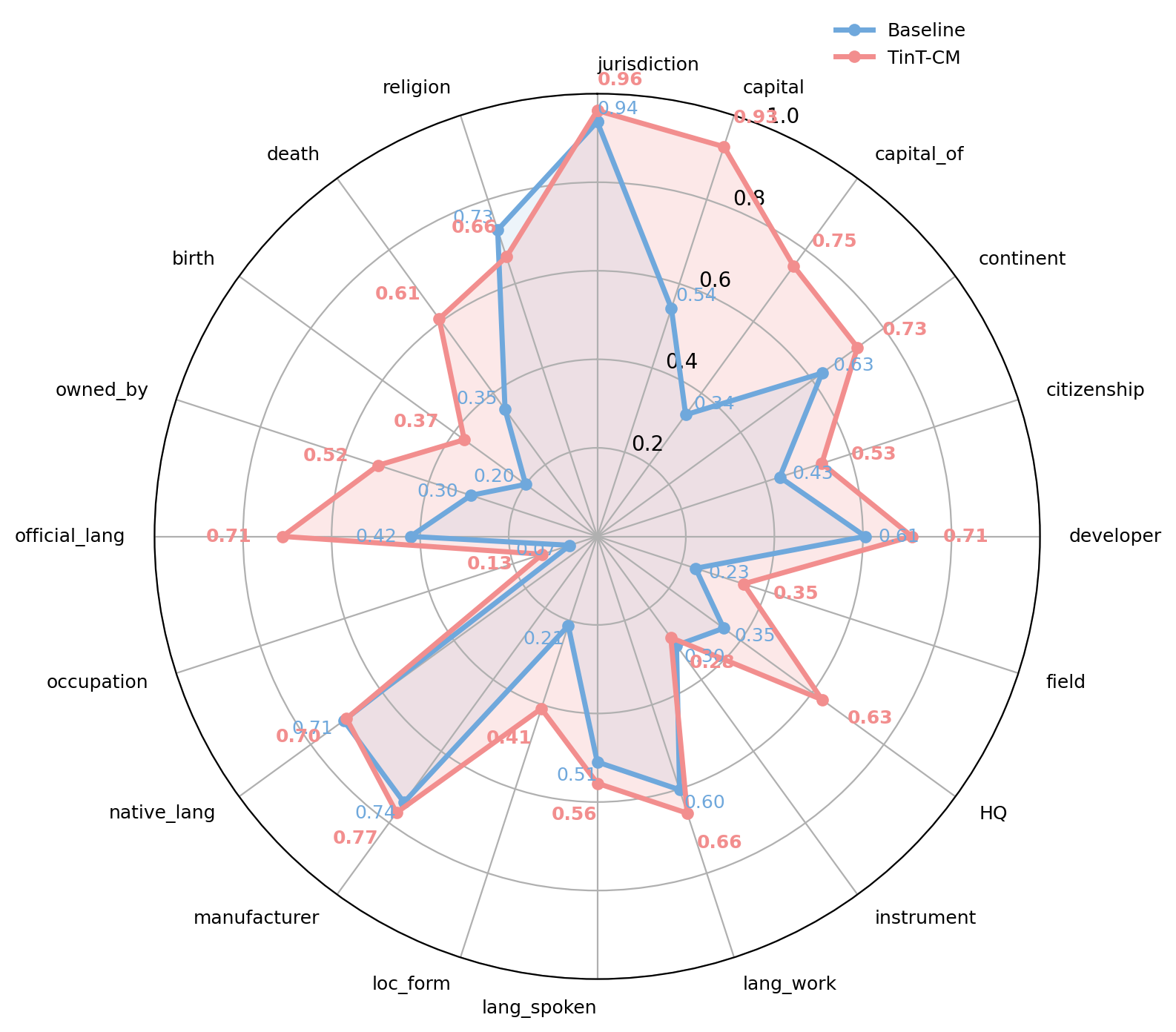}
    \caption{Telugu}
\end{subfigure}

\caption{
Accuracy of different category of facts across languages.Here, we present scores of only five representative languages; the remaining languages are shown in Figure~\ref{fig:radar_remaining_languages} in the Appendix.}
\label{fig:radar_fact_category_avg}

\end{figure*}

\subsection{Language Family and Category Effects}
\label{sec:res_lang_cat}

\paragraph{Language Groups.} We group the evaluated languages into two families: Indo-Aryan (\texttt{as}, \texttt{bn}, \texttt{doi}, \texttt{gu}, \texttt{hi}, \texttt{gom}, \texttt{mai}, \texttt{mr}, \texttt{ne}, \texttt{or}, \texttt{pa}, \texttt{sa}, \texttt{sd}, \texttt{ur}) and Dravidian (\texttt{kn}, \texttt{ml}, \texttt{ta}, \texttt{te}). TinT achieves stronger gains for Indo-Aryan languages (avg.\ $+18\%$ accuracy, $+14\%$ CLC) compared to Dravidian languages ($+14\%$ accuracy, $+13\%$ CLC).

\paragraph{Category Analysis.} We analyze TinT prompting performance across 20 fact categories per language using radar charts to compare baseline and TinT accuracy. Figure~\ref{fig:radar_fact_category_avg} reveals: (i) TinT consistently improves across most categories for all languages. (ii) Hindi and Bengali show uniform improvement, indicating TinT generalizes well to better-resourced languages. (iii) Assamese shows uneven gains (15-20\% for some categories, minimal for others), suggesting inconsistent coverage for low-resource facts. (iv) Odia shows minimal improvement, indicating implicit translation cannot overcome fundamental representation gaps. (v) Semantic and cultural fact categories remain challenging across all languages, highlighting persistent limitations.

\subsection{Pairwise Cross-Lingual Consistency}
\label{sec:res_pairwise}

To understand whether models maintain consistent predictions across all language pairs, we compute cross-lingual consistency (CLC) separately for each pair of the 18 languages, comparing baseline and TinT-CM predictions for all 18 languages using Llama-3.1-8B-Instruct. Figure~\ref{fig:triangular_heatmap_llama31_8b_titcm} reveals three key patterns. (i) Baseline CLC values span a wide range with high variance (0.4--0.8), whereas (ii) TinT-CM consistently improved scores, converging CLC scores to a higher range (0.5--0.9), suggesting that implicit translation enforces higher range cross-lingual consistency. (iii) Notably, Odia emerges as an outlier with the lowest CLC scores in both baseline and TinT-CM settings (0.38--0.60), likely due to limited pretraining data and script complexity, suggesting that implicit translation alone cannot overcome fundamental representational gaps for severely under-resourced languages.

\subsection{Romanization Ablation and Literature Comparison}
\label{sec:res_roman_litcomp}

\begin{figure}[h]
  \centering
  \includegraphics[width=\columnwidth]{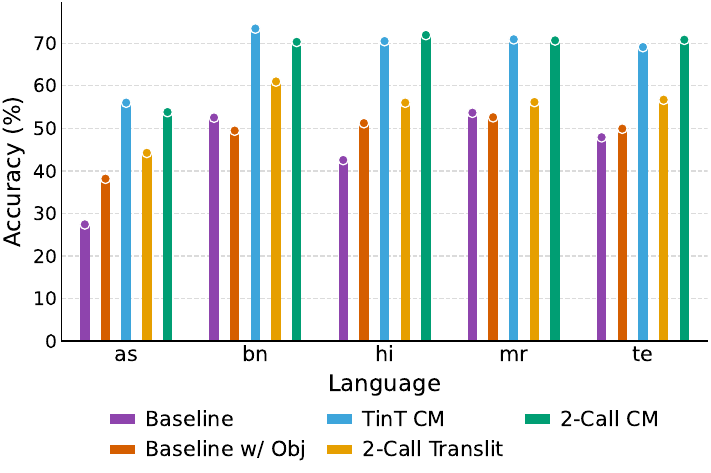}
  \caption{Ablation isolating the effect of code-mixing from romanization.}
  \label{fig:ablation_romanization}
\end{figure}

\paragraph{Ablation of code-mixing vs. romanization.} To isolate whether performance gains from code-mixed conversion arise from lexical English anchors or from script uniformity alone, we compare three variants: baseline (native script), romanized (full transliteration to Latin script), and code-mixed (English-Indic mixing). Figure~\ref{fig:ablation_romanization} shows that code-mixed variants substantially outperform both native-script (low-res.) and romanized baselines across languages, with improvements of 10-20\% accuracy. Notably, romanized variants provide suboptimal gains over native-script inputs, indicating that the benefits of code-mixing stem from meaningful bilingual lexical cues rather than script conversion alone. 
This distinction is also evident qualitatively in Figure~\ref{fig:qualitative_examples},
where transliteration fails to recover the correct prediction in several cases, while the corresponding code-mixed representation succeeds.


\paragraph{Comparison with prior work.}
Table~\ref{tab:prior_comparison} shows that our approach (TinT) achieves significantly higher accuracy compared to previous crosslingual consistency evaluation methods.

\begin{table}[h]
\centering
\small
\setlength{\tabcolsep}{2pt}
\resizebox{\columnwidth}{!}{%
\begin{tabular}{lcccccccccc}
\toprule
& \multicolumn{2}{c}{\textbf{AS}} 
& \multicolumn{2}{c}{\textbf{BN}} 
& \multicolumn{2}{c}{\textbf{HI}} 
& \multicolumn{2}{c}{\textbf{MR}} 
& \multicolumn{2}{c}{\textbf{TE}} \\
\cmidrule(lr){2-3} 
\cmidrule(lr){4-5} 
\cmidrule(lr){6-7}
\cmidrule(lr){8-9}
\cmidrule(lr){10-11}
\textbf{Method} 
& \textbf{ACC} & \textbf{CLC} 
& \textbf{ACC} & \textbf{CLC} 
& \textbf{ACC} & \textbf{CLC}
& \textbf{ACC} & \textbf{CLC}
& \textbf{ACC} & \textbf{CLC} \\
\midrule

\textsc{SubInj}
& 0.533 & 0.614 
& 0.680 & 0.692 
& 0.589 & 0.632
& 0.649 & 0.691
& 0.635 & 0.669 \\

\textsc{CoT} 
& 0.496 & 0.574
& 0.641 & 0.655
& 0.656 & 0.643
& 0.664 & 0.649
& 0.523 & 0.592 \\

\midrule

\textbf{Ours} 
& 0.656 & 0.741
& 0.753 & 0.796 
& 0.786 & 0.783
& 0.781 & 0.791
& 0.697 & 0.764 \\

\bottomrule
\end{tabular}%
}
\caption{Comparison with prior work \textsc{SubInj} \cite{liu2025entity} , including Chain-of-Thought (\textsc{CoT}) \cite{10.5555/3600270.3602070} prompting, on five languages.}
\label{tab:prior_comparison}
\end{table}

\subsection{Generalization and Flip Point}
\label{sec:res_gen_flip}

\paragraph{Generalization.} TinT generalizes beyond \textsc{IndicKLAR}: evaluated on a contextually mediated knowledge recall benchmark~\cite{liu2026evaluating} translated into five representative Indian languages, TinT consistently improves over the baseline across all languages and model families (see Appendix~\ref{app:Generalization_Knowledge_Recall}).

\paragraph{Locating the Flip Point.}
To locate the flip point precisely, we compute transition counts across the trajectory Baseline $\rightarrow$ TinT-CM $\rightarrow$ TinT-EN for five representative languages and three models (Table~\ref{tab:flip_transitions}). For every instance that the baseline answers incorrectly, we record whether it becomes correct under TinT-CM and, of those that remain incorrect, whether it becomes correct under TinT-EN. The large majority of recoveries occur at the first boundary (Baseline $\rightarrow$ TinT-CM), while the subsequent TinT-CM $\rightarrow$ TinT-EN transition yields only a small residual gain (Figure~\ref{fig:flip_point}). The detailed scores are provided in Appendix-\ref{app:flip_point}. This pattern holds uniformly across model families and languages, providing direct empirical evidence that the flip point lies between the native and code-mixed regimes, whether the code-mixed representation is supplied explicitly or induced internally by TinT.

\begin{figure}[t]
    \centering
    \includegraphics[width=\columnwidth]{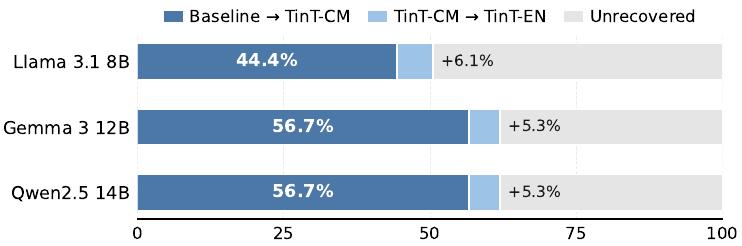}
    \caption{
Prediction recovery along the Baseline $\rightarrow$ TinT-CM $\rightarrow$ TinT-EN trajectory across five representative languages. Percentages show the proportion of baseline-incorrect instances recovered at each transition; the remaining instances are unrecovered.
}
    \label{fig:flip_point}
\end{figure}

\section{Analysis}
\label{sec:analysis}

\subsection{Interpretability Case Study}
\label{sec:interpretability_case_study}

\begin{figure}[h]
    \centering
    \includegraphics[width=\columnwidth]{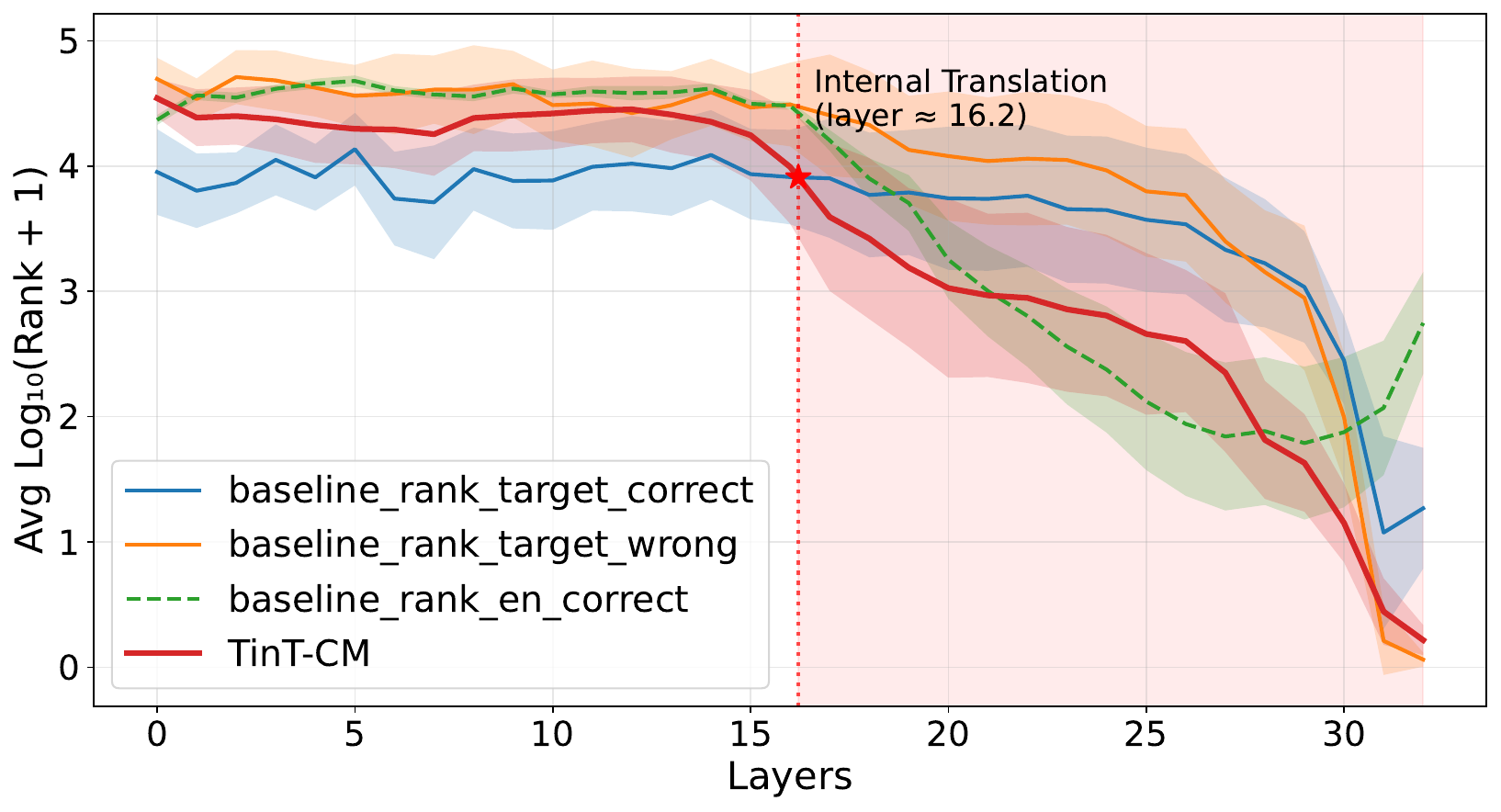}
    \caption{Layer-wise rank of the gold answer token among the model's output distribution for \textbf{Base} prompting and the best \textbf{TinT} variant across layers of Llama-3.1-8B-Instruct. Shaded bands show the standard deviation across languages.}
    \label{fig:rank-tit-vs-base}
\end{figure}

\begin{figure*}[h]
    \centering
    \includegraphics[width=0.9\textwidth]{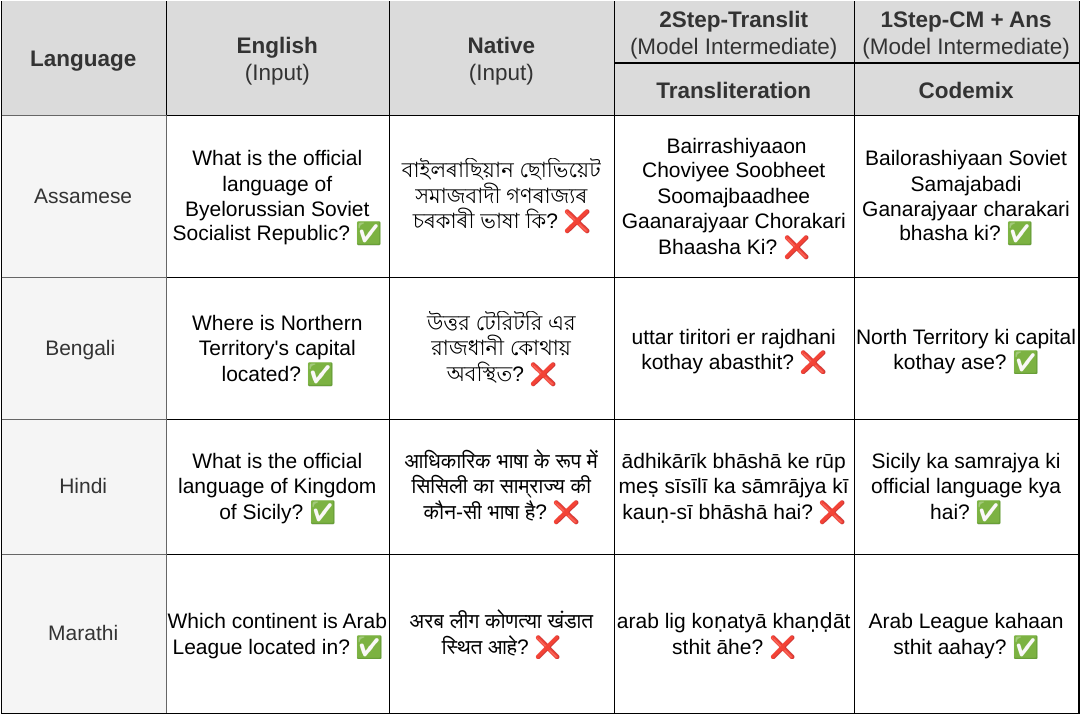}
    \caption{
    Qualitative comparison of different prompt representations across Assamese,
    Bengali, Hindi, and Marathi. A tick and cross denote correct and incorrect
    predictions, respectively.
    }
    \label{fig:qualitative_examples}
\end{figure*}
We focus on a single representative model (Llama-3.1-8B-Instruct); the trends are averaged across all 18 Indian languages and should be read as illustrative rather than exhaustive.
At each layer, we compute the rank of the gold answer token within the model's predicted distribution: a rank of 1 means the model assigns it the highest probability, while higher ranks indicate it is deprioritised relative to other tokens (see Appendix~\ref{sec:layerwise-rank-analysis} for details). Figure~\ref{fig:rank-tit-vs-base} tracks four such trajectories over samples the baseline answered \emph{incorrectly}: under baseline, the correct (\texttt{target\_correct}) and finally-generated incorrect (\texttt{target\_wrong}) target tokens and the English equivalent of the correct answer (\texttt{en\_correct}); and the correct target token under \texttt{TinT-CM} on the same indices. The TinT-CM and baseline curves remain largely indistinguishable through the first $\sim$16 layers. After this, TinT-CM separates: the gold answer's rank drops steeply toward zero in later layers, while the baseline's stays elevated. This gap is concentrated in the deeper layers, consistent with late-layer cross-lingual reorganization---though not by itself direct evidence of it.

\subsection{Error Analysis}
\label{sec:error_analysis}

To understand why low-resource recall fails and how conversion-based strategies help, we qualitatively compare prompting behaviors across four languages. 
We observe three recurring failure modes. First, the model sometimes produces little or no meaningful code-mixed conversion, instead reproducing the original native-language query or relying primarily on transliteration; this limits the benefit of English lexical anchors. For example, the Assamese case in Figure~\ref{fig:qualitative_examples} shows this behavior, where transliteration alone fails while the code-mixed representation leads to the correct prediction.Second, when intermediate transliteration is used, the model occasionally generates entity names that are phonetically similar to the intended entities but contain incorrect or uncommon spellings, leading to mismatches. Third, conversion is sometimes incomplete: only the subject or entity is transliterated or code-mixed while the remainder of the query is left unchanged, weakening the overall effect of the strategy.




\section{Conclusion}

We introduced \textsc{IndicKLAR}, an Indic extension of KLAR-CLC covering 18 of the 22 scheduled Indian languages and code-mixed variants for 11 widely used language pairs---human-verified across both monolingual and code-mixed forms---enabling systematic evaluation of crosslingual knowledge consistency across a linguistically diverse and practically significant setting. Our evaluation reveals a pronounced consistency gap between English and native Indian language settings, whereas code-mixed variants consistently narrow this gap.

To further reduce this gap without requiring an explicit translation step, we introduced \emph{Translate-in-Thought (TinT)} prompting, a single-step implicit strategy that instructs models to perform language conversion internally without explicit intermediate outputs. TinT shows statistically significant gains over the baseline across languages and model families while offering lower inference time compared to other prompting strategies. We further show that TinT generalizes beyond \textsc{IndicKLAR} to contextually mediated knowledge recall, suggesting that implicit translation prompting is broadly effective for crosslingual factual knowledge recall. The effectiveness of TinT improves with model size, which we attribute to larger models possessing stronger latent translation ability. Our interpretability case study (\S\ref{sec:interpretability_case_study}) shows that the layer-wise rank gap between TinT and the baseline is concentrated in the deeper layers, consistent with late-layer cross-lingual reorganization. Taken together, our findings suggest that the \textit{flip point}---the boundary between incorrect and correct prediction---lies between the native and code-mixed settings in the performance trajectory of low-resource $\rightarrow$ code-mixed $\rightarrow$ English, whether viewed as a property of the input surface form or the model's internal reasoning process. Future work includes expanding coverage to the remaining scheduled Indian languages and additional task formats (reasoning, generation), and connecting TinT's behavioral gains to mechanistic evidence in the model's internal representations.

\section*{Limitations}
\label{sec:limitations}
While \textsc{IndicKLAR} covers 18 of the 22 scheduled Indian languages, the monolingual translations and code-mixed variants are synthetically generated, which may not fully capture the natural variation and dialectal diversity present in real-world multilingual communication. Although both monolingual and code-mixed variants were human-verified for 11 language settings, the monolingual variants of the remaining 7 languages were not subject to manual review. The four scheduled Indian languages not covered by \textsc{IndicKLAR}---Bodo, Kashmiri, Manipuri/Meitei, Santali--- were excluded primarily due to limited translation-system quality and annotator access at evaluation time.

Our evaluation is conducted on a single benchmark derived from \textsc{KLAR-CLC}, which focuses on crosslingual knowledge consistency. The findings may not generalize to other NLP tasks such as reasoning, summarization, or open-ended generation in low-resource Indian language settings.

The prompting strategies explored in this work are evaluated on models with up to 14B parameters, which are at the smaller end of the contemporary scale. It remains an open question whether the same trends hold for larger models or models from different families. Additionally, our inference-time analysis is limited to \texttt{Llama-3.1-8B-Instruct}, and runtime characteristics may differ substantially across hardware configurations and model sizes.

All experiments use a single random seed, and we do not report confidence intervals around the reported accuracy and CLC values; quoted improvements should be interpreted as point estimates.

The code-mixed variants are generated by GPT APIs and may inherit its systematic preferences (e.g., word-class substitution patterns or English-token-ratio distribution). Despite native-speaker verification, our code-mixed distribution is therefore one of many plausible realizations of natural code-mixing rather than a sampled cross-section of it.

The interpretability and qualitative analyses in \S\ref{sec:analysis} are based on a single model (\texttt{Llama-3.1-8B-Instruct}) and, for the qualitative example, a single language (Hindi); mechanistic conclusions should not be inferred from these illustrations alone.

Finally, the code-mixed variants were generated for only 11 of the 18 Indian languages, constrained by available linguist coverage. Languages such as Dogri, Kannada, Konkani, Maithili, Nepali, Sindhi, and Urdu are represented only in their monolingual form, which limits the scope of code-mixed analysis for these languages.

While \textsc{IndicKLAR} focuses on text-based factual knowledge, cross-lingual consistency challenges extend to other modalities. Recent work on vision-language models \cite{gautam-etal-2025-mind} reveals similar high-resource versus low-resource performance gaps, suggesting these issues may be fundamental across modalities. Future work can investigate whether our findings generalize to visual and speech content in Indian languages. 
\section*{Ethical consideration}

AI-assisted writing tools were used to check grammar and improve language quality during manuscript preparation. We thank the native speakers who voluntarily contributed to the human quality verification of \textsc{IndicKLAR} for their careful review of the native and code-mixed variants across all 11 language settings. 

\paragraph{Annotator Demographics.}
\label{para:human_veri_demographics}
Human quality verification was conducted on a voluntary basis by 10 graduate students with native or near-native proficiency in the respective language, either as their mother tongue or through demonstrated linguistic expertise. Annotators ranged in age from 20--25 years, with a gender distribution of 7 male and 3 female participants, collectively covering 10 Indian states. Each annotator was assigned to their language of proficiency and asked to assess whether the samples appeared fluent and natural in that language. 
Rather than independently reading all 2,619 structurally identical sentences per language, our verification strategy focused on the underlying templates. Since the data are generated from a fixed set of 20 category-specific templates (5 templates $\times$ 20 relations) inherited from \textsc{KLAR-CLC}, annotators evaluated the fluency, naturalness, and semantic correctness of the translated and code-mixed \emph{templates} for each relation. This template-level verification ensures that all instantiations maintain quality across the dataset.
The verification task involved careful examination of the assigned language setting and took approximately 1--2 hours per language to complete.

\bibliography{main}

\clearpage
\appendix

\section{Experimental Settings}
\label{sec:appendix-experimental-settings}

This section summarizes the experimental details reported in the main paper. All experiments were conducted on a single NVIDIA A100 GPU (80GB). We used HuggingFace\footnote{\url{https://huggingface.co/}} model checkpoints and employed the vLLM library~\cite{kwon2023efficient} for efficient batch inference and faster decoding.

\begin{table}[h]
\centering
\setlength{\tabcolsep}{4pt}
\resizebox{\columnwidth}{!}{
\begin{tabular}{lcl}
\toprule
\textbf{Family} & \textbf{Parameters} & \textbf{Checkpoint} \\
\midrule

Llama 3.1 & 8B & \texttt{meta-llama/Llama-3.1-8B-Instruct} \\
Llama 3.2 & 1B & \texttt{meta-llama/Llama-3.2-1B-Instruct} \\
Llama 3.2 & 3B & \texttt{meta-llama/Llama-3.2-3B-Instruct} \\

Qwen 2.5 & 1.5B & \texttt{Qwen/Qwen2.5-1.5B-Instruct} \\
Qwen 2.5 & 7B & \texttt{Qwen/Qwen2.5-7B-Instruct} \\
Qwen 2.5 & 14B & \texttt{Qwen/Qwen2.5-14B-Instruct} \\


Gemma 3 & 1B & \texttt{google/gemma-3-1b-it} \\
Gemma 3 & 4B & \texttt{google/gemma-3-4b-it} \\
Gemma 3 & 12B & \texttt{google/gemma-3-12b-it} \\

\bottomrule
\end{tabular}
}
\caption{List of models used across our experiments.}
\label{tab:appendix-model-details}
\end{table}

\begin{table}[ht]
\centering
\resizebox{\columnwidth}{!}{
\begin{tabular}{|l|l|c|}
\hline
\textbf{Language} & \textbf{ISO code} & \textbf{CM availability} \\ \hline
Assamese & as & \checkmark \\ \hline
Bengali (Bangla) & bn & \checkmark \\ \hline
Dogri & doi &  \\ \hline
Gujarati & gu & \checkmark \\ \hline
Hindi & hi & \checkmark \\ \hline
Kannada & kn &  \\ \hline
Konkani & gom &  \\ \hline
Maithili & mai &  \\ \hline
Malayalam & ml & \checkmark \\ \hline
Marathi & mr & \checkmark \\ \hline
Nepali & ne &  \\ \hline
Odia & or & \checkmark \\ \hline
Punjabi & pa & \checkmark \\ \hline
Sanskrit & sa & \checkmark \\ \hline
Sindhi & sd &  \\ \hline
Tamil & ta & \checkmark \\ \hline
Telugu & te & \checkmark \\ \hline
Urdu & ur &  \\ \hline
\end{tabular}
}
\caption{Languages and their ISO codes. Notation: \checkmark indicates languages with constructed code-mixed (CM) variants, based on availability of linguists.}
\label{tab:languages_iso}
\end{table}

\subsection{Evaluation Metrics}
\label{app:eval_metrics}

We evaluate model performance using two metrics: \textbf{Accuracy} and \textbf{Cross-Lingual Consistency (CLC)}.

\paragraph{Accuracy.}
Accuracy measures the proportion of correctly predicted facts across all evaluation instances. Since generative language models may produce partial continuations, we employ a relaxed prefix-matching \cite{wang-etal-2025-lost-multilinguality}. A prediction is considered correct if the generated output is a non-empty case-insensitive prefix of the ground-truth answer, or vice versa. Formally, given a prediction \(\hat{y}_i\) and ground-truth answer \(y_i\), the correctness function is defined as:

\begin{equation}
\mathrm{Match}(\hat{y}_i, y_i) =
\begin{cases}
1, & \text{if } \hat{y}_i \preceq y_i \ \text{or} \ y_i \preceq \hat{y}_i \\
0, & \text{otherwise}
\end{cases}
\end{equation}

where \(\preceq\) denotes the prefix relation after lowercasing and whitespace normalization. The final accuracy is computed as:

\begin{equation}
\text{Accuracy} =
\frac{1}{N}
\sum_{i=1}^{N}
\mathbb{1}_{\text{prefix}}(\hat{y}_i, y_i)
\end{equation}

where \(N\) denotes the total number of evaluation samples.

\paragraph{Cross-Lingual Consistency (CLC).}
In addition to accuracy, we measure Cross-Lingual Consistency (CLC)\cite{qi-etal-2023-cross} to quantify whether the model retrieves consistent knowledge across different languages.CLC is computed using the overlap ratio between the sets of correctly predicted sample indices across language pairs. Given two languages \(L_a\) and \(L_b\), with corresponding correct prediction sets \(C_a\) and \(C_b\), the consistency score is defined as:

\begin{equation}
\text{CLC}(L_a, L_b) =
\frac{|C_a \cap C_b|}
{|C_a \cup C_b|}
\end{equation}

The final CLC score is obtained by averaging the pairwise consistency scores across all evaluated languages. This metric captures the extent to which multilingual models preserve factual knowledge consistently across linguistic variations.

\paragraph{Layer-wise Rank Analysis}
\label{sec:layerwise-rank-analysis}

Following Logit Lens analysis \citep{logit-lens, wang-etal-2025-lost-multilinguality}, we project intermediate hidden states at each transformer layer into the vocabulary space using the model's unembedding matrix and track the rank of the gold answer token across layers. The rank of a token is defined as the number of vocabulary tokens assigned a higher logit value by the model, where lower rank indicates stronger preference for the correct answer. Since answers may consist of multiple subword tokens, we report the minimum rank among all constituent subword tokens of the gold answer. For visualization stability, we report $\log_{10}(\mathrm{rank}+1)$ averaged across evaluation samples.


\begin{table}[t]
\centering
\resizebox{\columnwidth}{!}{
\begin{tabular}{lccc}
\toprule
\textbf{Variant} & \textbf{\#Inst} & \textbf{Avg Qry Len} & \textbf{Avg Ans Len} \\
\midrule
Native Indian & 2619 & 9.23 & 1.13 \\
Code-Mixed Variants & 2619 & 9.20 & 1.02 \\
\bottomrule
\end{tabular}
}
\caption{Dataset statistics for \textsc{IndicKLAR} (open-source under CC-BY-4.0 license).}
\label{tab:dataset_stats}
\end{table}

\subsection{Inference Time Analysis}
\label{app:inference_time_analysis}
Table~\ref{tab:runtime_comparison} reports the average per-sample end-to-end runtime across prompting strategies for all three models. Across all models, multi-step strategies (2Step-CM, 2Step-EN, 2Step-Translit) and answer-inclusive one-step variants (1Step-CM+Ans, 1Step-EN+Ans) consistently incur higher latency than the baseline. Gemma-3-12B and Qwen2.5-14B show this most clearly, with 2Step-Translit reaching 0.60s and 0.71s respectively---nearly double their baselines (0.36s and 0.34s). Both TinT variants (TinT-CM, TinT-EN) remain the most efficient across all three models: Llama-3.1-8B-Instruct runs at just 0.09s under both TinT settings, while Gemma-3-12B and Qwen2.5-14B reach at most 0.26s and 0.37s---well below their respective multi-step counterparts. These results demonstrate that TinT achieves its reasoning benefits without any additional inference overhead, making it practical for large-scale deployment.








\begin{table}[t]
\centering
\small
\setlength{\tabcolsep}{4pt}
\begin{tabular}{lccc}
\toprule
\textbf{Strategy} &
\textbf{Llama-3.1} &
\textbf{Gemma-3} &
\textbf{Qwen2.5} \\
\midrule
Baseline        & 0.13 & 0.36 & 0.34 \\
Obj-Baseline    & 0.08 & 0.23 & 0.32 \\
2Step-CM        & 0.17 & 0.57 & 0.66 \\
2Step-EN        & 0.12 & 0.51 & 0.57 \\
2Step-Translit  & 0.15 & 0.60 & 0.71 \\
1Step-CM+Ans    & 0.16 & 0.53 & 0.59 \\
1Step-EN+Ans    & 0.12 & 0.44 & 0.57 \\
TinT-CM         & 0.09 & 0.26 & 0.36 \\
TinT-EN         & 0.09 & 0.25 & 0.37 \\
\bottomrule
\end{tabular}
\caption{
Average per-sample end-to-end runtime (in seconds) across different prompting settings for the \texttt{meta-llama/Llama-3.1-8B-Instruct}, \texttt{google/gemma-3-12b-it}, and \texttt{Qwen/Qwen2.5-14B-Instruct} models.
}
\label{tab:runtime_comparison}
\end{table}

\subsection{Performance--Efficiency Tradeoffs}

We further analyze the tradeoffs between prompting performance and inference efficiency by comparing the best-performing variants of 1Step, 2Step, and TinT prompting across all languages and models. For a fair comparison, we first compute the improvement of each prompting strategy over baseline for every language and model pair, and then average these improvements across languages and models. For accuracy, 1Step and 2Step prompting achieve the highest overall gains over baseline, with average improvements of $+0.1902$ and $+0.1899$ respectively, while TinT achieves a competitive improvement of $+0.1812$. For CLC evaluation, TinT achieves the strongest overall improvement ($+0.1492$), outperforming both 1Step ($+0.1416$) and 2Step ($+0.1274$). Across both evaluations, the English-guided variants emerge as the dominant settings most frequently selected as the best-performing variants. Importantly, TinT achieves these competitive gains at substantially lower inference cost. While \texttt{1Step-CM+Ans} and \texttt{2Step-CM} require approximately $0.16$s and $0.17$s per sample respectively, TinT variants require only $0.09$s per sample. Despite this reduced runtime, TinT maintains performance comparable to stronger multi-step prompting methods, showing that TinT prompting can provide an effective balance between multilingual performance and computational efficiency.

Similar inference-time trends were also consistently observed for Gemma-3-12B-it and Qwen2.5-14B-Instruct (Appendix~\ref{app:inference_time_analysis}).

\subsection{Scaling Behavior and TinT Analysis}
\label{app:scaling_tint_analysis}
TinT's effectiveness scales clearly with model size. Smaller models such as \texttt{Llama-3.2-1B} and \texttt{Qwen2.5-1.5B} do not reliably outperform the baseline, particularly for CLC: \texttt{Llama-3.2-1B} yields positive CLC gains under \texttt{TinT-CM} for only $11$ of $18$ languages and under \texttt{TinT-EN} for just one, while \texttt{Qwen2.5-1.5B} improves on $15$. This instability largely disappears at scale—\texttt{Llama-3.1-8B}, \texttt{Qwen2.5-7B}, and \texttt{Qwen2.5-14B} all achieve positive accuracy and CLC gains across every language under both variants. The trend is sharpest within the Qwen family, where the average \texttt{TinT-EN} accuracy gain rises from $0.062$ for \texttt{Qwen2.5-1.5B} to $0.244$ and $0.288$ for the $7$B and $14$B variants.

The two variants also diverge by scale. \texttt{TinT-EN} dominates for most medium and large models, whereas \texttt{Llama-3.2-1B} favors \texttt{TinT-CM} across all $18$ languages on both metrics—suggesting smaller models struggle to exploit implicit English-guided reasoning. Nonetheless, the variants remain close on average: across all models, \texttt{TinT-CM} and \texttt{TinT-EN} yield mean accuracy gains of $0.165$ and $0.178$, and CLC gains of $0.140$ and $0.150$, indicating TinT stays effective even directly in code-mixed settings.

Finally, \texttt{Gemma-3-1B} bucks the small-model trend, improving across all languages and both metrics. This suggests multilingual stability under implicit prompting depends not only on parameter count but also on model family and training characteristics.

\subsection{Statistical Analysis}
\label{app:Statistical_significance}
To assess whether the observed improvements across prompting strategies were statistically significant, we conducted paired McNemar tests over 2,619 shared evaluation samples. Both TinT-CM and TinT-EN significantly outperformed the baseline setting. TinT-CM achieved a significant improvement over the baseline ($\chi^2 = 534.75$, $p < 10^{-10}$), while TinT-EN also significantly improved over the baseline ($\chi^2 = 603.56$, $p < 10^{-10}$). Collectively, these results establish the following performance ordering:
\[
\text{Baseline} < \text{TinT}.
\]

\subsection{Pretraining Regime}
\label{sec:res_pretrain}

\paragraph{Base vs.\ Instruct.}
Table~\ref{tab:pretraining_comparison} compares base and instruction-tuned model variants under TinT-CM and TinT-EN settings. In the majority of cases, base and instruction-tuned variants perform comparably across both model families and all five languages. However, isolated exceptions exist: Llama-3.1-8B under TinT-CM shows a gap of up to 9.2 points on ASM (0.732 vs.\ 0.641), and similar but smaller divergences appear for Qwen2.5-14B on ASM under TinT-CM (0.699 vs.\ 0.669). These occasional gaps do not follow a consistent pattern across languages or prompting settings, suggesting they reflect task-specific variance rather than a systematic effect of instruction tuning. Overall, TinT's effectiveness relies primarily on the model's inherent multilingual representations rather than explicit instruction-following. 
\begin{table}[h]
\centering
\setlength{\tabcolsep}{2pt}
\resizebox{\columnwidth}{!}{%
\begin{tabular}{llccccc}
\toprule
\textbf{Model Family} &
\textbf{Setting} &
\textbf{AS} &
\textbf{BN} &
\textbf{HI} &
\textbf{MR} &
\textbf{TE} \\
\midrule
Llama-3.1-8B
& TinT-CM
& 0.732
& 0.796
& 0.793
& 0.794
& 0.749 \\
Llama-3.1-8B-Inst
& TinT-CM
& 0.641
& 0.722
& 0.764
& 0.751
& 0.675 \\
Llama-3.1-8B
& TinT-EN
& 0.721
& 0.802
& 0.806
& 0.811
& 0.742 \\
Llama-3.1-8B-Inst
& TinT-EN
& 0.656
& 0.753
& 0.786
& 0.781
& 0.697 \\
\midrule
Qwen2.5-14B
& TinT-CM
& 0.699
& 0.844
& 0.835
& 0.838
& 0.724 \\
Qwen2.5-14B-Inst
& TinT-CM
& 0.669
& 0.825
& 0.816
& 0.817
& 0.732 \\
Qwen2.5-14B
& TinT-EN
& 0.703
& 0.858
& 0.842
& 0.844
& 0.738 \\
Qwen2.5-14B-Inst
& TinT-EN
& 0.689
& 0.841
& 0.831
& 0.831
& 0.753 \\
\bottomrule
\end{tabular}
}
\caption{Comparison between base and instruction-tuned models under
Translate-in-Thought prompting settings across five representative Indian
languages.}
\label{tab:pretraining_comparison}
\end{table}

\subsection{Illustrating Cross-Lingual Consistency}
\label{app:ronaldo-example}
We present one qualitative example intended to illustrate rather than quantify the mechanism. When asked ``Which country is Ronaldo a citizen of?'' in Hindi, baseline prompting incorrectly predicts Portugal, ignoring the candidate list and relying instead on memorised associations with Cristiano Ronaldo. Pure transliteration (romanising to Latin script) constrains the model to the candidate set but still predicts incorrectly. In contrast, both \textsc{1Step-CM+Ans} and \textsc{TinT-CM} convert the Hindi word for ``citizen'' while preserving grammatical structure and produce the correct answer. We do not claim this single example is representative of the dominant failure mode; the per-category numbers in \S\ref{sec:res_lang_cat} provide the aggregate picture.

\begin{center}
{\inlinepdfcrop[0.33\columnwidth]{0bp 0bp 0bp 0bp}{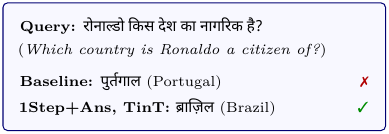}}
 \label{fig:hindi_qualitative_analysis}
\end{center}

\subsection{Entity--Context Ablation}
\label{app:entity_context}

We evaluate two controlled configurations with Llama-3.1-8B-Instruct: (i) English entities with native-language context, and (ii) native-language entities with English context. Table~\ref{tab:entity_context} reports accuracy and CLC across five representative Indian languages. The English-entity configuration consistently performs better than the native-entity configuration, indicating that entity representation contributes substantially to the observed performance differences.

\begin{table*}[t]
\centering
\small
\begin{tabular}{lcccccccccc}
\toprule
Ablation
& \multicolumn{2}{c}{ASM}
& \multicolumn{2}{c}{BEN}
& \multicolumn{2}{c}{HIN}
& \multicolumn{2}{c}{MAR}
& \multicolumn{2}{c}{TEL} \\
& Acc. & CLC
& Acc. & CLC
& Acc. & CLC
& Acc. & CLC
& Acc. & CLC \\
\midrule
Native context, English entity
& 0.533 & 0.614
& 0.680 & 0.692
& 0.589 & 0.632
& 0.649 & 0.691
& 0.635 & 0.669 \\

English context, native entity
& 0.389 & 0.532
& 0.534 & 0.611
& 0.521 & 0.568
& 0.518 & 0.591
& 0.508 & 0.585 \\
\bottomrule
\end{tabular}
\caption{Entity--context ablation using Llama-3.1-8B-Instruct across five representative Indian languages. We independently vary the language of the entities and surrounding context.}
\label{tab:entity_context}
\end{table*}

\begin{table*}[t]
\centering

\setlength{\tabcolsep}{3pt}
\resizebox{0.8\textwidth}{!}{%
\begin{tabular}{llcccccccccc}
\toprule

& 
& \multicolumn{2}{c}{\textbf{AS}} 
& \multicolumn{2}{c}{\textbf{BN}} 
& \multicolumn{2}{c}{\textbf{HI}} 
& \multicolumn{2}{c}{\textbf{MR}} 
& \multicolumn{2}{c}{\textbf{TE}} \\

\cmidrule(lr){3-4}
\cmidrule(lr){5-6}
\cmidrule(lr){7-8}
\cmidrule(lr){9-10}
\cmidrule(lr){11-12}

\textbf{Model} 
& \textbf{Setting}

& \textbf{ACC} & \textbf{CLC} 
& \textbf{ACC} & \textbf{CLC} 
& \textbf{ACC} & \textbf{CLC}
& \textbf{ACC} & \textbf{CLC}
& \textbf{ACC} & \textbf{CLC} \\

\midrule

\multirow{2}{*}{\textbf{LLaMA~3.1~8B}}
& Baseline
& 0.422 & 0.475
& 0.604 & 0.566
& 0.575 & 0.539
& 0.559 & 0.529
&  0.482 & 0.490 \\

& TinT
& 0.735 & 0.766
& 0.833 & 0.825
& 0.843 & 0.806
& 0.830 & 0.820
& 0.784 & 0.783 \\

\midrule

\multirow{2}{*}{\textbf{Gemma~3~12B}}
& Baseline
& 0.580 & 0.652
& 0.784 & 0.781
& 0.796 & 0.779
& 0.803 & 0.786
& 0.797 & 0.776 \\

& TinT
& 0.792 & 0.813
& 0.900 & 0.889
& 0.910 & 0.888
& 0.907 & 0.890
& 0.898 & 0.874 \\

\midrule

\multirow{2}{*}{\textbf{Qwen~2.5~14B}}
& Baseline
& 0.520 & 0.575
&  0.706 & 0.686
& 0.695 & 0.671
& 0.729 & 0.682
& 0.565 & 0.589 \\

& TinT
& 0.708 & 0.728
& 0.854 & 0.816
& 0.859 & 0.811
& 0.844 & 0.813
& 0.756 & 0.757\\

\bottomrule
\end{tabular}%
}
\caption{Comparison between Baseline and TinT settings across Assamese (\texttt{as}), Bengali (\texttt{bn}), Hindi (\texttt{hi}), Marathi (\texttt{mr}), and Telugu (\texttt{te}). Accuracy (ACC) and Cross-Lingual Consistency (CLC) are reported for models.}
\label{tab:tint_vs_baseline_newdata}
\end{table*}

\subsection{Tokenization--Accuracy Correlation}
\label{app:tokenization}

We examine the relationship between average prompt tokenization length and language-wise native-script accuracy across the 18 Indic languages for each of the three main models. Table~\ref{tab:tokenization-correlation} reports the Pearson and Spearman correlations. All three models show a negative Pearson correlation, indicating that languages with longer tokenization tend to exhibit lower native-script accuracy. The corresponding Spearman correlations show the same direction, although the association is not statistically significant for Qwen2.5-14B-Instruct.

\begin{table}[t]
\centering
\small
\begin{tabular}{lcc}
\toprule
Model & Pearson $r$ & Spearman $\rho$ \\
\midrule
Llama-3.1-8B-Instruct
& $-0.717^{***}$
& $-0.629^{**}$ \\

Gemma-3-12B-it
& $-0.693^{**}$
& $-0.542^{*}$ \\

Qwen2.5-14B-Instruct
& $-0.570^{*}$
& $-0.356$ \\
\bottomrule
\end{tabular}
\caption{
Correlation between per-language average prompt tokenization length and language-wise native-script accuracy across 18 Indic languages. Pearson and Spearman correlations are two-tailed. $^{*}p<0.05$, $^{**}p<0.01$, $^{***}p<0.001$.
}
\label{tab:tokenization-correlation}
\end{table}

\subsection{Regional Language Model Evaluation}
\label{app:sarvam}

To assess whether the observed code-mixing gains extend beyond general-purpose multilingual models, we additionally evaluate Sarvam-1\footnote{\url{https://huggingface.co/sarvamai/sarvam-1}}, an Indic-focused language model. Across the nine evaluated Indic languages, code-mixed inputs consistently approach the English accuracy level. For example, Hindi improves from 0.439 in the native setting to 0.729 under code-mixing, compared with 0.737 in English, while Odia improves from 0.450 to 0.736, compared with 0.737 in English. These results suggest that the observed code-mixing effect is not specific to the general-purpose model families used in our main evaluation.%

\begin{table}[t]
\centering
\small
\begin{tabular}{lcc}
\toprule
Language & Native & Code-Mixed \\
\midrule
Bengali   & 0.575 & 0.727 \\
Gujarati  & 0.612 & 0.738 \\
Hindi     & 0.439 & 0.729 \\
Malayalam & 0.570 & 0.734 \\
Marathi   & 0.637 & 0.725 \\
Odia     & 0.450 & 0.736 \\
Punjabi   & 0.596 & 0.729 \\
Tamil     & 0.516 & 0.731 \\
Telugu    & 0.596 & 0.733 \\
\midrule
English   & \multicolumn{2}{c}{0.737} \\
\bottomrule
\end{tabular}
\caption{%
Accuracy of Sarvam-1 across native and code-mixed Indic inputs. 
The English score is language-agnostic and corresponds to the pooled 
English-prompt evaluation.
}
\label{tab:sarvam}
\end{table}

\section{Generalization to Contextually Mediated Knowledge Recall}
\label{app:Generalization_Knowledge_Recall}
To further validate our findings, we evaluate TinT on the contextually mediated knowledge recall benchmark of \citet{liu2026evaluating}, translated into our five representative Indian languages. The dataset statistics are presented in Table \ref{tab:contextual_dataset_stats}.
\begin{table}[h]
\centering
\resizebox{\columnwidth}{!}{
\begin{tabular}{lccc}
\toprule
\textbf{Variant} & \textbf{\#Inst} & \textbf{Avg Qry Len} & \textbf{Avg Ans Len} \\
\midrule
\citet{liu2026evaluating} & 1742 & 22.00 & 1.02 \\
\bottomrule
\end{tabular}
}
\caption{Dataset statistics for the contextually mediated knowledge recall benchmark of \citet{liu2026evaluating}.}
\label{tab:contextual_dataset_stats}
\end{table}
 This benchmark differs from \textsc{IndicKLAR} in that knowledge is accessed through naturalistic referential context rather than direct entity queries, making it a stronger test of crosslingual reasoning. Sample lengths are also considerably longer than those in KLAR-CLC. As shown in Table~\ref{tab:tint_vs_baseline_newdata}, TinT consistently improves over the baseline across all five languages and all three model families, demonstrating that the gains from Translate-in-Thought prompting generalize beyond direct knowledge recall to more contextually grounded settings.

\section{Language-wise Accuracy and CLC Scores on \textsc{IndicKLAR}}
\label{app:acc_clc_main_exp_results}

We report the complete language-wise Accuracy and CLC scores for all evaluated models and prompting strategies on \textsc{IndicKLAR}. The corresponding aggregate results are summarized in the main paper. Tables~\ref{tab:accuracy_18_languages} and~\ref{tab:clc_18_language} provide the corresponding language-wise results.

{
\onecolumn
\scriptsize{
\setlength{\tabcolsep}{4pt}
\renewcommand{\arraystretch}{1.1}
\definecolor{baselinerow}{gray}{0.90}

\begin{longtable}{l l *{2}{c} c *{3}{c} *{3}{c}}
\label{tab:accuracy_18_languages} \\

\toprule
\textbf{Lang} & \textbf{Strategy} & \textbf{Llama 3.1} & \multicolumn{2}{c}{\textbf{Llama 3.2}} & \multicolumn{3}{c}{\textbf{Gemma 3}} & \multicolumn{3}{c}{\textbf{Qwen 2.5}} \\

\cmidrule(lr){3-3}
\cmidrule(lr){4-5}
\cmidrule(lr){6-8}
\cmidrule(lr){9-11}

 &  & \textbf{8B} & \textbf{1B} & \textbf{3B} & \textbf{1B} & \textbf{4B} & \textbf{12B} & \textbf{1.5B} & \textbf{7B} & \textbf{14B} \\

\midrule

English
 & Baseline & 0.842 & 0.744 & 0.805 & 0.672 & 0.763 & 0.870 & 0.833 & 0.879 & 0.840 \\
\midrule

\multirow{7}{*}{Assamese}
 & Baseline & 0.475 & 0.122 & 0.206 & 0.110 & 0.382 & 0.560 & 0.196 & 0.323 & 0.409 \\

 & 2Step-CM & +0.195 & +0.106 & +0.255 & +0.204 & +0.157 & +0.131 & +0.064 & +0.219 & +0.207 \\

 & 2Step-EN & +0.223 & -0.040 & +0.282 & +0.218 & +0.169 & +0.207 & +0.056 & +0.244 & +0.287 \\
\addlinespace[2pt]

 & 1Step-CM+Ans & +0.231 & +0.035 & +0.218 & +0.176 & +0.177 & +0.231 & +0.103 & +0.227 & +0.278 \\

 & 1Step-EN+Ans & +0.278 & -0.038 & +0.223 & +0.213 & +0.220 & +0.260 & +0.082 & +0.251 & +0.310 \\
\addlinespace[2pt]

 & TinT-CM & +0.166 & +0.123 & +0.211 & +0.210 & +0.173 & +0.204 & +0.032 & +0.235 & +0.260 \\

 & TinT-EN & +0.181 & -0.004 & +0.248 & +0.241 & +0.183 & +0.221 & +0.043 & +0.250 & +0.281 \\

\midrule


\multirow{7}{*}{Bengali}
 & Baseline & 0.584 & 0.196 & 0.403 & 0.287 & 0.525 & 0.741 & 0.234 & 0.440 & 0.554 \\

 & 2Step-CM & +0.194 & +0.056 & +0.207 & +0.126 & +0.178 & +0.118 & +0.120 & +0.240 & +0.217 \\

 & 2Step-EN & +0.239 & -0.023 & +0.263 & +0.161 & +0.245 & +0.158 & +0.124 & +0.310 & +0.302 \\
\addlinespace[2pt]

 & 1Step-CM+Ans & +0.192 & -0.047 & +0.172 & +0.086 & +0.206 & +0.149 & +0.124 & +0.261 & +0.279 \\

 & 1Step-EN+Ans & +0.243 & -0.118 & +0.134 & +0.193 & +0.252 & +0.159 & +0.087 & +0.288 & +0.308 \\
\addlinespace[2pt]

 & TinT-CM & +0.138 & +0.023 & +0.123 & +0.207 & +0.195 & +0.144 & +0.063 & +0.262 & +0.271 \\

 & TinT-EN & +0.169 & -0.074 & +0.158 & +0.224 & +0.200 & +0.149 & +0.074 & +0.276 & +0.288 \\

\midrule


\multirow{7}{*}{Dogri}
 & Baseline & 0.561 & 0.238 & 0.426 & 0.246 & 0.486 & 0.716 & 0.230 & 0.489 & 0.505 \\

 & 2Step-CM & +0.211 & +0.157 & +0.155 & +0.156 & +0.184 & +0.096 & +0.145 & +0.199 & +0.276 \\

 & 2Step-EN & +0.254 & +0.066 & +0.182 & +0.128 & +0.176 & +0.132 & +0.167 & +0.222 & +0.330 \\
\addlinespace[2pt]

 & 1Step-CM+Ans & +0.218 & +0.017 & +0.163 & +0.077 & +0.196 & +0.147 & +0.218 & +0.222 & +0.321 \\

 & 1Step-EN+Ans & +0.268 & -0.026 & +0.196 & +0.154 & +0.234 & +0.163 & +0.134 & +0.229 & +0.350 \\
\addlinespace[2pt]

 & TinT-CM & +0.198 & +0.175 & +0.132 & +0.171 & +0.179 & +0.143 & +0.126 & +0.213 & +0.308 \\

 & TinT-EN & +0.221 & +0.070 & +0.152 & +0.204 & +0.207 & +0.145 & +0.134 & +0.222 & +0.326 \\

\midrule


\multirow{7}{*}{Gujarati}
 & Baseline & 0.476 & 0.198 & 0.363 & 0.188 & 0.477 & 0.721 & 0.241 & 0.292 & 0.423 \\

 & 2Step-CM & +0.254 & +0.024 & +0.177 & +0.163 & +0.207 & +0.108 & +0.023 & +0.308 & +0.341 \\

 & 2Step-EN & +0.304 & -0.115 & +0.226 & +0.176 & +0.220 & +0.160 & +0.031 & +0.370 & +0.384 \\
\addlinespace[2pt]

 & 1Step-CM+Ans & +0.257 & -0.066 & +0.106 & +0.148 & +0.163 & +0.158 & +0.055 & +0.328 & +0.351 \\

 & 1Step-EN+Ans & +0.300 & -0.121 & +0.130 & +0.224 & +0.226 & +0.167 & +0.034 & +0.324 & +0.393 \\
\addlinespace[2pt]

 & TinT-CM & +0.220 & -0.021 & +0.118 & +0.180 & +0.178 & +0.146 & +0.015 & +0.324 & +0.339 \\

 & TinT-EN & +0.247 & -0.117 & +0.143 & +0.217 & +0.175 & +0.151 & +0.015 & +0.331 & +0.361 \\

\midrule


\multirow{7}{*}{Hindi}
 & Baseline & 0.592 & 0.258 & 0.386 & 0.236 & 0.512 & 0.719 & 0.286 & 0.470 & 0.523 \\

 & 2Step-CM & +0.174 & +0.008 & +0.256 & +0.167 & +0.207 & +0.122 & +0.119 & +0.210 & +0.273 \\

 & 2Step-EN & +0.210 & +0.039 & +0.315 & +0.229 & +0.233 & +0.151 & +0.136 & +0.278 & +0.317 \\
\addlinespace[2pt]

 & 1Step-CM+Ans & +0.211 & -0.019 & +0.248 & +0.138 & +0.193 & +0.155 & +0.079 & +0.254 & +0.295 \\

 & 1Step-EN+Ans & +0.233 & -0.081 & +0.255 & +0.212 & +0.243 & +0.154 & +0.074 & +0.263 & +0.326 \\
\addlinespace[2pt]

 & TinT-CM & +0.172 & +0.060 & +0.228 & +0.170 & +0.201 & +0.153 & +0.064 & +0.239 & +0.292 \\

 & TinT-EN & +0.194 & -0.036 & +0.261 & +0.205 & +0.200 & +0.148 & +0.072 & +0.257 & +0.307 \\

\midrule


\multirow{7}{*}{Kannada}
 & Baseline & 0.483 & 0.197 & 0.364 & 0.110 & 0.443 & 0.717 & 0.186 & 0.388 & 0.442 \\

 & 2Step-CM & +0.276 & +0.114 & +0.168 & +0.152 & +0.194 & +0.122 & +0.008 & +0.151 & +0.250 \\

 & 2Step-EN & +0.304 & -0.076 & +0.222 & +0.169 & +0.235 & +0.157 & +0.004 & +0.219 & +0.332 \\
\addlinespace[2pt]

 & 1Step-CM+Ans & +0.220 & -0.023 & +0.140 & +0.172 & +0.157 & +0.150 & +0.117 & +0.169 & +0.294 \\

 & 1Step-EN+Ans & +0.286 & -0.085 & +0.107 & +0.204 & +0.226 & +0.169 & +0.055 & +0.182 & +0.322 \\
\addlinespace[2pt]

 & TinT-CM & +0.187 & +0.070 & +0.064 & +0.227 & +0.171 & +0.144 & +0.039 & +0.153 & +0.259 \\

 & TinT-EN & +0.213 & -0.049 & +0.132 & +0.271 & +0.167 & +0.153 & +0.031 & +0.170 & +0.275 \\

\midrule


\multirow{7}{*}{Konkani}
 & Baseline & 0.451 & 0.196 & 0.320 & 0.158 & 0.446 & 0.664 & 0.152 & 0.265 & 0.426 \\

 & 2Step-CM & +0.288 & +0.213 & +0.234 & +0.249 & +0.201 & +0.163 & +0.076 & +0.343 & +0.331 \\

 & 2Step-EN & +0.359 & +0.069 & +0.277 & +0.263 & +0.221 & +0.179 & +0.064 & +0.395 & +0.341 \\
\addlinespace[2pt]

 & 1Step-CM+Ans & +0.315 & -0.019 & +0.187 & +0.245 & +0.175 & +0.202 & +0.068 & +0.388 & +0.378 \\

 & 1Step-EN+Ans & +0.362 & -0.033 & +0.217 & +0.311 & +0.219 & +0.211 & +0.011 & +0.410 & +0.394 \\
\addlinespace[2pt]

 & TinT-CM & +0.285 & +0.230 & +0.199 & +0.255 & +0.183 & +0.204 & +0.101 & +0.374 & +0.362 \\

 & TinT-EN & +0.300 & +0.110 & +0.243 & +0.287 & +0.208 & +0.212 & +0.065 & +0.393 & +0.379 \\

\midrule


\multirow{7}{*}{Maithili}
 & Baseline & 0.582 & 0.316 & 0.457 & 0.265 & 0.551 & 0.738 & 0.265 & 0.413 & 0.502 \\

 & 2Step-CM & +0.197 & +0.007 & +0.168 & +0.146 & +0.171 & +0.108 & +0.149 & +0.257 & +0.279 \\

 & 2Step-EN & +0.222 & -0.130 & +0.198 & +0.140 & +0.161 & +0.130 & +0.174 & +0.305 & +0.323 \\
\addlinespace[2pt]

 & 1Step-CM+Ans & +0.229 & -0.027 & +0.159 & +0.072 & +0.159 & +0.131 & +0.118 & +0.304 & +0.322 \\

 & 1Step-EN+Ans & +0.266 & -0.084 & +0.160 & +0.167 & +0.192 & +0.140 & +0.096 & +0.318 & +0.347 \\
\addlinespace[2pt]

 & TinT-CM & +0.191 & +0.053 & +0.144 & +0.166 & +0.164 & +0.140 & +0.083 & +0.296 & +0.316 \\

 & TinT-EN & +0.207 & -0.058 & +0.171 & +0.211 & +0.168 & +0.137 & +0.107 & +0.309 & +0.339 \\

\midrule


\multirow{7}{*}{Malayalam}
 & Baseline & 0.528 & 0.153 & 0.354 & 0.049 & 0.480 & 0.721 & 0.270 & 0.400 & 0.516 \\

 & 2Step-CM & +0.211 & +0.056 & +0.175 & +0.100 & +0.158 & +0.092 & -0.029 & +0.164 & +0.159 \\

 & 2Step-EN & +0.226 & -0.054 & +0.197 & +0.112 & +0.185 & +0.151 & -0.021 & +0.236 & +0.233 \\
\addlinespace[2pt]

 & 1Step-CM+Ans & +0.161 & +0.029 & +0.175 & +0.100 & +0.117 & +0.132 & +0.060 & +0.158 & +0.207 \\

 & 1Step-EN+Ans & +0.240 & -0.002 & +0.152 & +0.171 & +0.163 & +0.146 & +0.047 & +0.172 & +0.252 \\
\addlinespace[2pt]

 & TinT-CM & +0.155 & +0.082 & +0.127 & +0.144 & +0.124 & +0.134 & +0.007 & +0.148 & +0.200 \\

 & TinT-EN & +0.184 & +0.006 & +0.165 & +0.194 & +0.141 & +0.140 & +0.034 & +0.159 & +0.210 \\

\midrule


\multirow{7}{*}{Marathi}
 & Baseline & 0.592 & 0.242 & 0.399 & 0.303 & 0.537 & 0.764 & 0.310 & 0.443 & 0.579 \\

 & 2Step-CM & +0.190 & +0.020 & +0.221 & +0.095 & +0.170 & +0.081 & +0.050 & +0.224 & +0.201 \\

 & 2Step-EN & +0.226 & +0.049 & +0.259 & +0.152 & +0.190 & +0.126 & +0.059 & +0.293 & +0.258 \\
\addlinespace[2pt]

 & 1Step-CM+Ans & +0.196 & -0.005 & +0.213 & +0.064 & +0.169 & +0.124 & +0.064 & +0.263 & +0.249 \\

 & 1Step-EN+Ans & +0.234 & -0.035 & +0.224 & +0.126 & +0.221 & +0.133 & -0.032 & +0.276 & +0.279 \\
\addlinespace[2pt]

 & TinT-CM & +0.159 & +0.065 & +0.173 & +0.118 & +0.167 & +0.113 & +0.003 & +0.252 & +0.238 \\

 & TinT-EN & +0.189 & -0.023 & +0.211 & +0.166 & +0.175 & +0.120 & +0.001 & +0.272 & +0.252 \\

\midrule


\multirow{7}{*}{Nepali}
 & Baseline & 0.556 & 0.227 & 0.352 & 0.156 & 0.518 & 0.714 & 0.289 & 0.439 & 0.364 \\

 & 2Step-CM & +0.190 & +0.064 & +0.250 & +0.268 & +0.166 & +0.109 & +0.092 & +0.239 & +0.396 \\

 & 2Step-EN & +0.191 & -0.034 & +0.197 & +0.222 & +0.193 & +0.150 & +0.133 & +0.283 & +0.464 \\
\addlinespace[2pt]

 & 1Step-CM+Ans & +0.248 & +0.014 & +0.272 & +0.150 & +0.163 & +0.153 & +0.127 & +0.277 & +0.457 \\

 & 1Step-EN+Ans & +0.265 & -0.002 & +0.233 & +0.207 & +0.219 & +0.155 & +0.107 & +0.290 & +0.480 \\
\addlinespace[2pt]

 & TinT-CM & +0.206 & +0.113 & +0.211 & +0.290 & +0.163 & +0.145 & +0.054 & +0.271 & +0.443 \\

 & TinT-EN & +0.225 & -0.006 & +0.263 & +0.259 & +0.156 & +0.153 & +0.081 & +0.282 & +0.462 \\

\midrule


\multirow{7}{*}{Odia}
 & Baseline & 0.350 & 0.071 & 0.142 & 0.077 & 0.311 & 0.516 & 0.146 & 0.222 & 0.301 \\

 & 2Step-CM & +0.248 & +0.051 & +0.042 & +0.178 & +0.170 & +0.184 & -0.002 & +0.214 & +0.324 \\

 & 2Step-EN & +0.283 & +0.052 & +0.052 & +0.167 & +0.168 & +0.275 & -0.024 & +0.250 & +0.378 \\
\addlinespace[2pt]

 & 1Step-CM+Ans & +0.198 & +0.077 & +0.085 & +0.116 & +0.157 & +0.245 & -0.002 & +0.058 & +0.305 \\

 & 1Step-EN+Ans & +0.305 & +0.018 & +0.119 & +0.209 & +0.176 & +0.283 & +0.069 & +0.128 & +0.354 \\
\addlinespace[2pt]

 & TinT-CM & +0.188 & +0.207 & +0.037 & +0.184 & +0.178 & +0.215 & +0.009 & +0.210 & +0.291 \\

 & TinT-EN & +0.208 & +0.164 & +0.070 & +0.172 & +0.169 & +0.243 & +0.014 & +0.251 & +0.317 \\

\midrule


\multirow{7}{*}{Punjabi}
 & Baseline & 0.476 & 0.187 & 0.344 & 0.118 & 0.417 & 0.654 & 0.206 & 0.339 & 0.494 \\

 & 2Step-CM & +0.224 & +0.033 & +0.151 & +0.051 & +0.202 & +0.098 & +0.035 & +0.172 & +0.208 \\

 & 2Step-EN & +0.280 & -0.090 & +0.218 & +0.055 & +0.249 & +0.154 & +0.036 & +0.253 & +0.274 \\
\addlinespace[2pt]

 & 1Step-CM+Ans & +0.224 & -0.049 & +0.174 & +0.072 & +0.152 & +0.168 & +0.085 & +0.234 & +0.236 \\

 & 1Step-EN+Ans & +0.291 & -0.143 & +0.147 & +0.115 & +0.181 & +0.163 & +0.112 & +0.262 & +0.297 \\
\addlinespace[2pt]

 & TinT-CM & +0.197 & +0.005 & +0.129 & +0.081 & +0.161 & +0.149 & +0.032 & +0.222 & +0.236 \\

 & TinT-EN & +0.223 & -0.076 & +0.144 & +0.105 & +0.167 & +0.155 & +0.050 & +0.239 & +0.265 \\

\midrule


\multirow{7}{*}{Sanskrit}
 & Baseline & 0.509 & 0.225 & 0.312 & 0.190 & 0.470 & 0.695 & 0.215 & 0.377 & 0.517 \\

 & 2Step-CM & +0.236 & +0.120 & +0.229 & +0.202 & +0.165 & +0.125 & +0.128 & +0.272 & +0.255 \\

 & 2Step-EN & +0.282 & +0.005 & +0.264 & +0.189 & +0.151 & +0.158 & +0.157 & +0.320 & +0.296 \\
\addlinespace[2pt]

 & 1Step-CM+Ans & +0.256 & -0.004 & +0.177 & +0.129 & +0.171 & +0.159 & +0.112 & +0.328 & +0.312 \\

 & 1Step-EN+Ans & +0.311 & -0.023 & +0.281 & +0.191 & +0.240 & +0.176 & +0.065 & +0.340 & +0.334 \\
\addlinespace[2pt]

 & TinT-CM & +0.226 & +0.147 & +0.163 & +0.274 & +0.171 & +0.154 & +0.076 & +0.307 & +0.288 \\

 & TinT-EN & +0.261 & +0.033 & +0.195 & +0.271 & +0.190 & +0.158 & +0.076 & +0.323 & +0.313 \\

\midrule


\multirow{7}{*}{Sindhi}
 & Baseline & 0.478 & 0.172 & 0.249 & 0.129 & 0.416 & 0.540 & 0.188 & 0.369 & 0.509 \\

 & 2Step-CM & +0.121 & +0.069 & +0.079 & +0.181 & +0.119 & +0.110 & +0.101 & +0.121 & +0.074 \\

 & 2Step-EN & +0.192 & -0.055 & +0.105 & +0.152 & +0.145 & +0.208 & +0.045 & +0.081 & +0.073 \\
\addlinespace[2pt]

 & 1Step-CM+Ans & +0.193 & -0.007 & +0.156 & +0.142 & +0.123 & +0.207 & +0.143 & +0.183 & +0.128 \\

 & 1Step-EN+Ans & +0.247 & -0.066 & +0.150 & +0.233 & +0.175 & +0.249 & +0.164 & +0.198 & +0.153 \\
\addlinespace[2pt]

 & TinT-CM & +0.153 & +0.112 & +0.117 & +0.205 & +0.141 & +0.188 & +0.127 & +0.178 & +0.114 \\

 & TinT-EN & +0.187 & +0.008 & +0.140 & +0.252 & +0.153 & +0.204 & +0.105 & +0.189 & +0.148 \\

\midrule


\multirow{7}{*}{Tamil}
 & Baseline & 0.440 & 0.181 & 0.329 & 0.285 & 0.488 & 0.700 & 0.200 & 0.303 & 0.392 \\

 & 2Step-CM & +0.231 & -0.001 & +0.166 & +0.139 & +0.190 & +0.067 & -0.018 & +0.165 & +0.219 \\

 & 2Step-EN & +0.271 & -0.049 & +0.219 & +0.149 & +0.226 & +0.165 & -0.024 & +0.194 & +0.277 \\
\addlinespace[2pt]

 & 1Step-CM+Ans & +0.231 & +0.004 & +0.199 & +0.089 & +0.185 & +0.164 & +0.054 & +0.172 & +0.276 \\

 & 1Step-EN+Ans & +0.289 & -0.080 & +0.148 & +0.186 & +0.250 & +0.175 & +0.089 & +0.207 & +0.297 \\
\addlinespace[2pt]

 & TinT-CM & +0.194 & +0.027 & +0.084 & +0.146 & +0.187 & +0.141 & +0.010 & +0.179 & +0.255 \\

 & TinT-EN & +0.218 & -0.082 & +0.134 & +0.194 & +0.181 & +0.142 & +0.010 & +0.198 & +0.266 \\

\midrule


\multirow{7}{*}{Telugu}
 & Baseline & 0.509 & 0.181 & 0.385 & 0.192 & 0.499 & 0.745 & 0.189 & 0.473 & 0.515 \\

 & 2Step-CM & +0.216 & +0.100 & +0.189 & +0.190 & +0.209 & +0.108 & +0.011 & +0.088 & +0.214 \\

 & 2Step-EN & +0.279 & -0.012 & +0.225 & +0.179 & +0.233 & +0.151 & +0.011 & +0.139 & +0.279 \\
\addlinespace[2pt]

 & 1Step-CM+Ans & +0.182 & +0.019 & +0.151 & +0.106 & +0.195 & +0.144 & +0.093 & +0.089 & +0.242 \\

 & 1Step-EN+Ans & +0.261 & +0.018 & +0.134 & +0.234 & +0.228 & +0.153 & +0.126 & +0.107 & +0.276 \\
\addlinespace[2pt]

 & TinT-CM & +0.166 & +0.084 & +0.107 & +0.198 & +0.168 & +0.128 & +0.020 & +0.074 & +0.217 \\

 & TinT-EN & +0.188 & -0.013 & +0.158 & +0.286 & +0.192 & +0.132 & +0.070 & +0.096 & +0.238 \\

\midrule


\multirow{7}{*}{Urdu}
 & Baseline & 0.570 & 0.265 & 0.407 & 0.137 & 0.483 & 0.688 & 0.251 & 0.485 & 0.590 \\

 & 2Step-CM & +0.139 & -0.058 & +0.138 & +0.124 & +0.152 & +0.081 & +0.116 & +0.096 & +0.092 \\

 & 2Step-EN & +0.204 & -0.092 & +0.239 & +0.134 & +0.195 & +0.155 & +0.107 & +0.172 & +0.180 \\
\addlinespace[2pt]

 & 1Step-CM+Ans & +0.190 & -0.078 & +0.163 & +0.093 & +0.132 & +0.138 & +0.126 & +0.158 & +0.174 \\

 & 1Step-EN+Ans & +0.233 & -0.140 & +0.151 & +0.172 & +0.203 & +0.147 & +0.140 & +0.189 & +0.189 \\
\addlinespace[2pt]

 & TinT-CM & +0.137 & -0.019 & +0.112 & +0.168 & +0.125 & +0.132 & +0.120 & +0.153 & +0.149 \\

 & TinT-EN & +0.181 & -0.100 & +0.166 & +0.205 & +0.145 & +0.133 & +0.128 & +0.170 & +0.160 \\

\midrule

\bottomrule

\caption{Consolidated multilingual evaluation accuracy results across all model families and prompting strategies. Scores are reported as accuracy in the range $[0,1]$.}

\end{longtable}
}

\twocolumn
}

{
\onecolumn

\scriptsize{
\setlength{\tabcolsep}{4pt}
\renewcommand{\arraystretch}{1.1}
\definecolor{baselinerow}{gray}{0.90}

\begin{longtable}{l l *{2}{c} c *{3}{c} *{3}{c}}
\label{tab:clc_18_language} \\

\toprule
\textbf{Lang} & \textbf{Strategy} & \textbf{Llama 3.1} & \multicolumn{2}{c}{\textbf{Llama 3.2}} & \multicolumn{3}{c}{\textbf{Gemma 3}} & \multicolumn{3}{c}{\textbf{Qwen 2.5}} \\

\cmidrule(lr){3-3}
\cmidrule(lr){4-5}
\cmidrule(lr){6-8}
\cmidrule(lr){9-11}

 &  & \textbf{8B} & \textbf{1B} & \textbf{3B} & \textbf{1B} & \textbf{4B} & \textbf{12B} & \textbf{1.5B} & \textbf{7B} & \textbf{14B} \\

\midrule


\multirow{7}{*}{Assamese}
 & Baseline & 0.518 & 0.263 & 0.322 & 0.197 & 0.463 & 0.637 & 0.362 & 0.408 & 0.480 \\

 & 2Step-CM & +0.170 & -0.000 & +0.190 & +0.095 & +0.113 & +0.063 & -0.024 & +0.143 & +0.142 \\

 & 2Step-EN & +0.184 & -0.133 & +0.189 & +0.094 & +0.118 & +0.135 & -0.025 & +0.167 & +0.209 \\
\addlinespace[2pt]

 & 1Step-CM+Ans & +0.202 & -0.003 & +0.158 & +0.112 & +0.156 & +0.174 & -0.057 & +0.172 & +0.218 \\

 & 1Step-EN+Ans & +0.254 & -0.113 & +0.163 & +0.140 & +0.167 & +0.201 & -0.074 & +0.195 & +0.248 \\
\addlinespace[2pt]

 & TinT-CM & +0.191 & +0.032 & +0.186 & +0.135 & +0.140 & +0.153 & -0.020 & +0.192 & +0.212 \\

 & TinT-EN & +0.204 & -0.070 & +0.210 & +0.153 & +0.164 & +0.165 & -0.012 & +0.205 & +0.230 \\

\midrule


\multirow{7}{*}{Bengali}
 & Baseline & 0.590 & 0.338 & 0.456 & 0.277 & 0.535 & 0.740 & 0.351 & 0.461 & 0.548 \\

 & 2Step-CM & +0.152 & -0.042 & +0.129 & +0.084 & +0.140 & +0.064 & +0.040 & +0.149 & +0.168 \\

 & 2Step-EN & +0.194 & -0.142 & +0.151 & +0.078 & +0.166 & +0.124 & +0.029 & +0.207 & +0.238 \\
\addlinespace[2pt]

 & 1Step-CM+Ans & +0.169 & -0.079 & +0.106 & +0.085 & +0.162 & +0.133 & -0.006 & +0.196 & +0.247 \\

 & 1Step-EN+Ans & +0.224 & -0.190 & +0.097 & +0.130 & +0.190 & +0.150 & -0.043 & +0.215 & +0.268 \\
\addlinespace[2pt]

 & TinT-CM & +0.166 & -0.041 & +0.127 & +0.131 & +0.158 & +0.123 & +0.039 & +0.202 & +0.236 \\

 & TinT-EN & +0.186 & -0.120 & +0.149 & +0.160 & +0.174 & +0.132 & +0.052 & +0.215 & +0.247 \\

\midrule


\multirow{7}{*}{Dogri}
 & Baseline & 0.572 & 0.323 & 0.454 & 0.268 & 0.516 & 0.747 & 0.357 & 0.449 & 0.534 \\

 & 2Step-CM & +0.182 & +0.014 & +0.140 & +0.116 & +0.159 & +0.046 & +0.049 & +0.175 & +0.202 \\

 & 2Step-EN & +0.217 & -0.106 & +0.136 & +0.078 & +0.157 & +0.092 & +0.038 & +0.228 & +0.252 \\
\addlinespace[2pt]

 & 1Step-CM+Ans & +0.188 & -0.025 & +0.121 & +0.113 & +0.174 & +0.118 & +0.028 & +0.214 & +0.264 \\

 & 1Step-EN+Ans & +0.247 & -0.106 & +0.098 & +0.142 & +0.193 & +0.136 & -0.032 & +0.239 & +0.288 \\
\addlinespace[2pt]

 & TinT-CM & +0.196 & +0.051 & +0.137 & +0.151 & +0.167 & +0.108 & +0.069 & +0.228 & +0.260 \\

 & TinT-EN & +0.219 & -0.037 & +0.147 & +0.172 & +0.187 & +0.111 & +0.086 & +0.244 & +0.271 \\

\midrule


\multirow{7}{*}{Gujarati}
 & Baseline & 0.537 & 0.290 & 0.442 & 0.256 & 0.507 & 0.746 & 0.331 & 0.402 & 0.496 \\

 & 2Step-CM & +0.199 & -0.005 & +0.124 & +0.091 & +0.166 & +0.057 & +0.004 & +0.199 & +0.232 \\

 & 2Step-EN & +0.230 & -0.151 & +0.140 & +0.093 & +0.178 & +0.116 & -0.004 & +0.248 & +0.274 \\
\addlinespace[2pt]

 & 1Step-CM+Ans & +0.199 & -0.044 & +0.078 & +0.102 & +0.169 & +0.124 & -0.020 & +0.230 & +0.270 \\

 & 1Step-EN+Ans & +0.257 & -0.148 & +0.077 & +0.149 & +0.199 & +0.143 & -0.068 & +0.239 & +0.305 \\
\addlinespace[2pt]

 & TinT-CM & +0.209 & -0.031 & +0.124 & +0.126 & +0.174 & +0.117 & +0.010 & +0.246 & +0.266 \\

 & TinT-EN & +0.226 & -0.115 & +0.135 & +0.166 & +0.184 & +0.125 & +0.024 & +0.261 & +0.283 \\

\midrule


\multirow{7}{*}{Hindi}
 & Baseline & 0.569 & 0.315 & 0.422 & 0.263 & 0.530 & 0.743 & 0.323 & 0.438 & 0.515 \\

 & 2Step-CM & +0.176 & +0.001 & +0.164 & +0.119 & +0.158 & +0.060 & +0.088 & +0.175 & +0.217 \\

 & 2Step-EN & +0.210 & -0.101 & +0.190 & +0.121 & +0.168 & +0.113 & +0.089 & +0.239 & +0.273 \\
\addlinespace[2pt]

 & 1Step-CM+Ans & +0.194 & -0.027 & +0.152 & +0.116 & +0.166 & +0.123 & +0.044 & +0.229 & +0.272 \\

 & 1Step-EN+Ans & +0.239 & -0.103 & +0.133 & +0.157 & +0.194 & +0.137 & -0.003 & +0.253 & +0.302 \\
\addlinespace[2pt]

 & TinT-CM & +0.190 & +0.050 & +0.168 & +0.159 & +0.170 & +0.115 & +0.108 & +0.236 & +0.270 \\

 & TinT-EN & +0.210 & -0.038 & +0.180 & +0.174 & +0.176 & +0.121 & +0.126 & +0.255 & +0.284 \\

\midrule


\multirow{7}{*}{Kannada}
 & Baseline & 0.565 & 0.314 & 0.464 & 0.191 & 0.527 & 0.735 & 0.357 & 0.445 & 0.515 \\

 & 2Step-CM & +0.177 & -0.002 & +0.102 & +0.082 & +0.127 & +0.067 & -0.065 & +0.109 & +0.162 \\

 & 2Step-EN & +0.206 & -0.147 & +0.123 & +0.088 & +0.148 & +0.122 & -0.078 & +0.166 & +0.227 \\
\addlinespace[2pt]

 & 1Step-CM+Ans & +0.153 & -0.033 & +0.076 & +0.123 & +0.137 & +0.131 & -0.044 & +0.151 & +0.227 \\

 & 1Step-EN+Ans & +0.228 & -0.161 & +0.050 & +0.150 & +0.167 & +0.151 & -0.088 & +0.172 & +0.250 \\
\addlinespace[2pt]

 & TinT-CM & +0.170 & +0.012 & +0.074 & +0.150 & +0.144 & +0.125 & -0.024 & +0.161 & +0.207 \\

 & TinT-EN & +0.182 & -0.096 & +0.109 & +0.189 & +0.146 & +0.132 & -0.034 & +0.174 & +0.218 \\

\midrule


\multirow{7}{*}{Konkani}
 & Baseline & 0.511 & 0.298 & 0.414 & 0.234 & 0.487 & 0.700 & 0.331 & 0.380 & 0.467 \\

 & 2Step-CM & +0.208 & +0.025 & +0.159 & +0.117 & +0.168 & +0.097 & -0.038 & +0.225 & +0.256 \\

 & 2Step-EN & +0.257 & -0.113 & +0.167 & +0.115 & +0.172 & +0.136 & -0.060 & +0.252 & +0.274 \\
\addlinespace[2pt]

 & 1Step-CM+Ans & +0.236 & -0.039 & +0.128 & +0.117 & +0.179 & +0.157 & -0.083 & +0.265 & +0.311 \\

 & 1Step-EN+Ans & +0.289 & -0.106 & +0.119 & +0.155 & +0.200 & +0.176 & -0.134 & +0.285 & +0.326 \\
\addlinespace[2pt]

 & TinT-CM & +0.244 & +0.057 & +0.161 & +0.147 & +0.175 & +0.153 & +0.031 & +0.270 & +0.305 \\

 & TinT-EN & +0.259 & -0.039 & +0.162 & +0.180 & +0.193 & +0.162 & +0.012 & +0.293 & +0.320 \\

\midrule


\multirow{7}{*}{Maithili}
 & Baseline & 0.603 & 0.353 & 0.485 & 0.286 & 0.547 & 0.753 & 0.382 & 0.458 & 0.529 \\

 & 2Step-CM & +0.158 & +0.006 & +0.115 & +0.114 & +0.153 & +0.064 & +0.048 & +0.166 & +0.214 \\

 & 2Step-EN & +0.185 & -0.161 & +0.120 & +0.086 & +0.151 & +0.102 & +0.028 & +0.225 & +0.263 \\
\addlinespace[2pt]

 & 1Step-CM+Ans & +0.166 & -0.047 & +0.107 & +0.111 & +0.159 & +0.120 & -0.004 & +0.223 & +0.271 \\

 & 1Step-EN+Ans & +0.218 & -0.117 & +0.087 & +0.145 & +0.187 & +0.137 & -0.047 & +0.245 & +0.293 \\
\addlinespace[2pt]

 & TinT-CM & +0.174 & +0.041 & +0.121 & +0.158 & +0.162 & +0.116 & +0.073 & +0.232 & +0.266 \\

 & TinT-EN & +0.190 & -0.050 & +0.110 & +0.185 & +0.177 & +0.120 & +0.084 & +0.247 & +0.280 \\

\midrule


\multirow{7}{*}{Malayalam}
 & Baseline & 0.555 & 0.294 & 0.430 & 0.101 & 0.529 & 0.728 & 0.319 & 0.397 & 0.517 \\

 & 2Step-CM & +0.176 & -0.025 & +0.130 & +0.089 & +0.109 & +0.059 & +0.021 & +0.177 & +0.145 \\

 & 2Step-EN & +0.193 & -0.166 & +0.133 & +0.100 & +0.128 & +0.121 & +0.008 & +0.229 & +0.213 \\
\addlinespace[2pt]

 & 1Step-CM+Ans & +0.160 & -0.024 & +0.112 & +0.117 & +0.117 & +0.121 & +0.005 & +0.197 & +0.206 \\

 & 1Step-EN+Ans & +0.237 & -0.114 & +0.094 & +0.175 & +0.145 & +0.144 & -0.011 & +0.220 & +0.241 \\
\addlinespace[2pt]

 & TinT-CM & +0.183 & +0.005 & +0.127 & +0.154 & +0.123 & +0.124 & +0.044 & +0.208 & +0.208 \\

 & TinT-EN & +0.204 & -0.069 & +0.149 & +0.193 & +0.137 & +0.126 & +0.054 & +0.215 & +0.213 \\

\midrule

\newpage

\multirow{7}{*}{Marathi}
 & Baseline & 0.577 & 0.301 & 0.451 & 0.288 & 0.547 & 0.760 & 0.338 & 0.459 & 0.552 \\

 & 2Step-CM & +0.174 & +0.014 & +0.130 & +0.091 & +0.130 & +0.053 & +0.062 & +0.165 & +0.179 \\

 & 2Step-EN & +0.206 & -0.079 & +0.152 & +0.085 & +0.147 & +0.106 & +0.057 & +0.220 & +0.233 \\
\addlinespace[2pt]

 & 1Step-CM+Ans & +0.189 & -0.007 & +0.128 & +0.105 & +0.156 & +0.118 & +0.040 & +0.212 & +0.241 \\

 & 1Step-EN+Ans & +0.238 & -0.083 & +0.102 & +0.142 & +0.182 & +0.135 & -0.031 & +0.231 & +0.266 \\
\addlinespace[2pt]

 & TinT-CM & +0.189 & +0.046 & +0.144 & +0.140 & +0.152 & +0.107 & +0.094 & +0.221 & +0.235 \\

 & TinT-EN & +0.205 & -0.024 & +0.157 & +0.165 & +0.167 & +0.114 & +0.105 & +0.238 & +0.246 \\

\midrule


\multirow{7}{*}{Nepali}
 & Baseline & 0.555 & 0.326 & 0.431 & 0.248 & 0.521 & 0.737 & 0.357 & 0.449 & 0.442 \\

 & 2Step-CM & +0.174 & +0.012 & +0.138 & +0.135 & +0.151 & +0.060 & +0.060 & +0.171 & +0.282 \\

 & 2Step-EN & +0.186 & -0.120 & +0.113 & +0.098 & +0.174 & +0.114 & +0.054 & +0.220 & +0.337 \\
\addlinespace[2pt]

 & 1Step-CM+Ans & +0.204 & -0.032 & +0.138 & +0.119 & +0.158 & +0.134 & +0.027 & +0.210 & +0.342 \\

 & 1Step-EN+Ans & +0.251 & -0.111 & +0.108 & +0.149 & +0.195 & +0.145 & +0.004 & +0.225 & +0.365 \\
\addlinespace[2pt]

 & TinT-CM & +0.194 & +0.043 & +0.144 & +0.175 & +0.159 & +0.123 & +0.074 & +0.217 & +0.333 \\

 & TinT-EN & +0.213 & -0.033 & +0.166 & +0.184 & +0.171 & +0.133 & +0.079 & +0.234 & +0.351 \\

\midrule


\multirow{7}{*}{Odia}
 & Baseline & 0.440 & 0.189 & 0.241 & 0.138 & 0.357 & 0.587 & 0.299 & 0.352 & 0.397 \\

 & 2Step-CM & +0.178 & -0.029 & -0.009 & +0.072 & +0.174 & +0.122 & -0.095 & +0.126 & +0.239 \\

 & 2Step-EN & +0.203 & -0.063 & -0.014 & +0.065 & +0.165 & +0.209 & -0.110 & +0.162 & +0.280 \\
\addlinespace[2pt]

 & 1Step-CM+Ans & +0.131 & +0.012 & -0.019 & +0.068 & +0.176 & +0.204 & -0.104 & -0.021 & +0.247 \\

 & 1Step-EN+Ans & +0.230 & -0.058 & +0.035 & +0.120 & +0.177 & +0.235 & -0.075 & +0.053 & +0.291 \\
\addlinespace[2pt]

 & TinT-CM & +0.147 & +0.056 & -0.017 & +0.102 & +0.184 & +0.177 & -0.068 & +0.153 & +0.241 \\

 & TinT-EN & +0.176 & +0.004 & +0.015 & +0.086 & +0.181 & +0.198 & -0.061 & +0.183 & +0.258 \\

\midrule


\multirow{7}{*}{Punjabi}
 & Baseline & 0.528 & 0.281 & 0.433 & 0.183 & 0.485 & 0.708 & 0.289 & 0.410 & 0.509 \\

 & 2Step-CM & +0.173 & -0.054 & +0.080 & -0.008 & +0.136 & +0.031 & -0.003 & +0.113 & +0.172 \\

 & 2Step-EN & +0.220 & -0.146 & +0.117 & -0.002 & +0.163 & +0.095 & -0.012 & +0.184 & +0.235 \\
\addlinespace[2pt]

 & 1Step-CM+Ans & +0.179 & -0.028 & +0.103 & +0.058 & +0.143 & +0.116 & +0.028 & +0.178 & +0.226 \\

 & 1Step-EN+Ans & +0.251 & -0.192 & +0.068 & +0.049 & +0.150 & +0.123 & -0.004 & +0.206 & +0.272 \\
\addlinespace[2pt]

 & TinT-CM & +0.184 & -0.086 & +0.121 & +0.041 & +0.132 & +0.104 & +0.017 & +0.185 & +0.231 \\

 & TinT-EN & +0.206 & -0.099 & +0.135 & +0.097 & +0.166 & +0.112 & +0.075 & +0.201 & +0.249 \\

\midrule


\multirow{7}{*}{Sanskrit}
 & Baseline & 0.536 & 0.314 & 0.386 & 0.254 & 0.524 & 0.733 & 0.282 & 0.436 & 0.530 \\

 & 2Step-CM & +0.198 & +0.010 & +0.172 & +0.109 & +0.123 & +0.064 & +0.081 & +0.183 & +0.191 \\

 & 2Step-EN & +0.236 & -0.105 & +0.178 & +0.090 & +0.132 & +0.111 & +0.083 & +0.226 & +0.243 \\
\addlinespace[2pt]

 & 1Step-CM+Ans & +0.208 & -0.023 & +0.141 & +0.124 & +0.153 & +0.125 & +0.040 & +0.232 & +0.261 \\

 & 1Step-EN+Ans & +0.272 & -0.111 & +0.162 & +0.144 & +0.186 & +0.148 & +0.007 & +0.251 & +0.285 \\
\addlinespace[2pt]

 & TinT-CM & +0.218 & +0.044 & +0.170 & +0.171 & +0.151 & +0.120 & +0.106 & +0.241 & +0.249 \\

 & TinT-EN & +0.239 & -0.034 & +0.186 & +0.182 & +0.168 & +0.129 & +0.127 & +0.257 & +0.265 \\

\midrule


\multirow{7}{*}{Sindhi}
 & Baseline & 0.493 & 0.268 & 0.337 & 0.211 & 0.452 & 0.606 & 0.263 & 0.394 & 0.474 \\

 & 2Step-CM & +0.121 & -0.034 & +0.048 & +0.081 & +0.108 & +0.063 & +0.028 & +0.105 & +0.118 \\

 & 2Step-EN & +0.176 & -0.155 & +0.046 & +0.048 & +0.121 & +0.148 & -0.001 & +0.082 & +0.115 \\
\addlinespace[2pt]

 & 1Step-CM+Ans & +0.169 & -0.056 & +0.086 & +0.063 & +0.128 & +0.168 & +0.035 & +0.162 & +0.186 \\

 & 1Step-EN+Ans & +0.235 & -0.141 & +0.070 & +0.128 & +0.158 & +0.204 & +0.025 & +0.173 & +0.211 \\
\addlinespace[2pt]

 & TinT-CM & +0.162 & -0.004 & +0.086 & +0.107 & +0.129 & +0.149 & +0.053 & +0.167 & +0.180 \\

 & TinT-EN & +0.192 & -0.087 & +0.087 & +0.141 & +0.145 & +0.163 & +0.033 & +0.183 & +0.202 \\

\midrule


\multirow{7}{*}{Tamil}
 & Baseline & 0.529 & 0.284 & 0.428 & 0.272 & 0.518 & 0.731 & 0.280 & 0.375 & 0.442 \\

 & 2Step-CM & +0.160 & -0.034 & +0.110 & +0.092 & +0.138 & +0.020 & +0.001 & +0.124 & +0.172 \\

 & 2Step-EN & +0.180 & -0.111 & +0.123 & +0.086 & +0.165 & +0.118 & -0.006 & +0.150 & +0.216 \\
\addlinespace[2pt]

 & 1Step-CM+Ans & +0.163 & -0.031 & +0.099 & +0.098 & +0.156 & +0.127 & -0.005 & +0.141 & +0.238 \\

 & 1Step-EN+Ans & +0.229 & -0.155 & +0.068 & +0.139 & +0.184 & +0.144 & +0.004 & +0.176 & +0.258 \\
\addlinespace[2pt]

 & TinT-CM & +0.176 & -0.033 & +0.082 & +0.133 & +0.155 & +0.116 & +0.037 & +0.162 & +0.227 \\

 & TinT-EN & +0.188 & -0.126 & +0.081 & +0.164 & +0.166 & +0.119 & +0.036 & +0.180 & +0.239 \\

\midrule


\multirow{7}{*}{Telugu}
 & Baseline & 0.555 & 0.324 & 0.458 & 0.255 & 0.533 & 0.747 & 0.298 & 0.408 & 0.530 \\

 & 2Step-CM & +0.163 & -0.008 & +0.118 & +0.104 & +0.141 & +0.052 & +0.003 & +0.168 & +0.171 \\

 & 2Step-EN & +0.212 & -0.135 & +0.119 & +0.078 & +0.163 & +0.117 & -0.005 & +0.212 & +0.226 \\
\addlinespace[2pt]

 & 1Step-CM+Ans & +0.148 & -0.036 & +0.097 & +0.096 & +0.161 & +0.125 & -0.006 & +0.199 & +0.217 \\

 & 1Step-EN+Ans & +0.238 & -0.145 & +0.074 & +0.155 & +0.185 & +0.144 & +0.003 & +0.220 & +0.246 \\
\addlinespace[2pt]

 & TinT-CM & +0.186 & -0.003 & +0.114 & +0.133 & +0.155 & +0.114 & +0.026 & +0.206 & +0.212 \\

 & TinT-EN & +0.199 & -0.071 & +0.133 & +0.173 & +0.179 & +0.124 & +0.053 & +0.222 & +0.228 \\

\midrule


\multirow{7}{*}{Urdu}
 & Baseline & 0.563 & 0.315 & 0.438 & 0.230 & 0.523 & 0.715 & 0.351 & 0.455 & 0.531 \\

 & 2Step-CM & +0.132 & -0.078 & +0.113 & +0.054 & +0.122 & +0.043 & +0.014 & +0.105 & +0.128 \\

 & 2Step-EN & +0.190 & -0.158 & +0.135 & +0.054 & +0.139 & +0.120 & +0.010 & +0.163 & +0.206 \\
\addlinespace[2pt]

 & 1Step-CM+Ans & +0.168 & -0.084 & +0.094 & +0.062 & +0.124 & +0.119 & -0.008 & +0.165 & +0.218 \\

 & 1Step-EN+Ans & +0.225 & -0.152 & +0.090 & +0.116 & +0.153 & +0.136 & -0.034 & +0.191 & +0.241 \\
\addlinespace[2pt]

 & TinT-CM & +0.154 & -0.026 & +0.107 & +0.100 & +0.120 & +0.107 & +0.008 & +0.171 & +0.197 \\

 & TinT-EN & +0.184 & -0.113 & +0.112 & +0.130 & +0.141 & +0.109 & +0.020 & +0.191 & +0.217 \\

\midrule

\bottomrule

\caption{Consolidated Cross-Lingual Consistency (CLC) results across all model families and prompting strategies. Scores denote the average pairwise Jaccard overlap between the sets of correctly predicted factual indices across multilingual language pairs.}

\end{longtable}
}

\twocolumn
}
\normalsize

{

\begin{table*}[h]
\centering
\small
\begin{tabular}{llccc}
\toprule
\textbf{Model} & \textbf{Language} & \textbf{Baseline Wrong} & \textbf{Baseline $\rightarrow$ TinT-CM} & \textbf{TinT-CM $\rightarrow$ TinT-EN} \\
 &  &  & \textbf{(Wrong to Correct)} & \textbf{(Wrong to Correct)} \\
\midrule
\multirow{5}{*}{Llama-3.1-8B-Instruct}
 & Assamese & 1375 & 561 & 61 \\
 & Bengali  & 1090 & 466 & 77 \\
 & Hindi    & 1068 & 534 & 64 \\
 & Marathi  & 1069 & 518 & 76 \\
 & Telugu   & 1286 & 536 & 84 \\
\midrule
\multirow{5}{*}{Gemma-3-12B-it}
 & Assamese & 1153 & 615 & 75 \\
 & Bengali  &  678 & 416 & 38 \\
 & Hindi    &  735 & 432 & 23 \\
 & Marathi  &  618 & 346 & 36 \\
 & Telugu   &  669 & 377 & 33 \\
\midrule
\multirow{5}{*}{Qwen2.5-14B-Instruct}
 & Assamese & 1549 & 742 & 87 \\
 & Bengali  & 1169 & 746 & 58 \\
 & Hindi    & 1248 & 805 & 58 \\
 & Marathi  & 1102 & 674 & 54 \\
 & Telugu   & 1271 & 629 & 77 \\
\bottomrule
\end{tabular}
\caption{
Transition analysis for the TinT setting (Baseline $\rightarrow$ TinT-CM $\rightarrow$ TinT-EN) across five representative Indian languages and three evaluated LLMs. The table reports the number of baseline-incorrect instances and the subset of those instances that transition from incorrect to correct under TinT-CM and then under TinT-EN, explicitly separating the recoveries achieved after the code-mixing transition from those achieved after the subsequent through-English reasoning transition.
}
\label{tab:flip_transitions}
\end{table*}
}

\section{Flip Point Analysis}
\label{app:flip_point}

We provide the full transition counts underlying the flip-point analysis in Section~\ref{sec:res_gen_flip}. For each baseline-incorrect instance, we record whether the prediction is recovered under TinT-CM and, if it remains incorrect, whether it is subsequently recovered under TinT-EN. Table~\ref{tab:flip_transitions} reports these transitions separately for each language and model.

\begin{figure*}[h]
\centering


\begin{subfigure}[t]{0.24\textwidth}
    \centering
    \includegraphics[width=\linewidth]{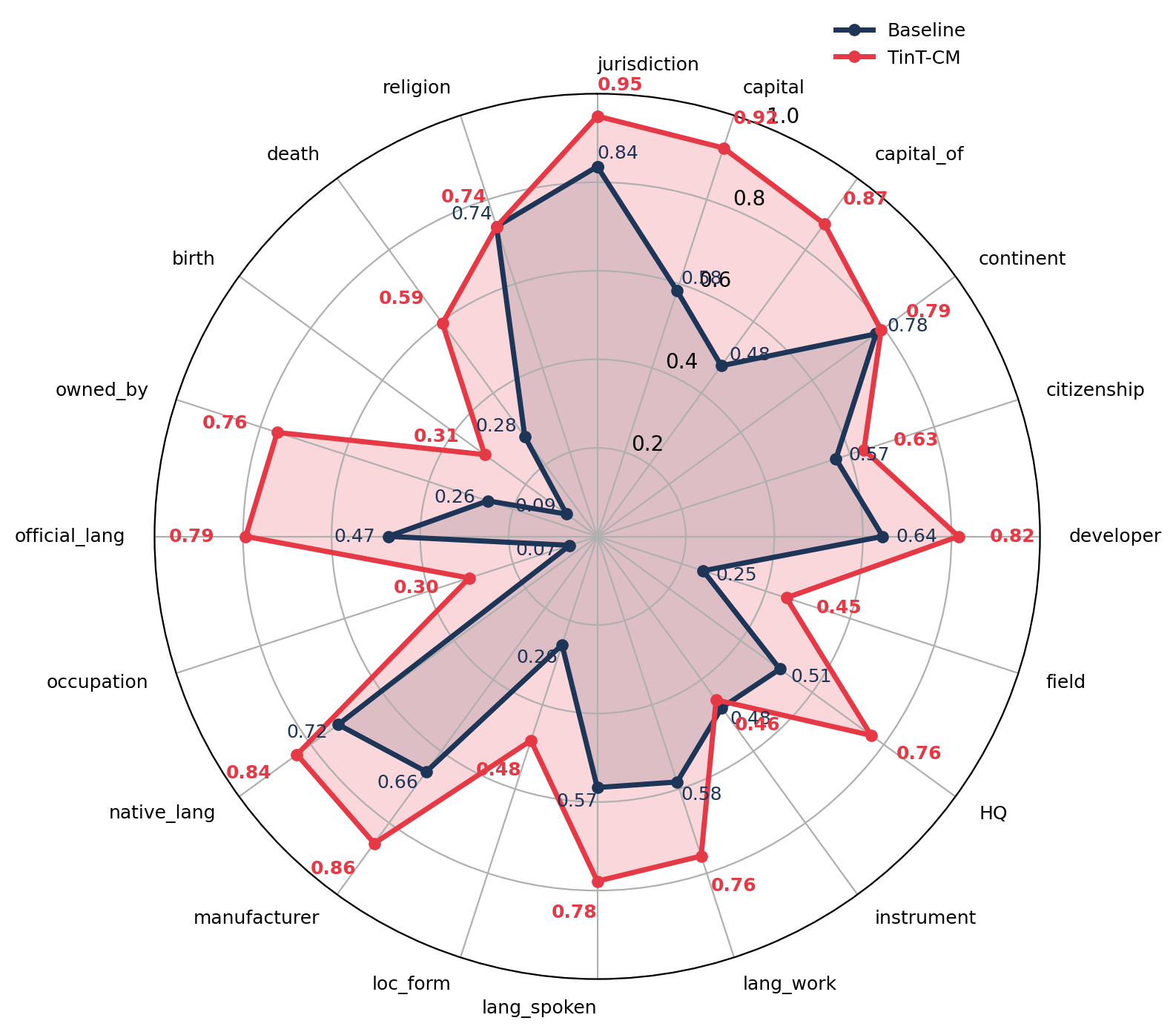}
    \caption{Dogri}
\end{subfigure}
\begin{subfigure}[t]{0.24\textwidth}
    \centering
    \includegraphics[width=\linewidth]{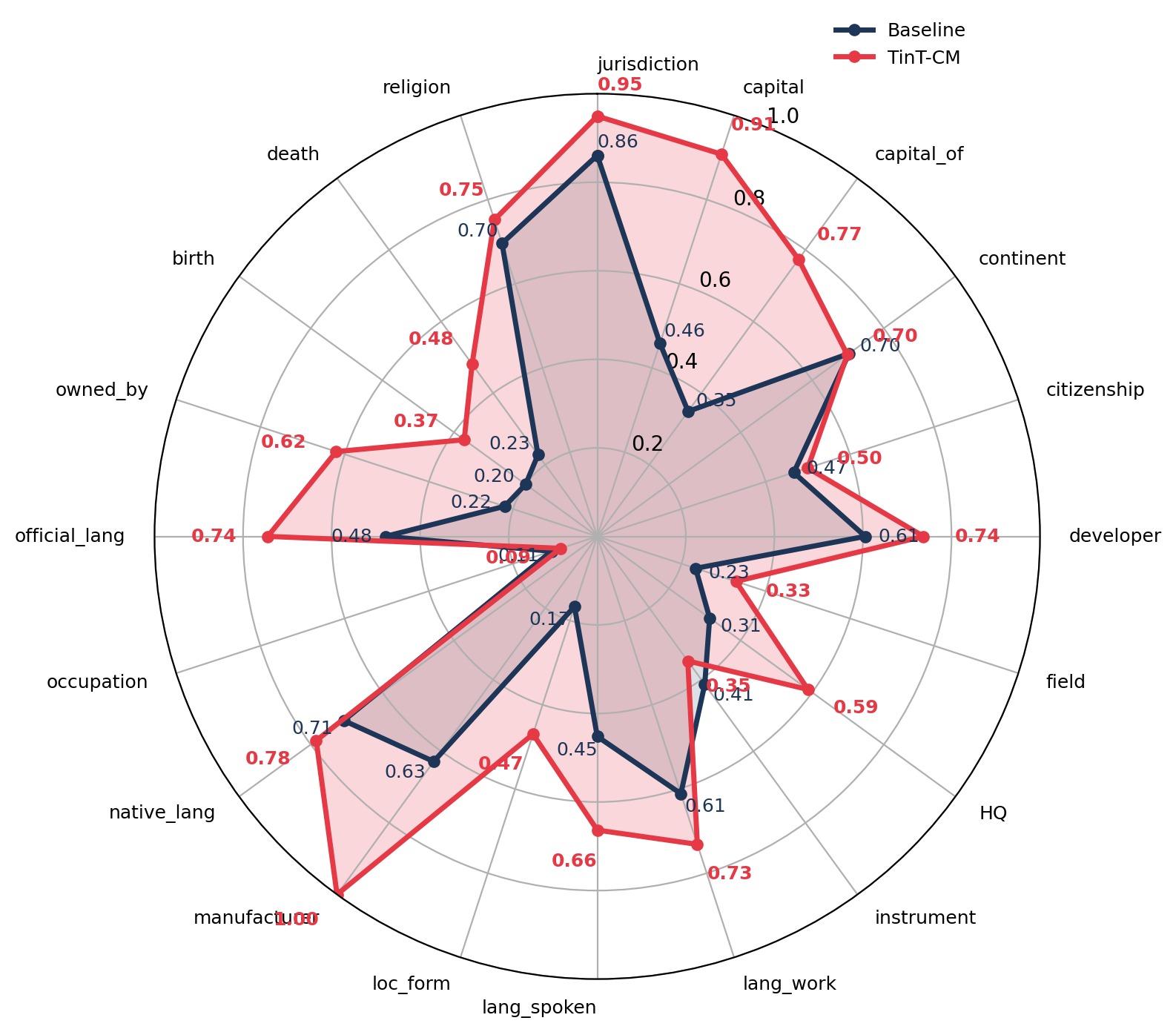}
    \caption{Gujarati}
\end{subfigure}
\begin{subfigure}[t]{0.24\textwidth}
    \centering
    \includegraphics[width=\linewidth]{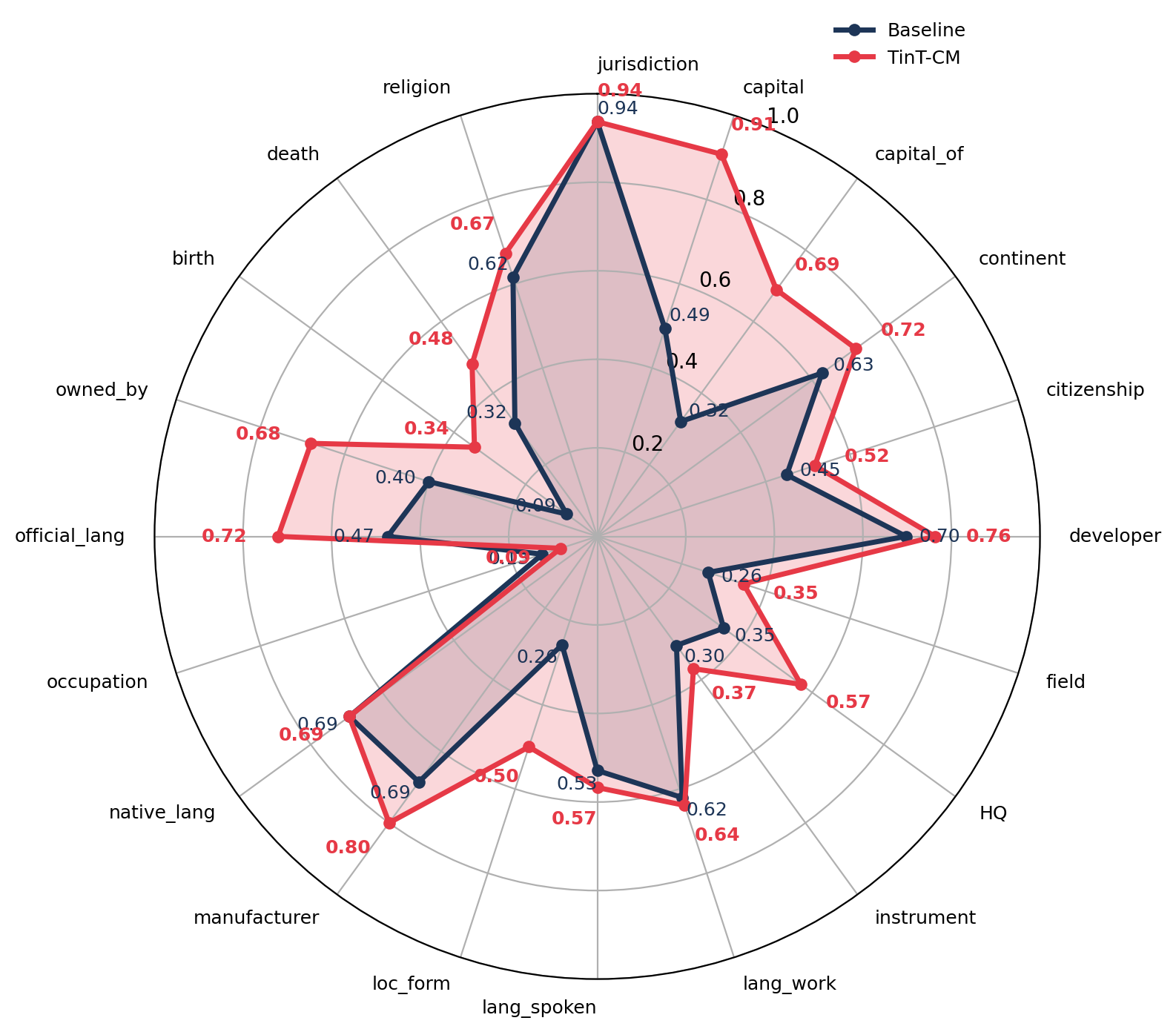}
    \caption{Kannada}
\end{subfigure}
\begin{subfigure}[t]{0.24\textwidth}
    \centering
    \includegraphics[width=\linewidth]{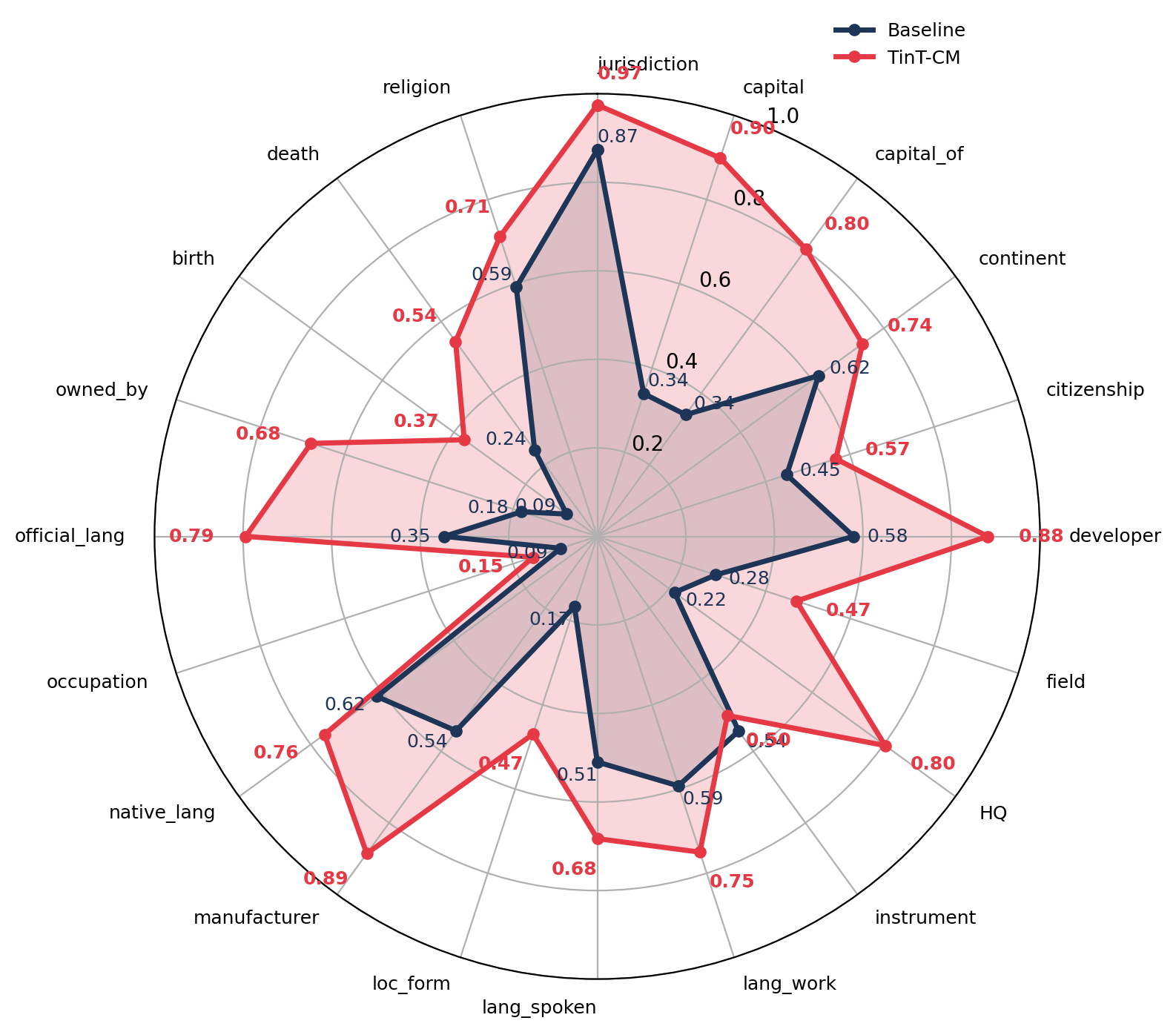}
    \caption{Konkani}
\end{subfigure}
\begin{subfigure}[t]{0.24\textwidth}
    \centering
    \includegraphics[width=\linewidth]{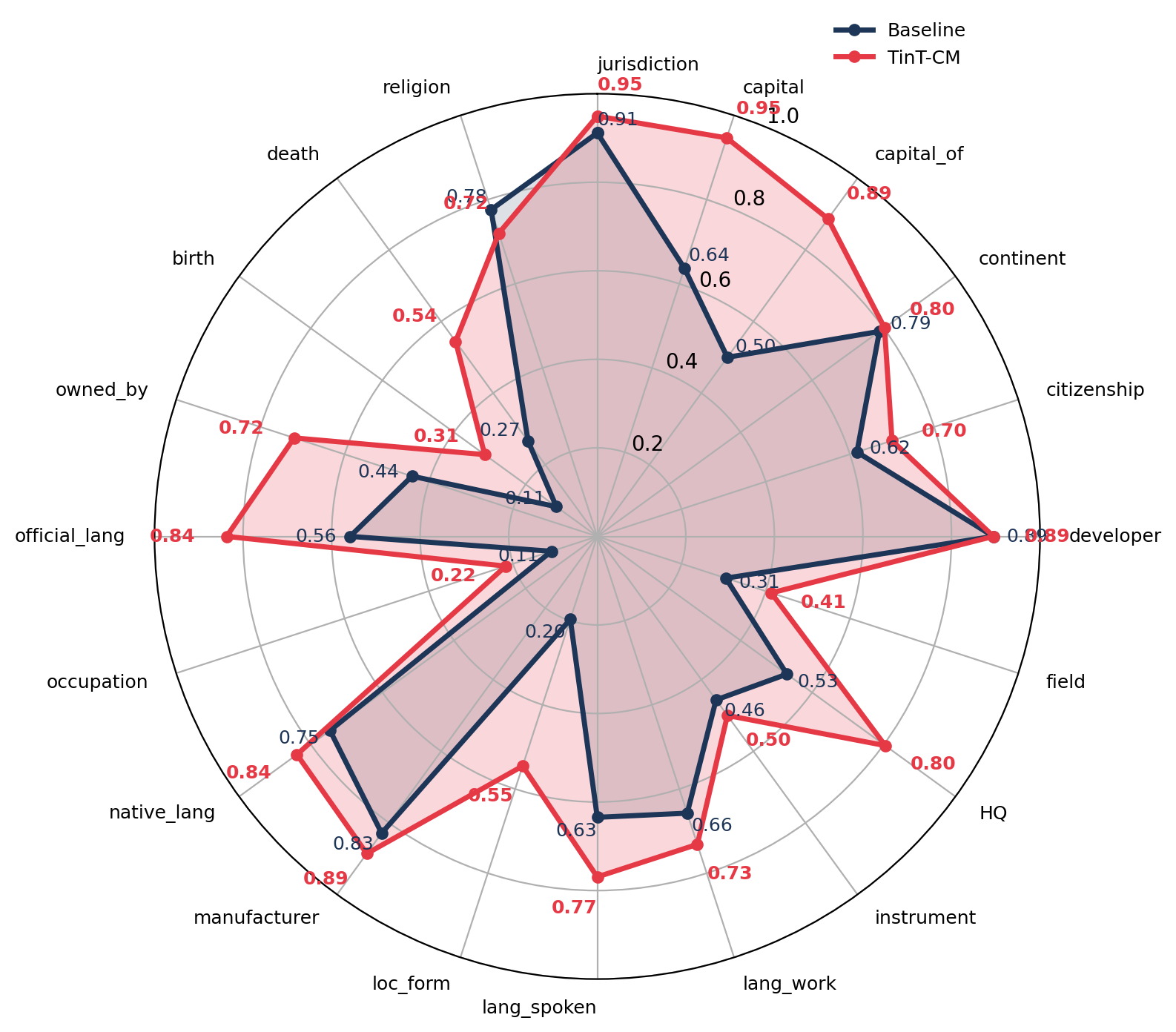}
    \caption{Maithili}
\end{subfigure}
\begin{subfigure}[t]{0.24\textwidth}
    \centering
    \includegraphics[width=\linewidth]{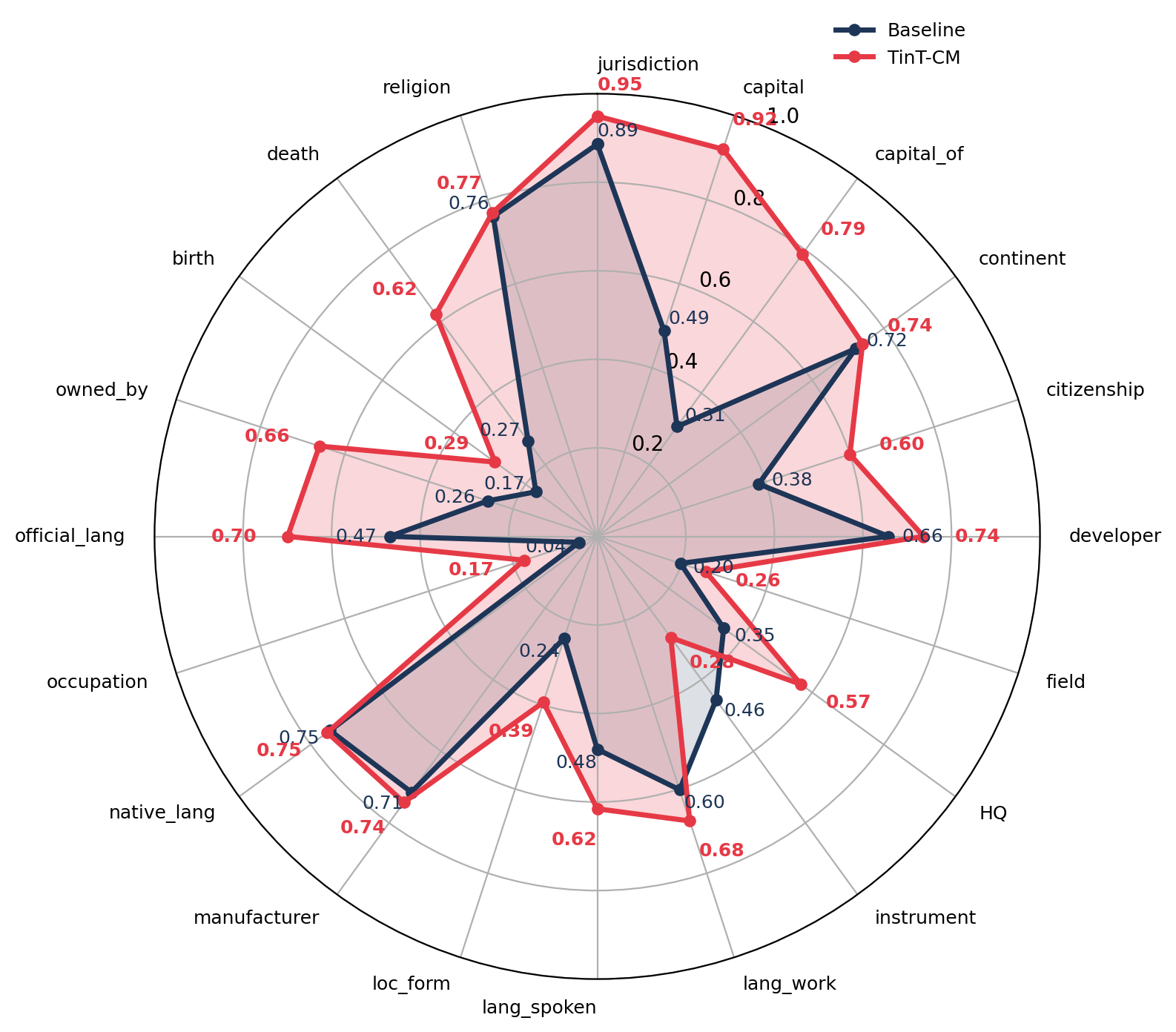}
    \caption{Malayalam}
\end{subfigure}
\begin{subfigure}[t]{0.24\textwidth}
    \centering
    \includegraphics[width=\linewidth]{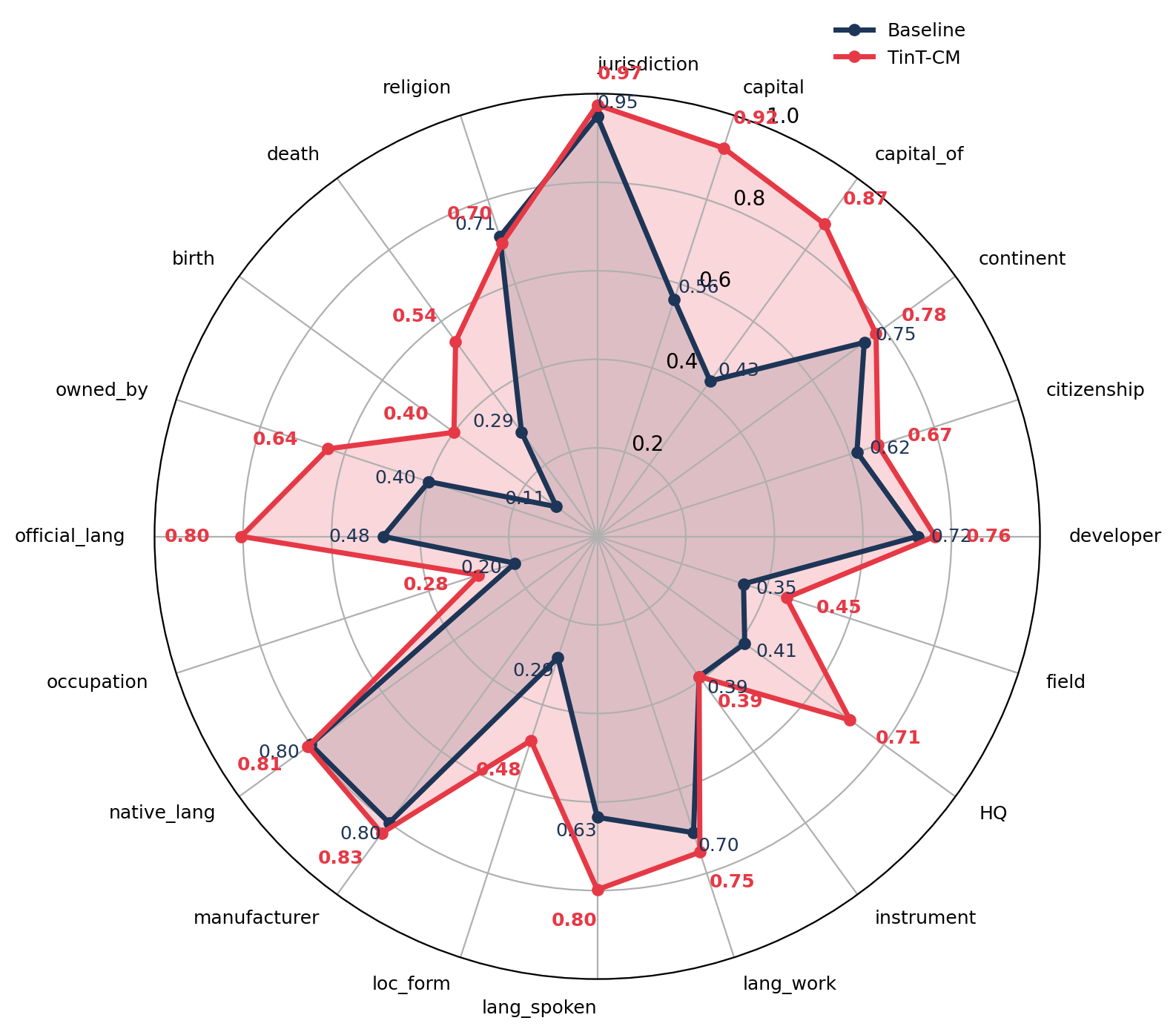}
    \caption{Marathi}
\end{subfigure}
\begin{subfigure}[t]{0.24\textwidth}
    \centering
    \includegraphics[width=\linewidth]{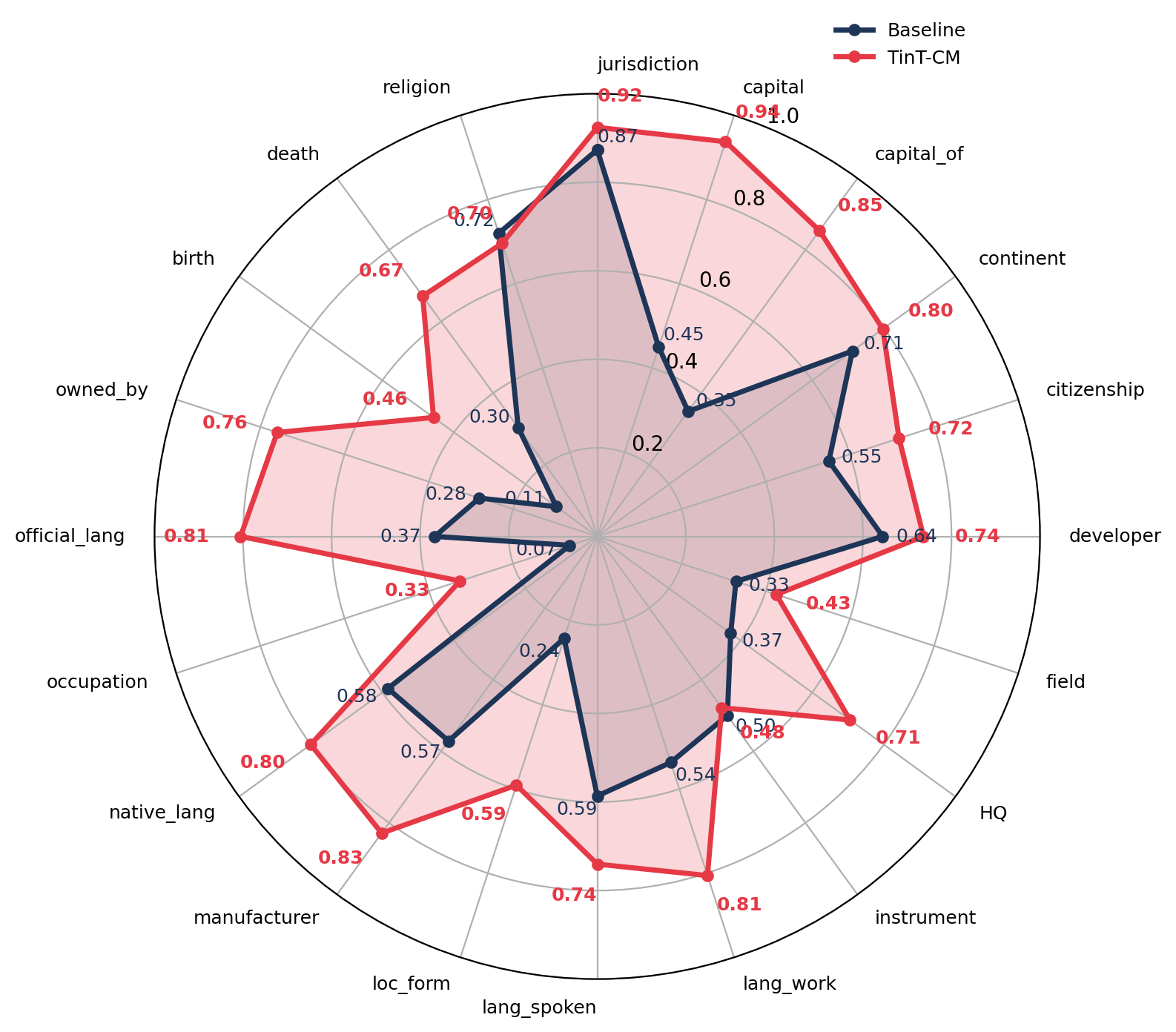}
    \caption{Nepali}
\end{subfigure}
\begin{subfigure}[t]{0.24\textwidth}
    \centering
    \includegraphics[width=\linewidth]{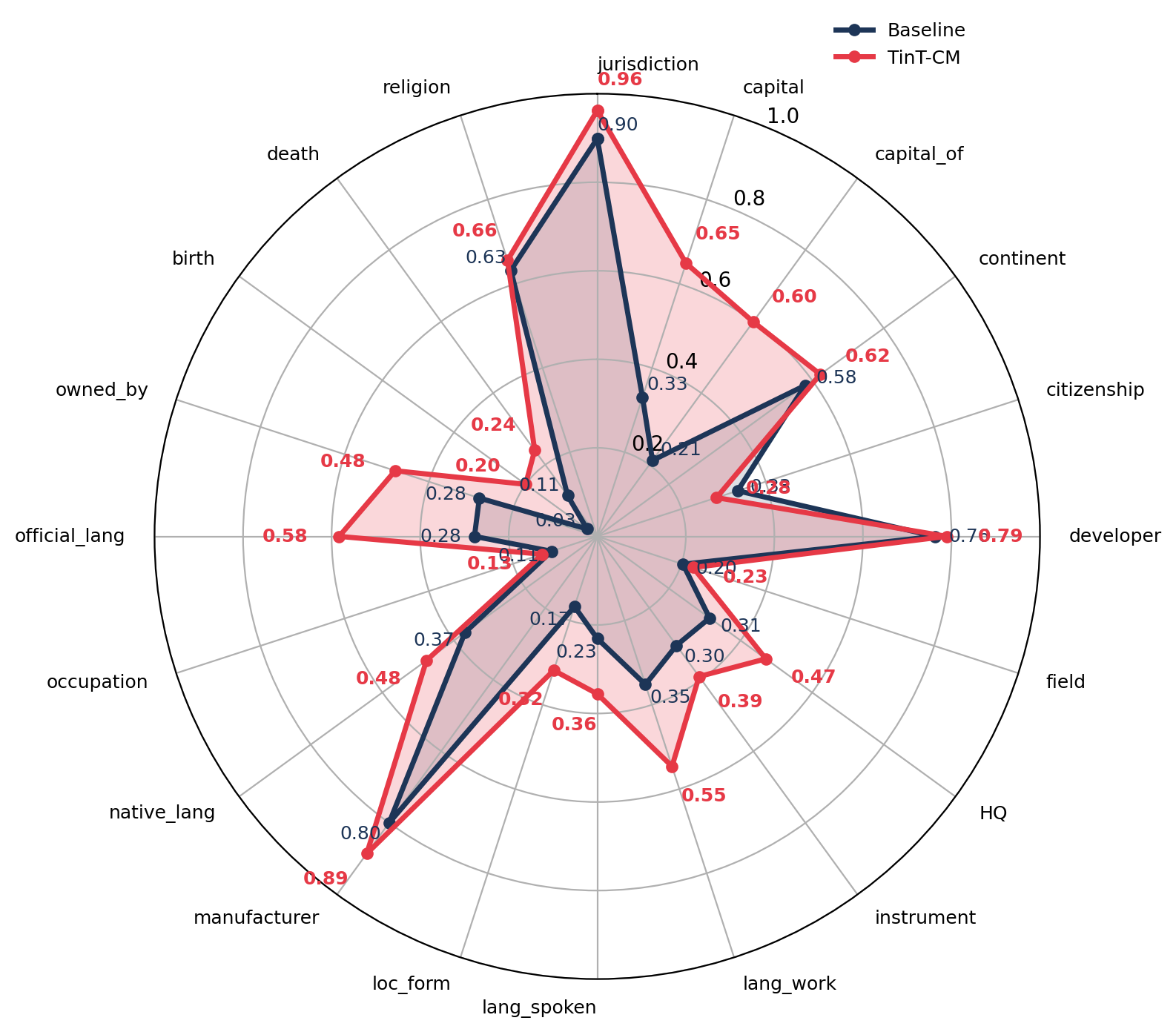}
    \caption{Odia}
\end{subfigure}
\begin{subfigure}[t]{0.24\textwidth}
    \centering
    \includegraphics[width=\linewidth]{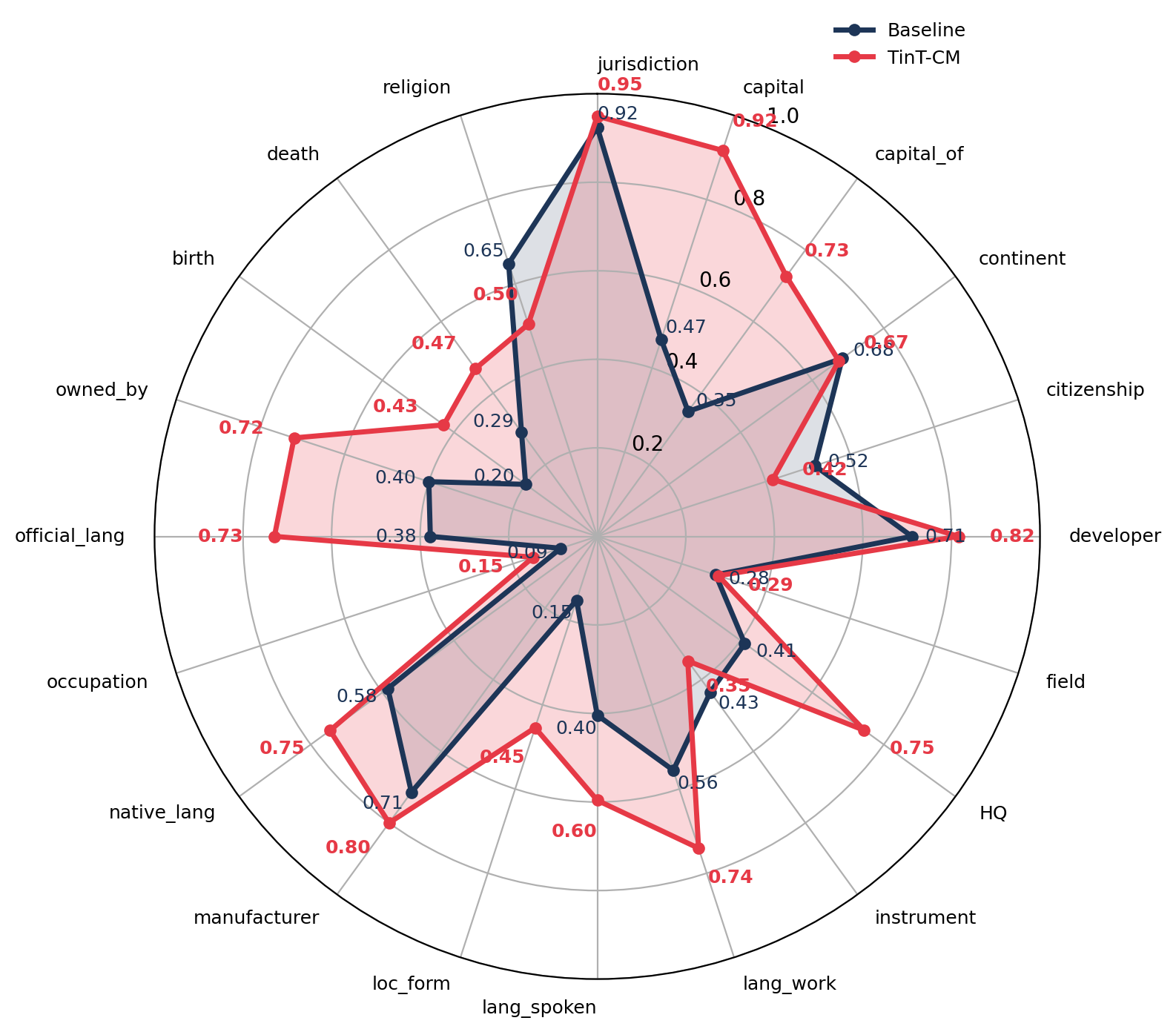}
    \caption{Punjabi}
\end{subfigure}
\begin{subfigure}[t]{0.24\textwidth}
    \centering
    \includegraphics[width=\linewidth]{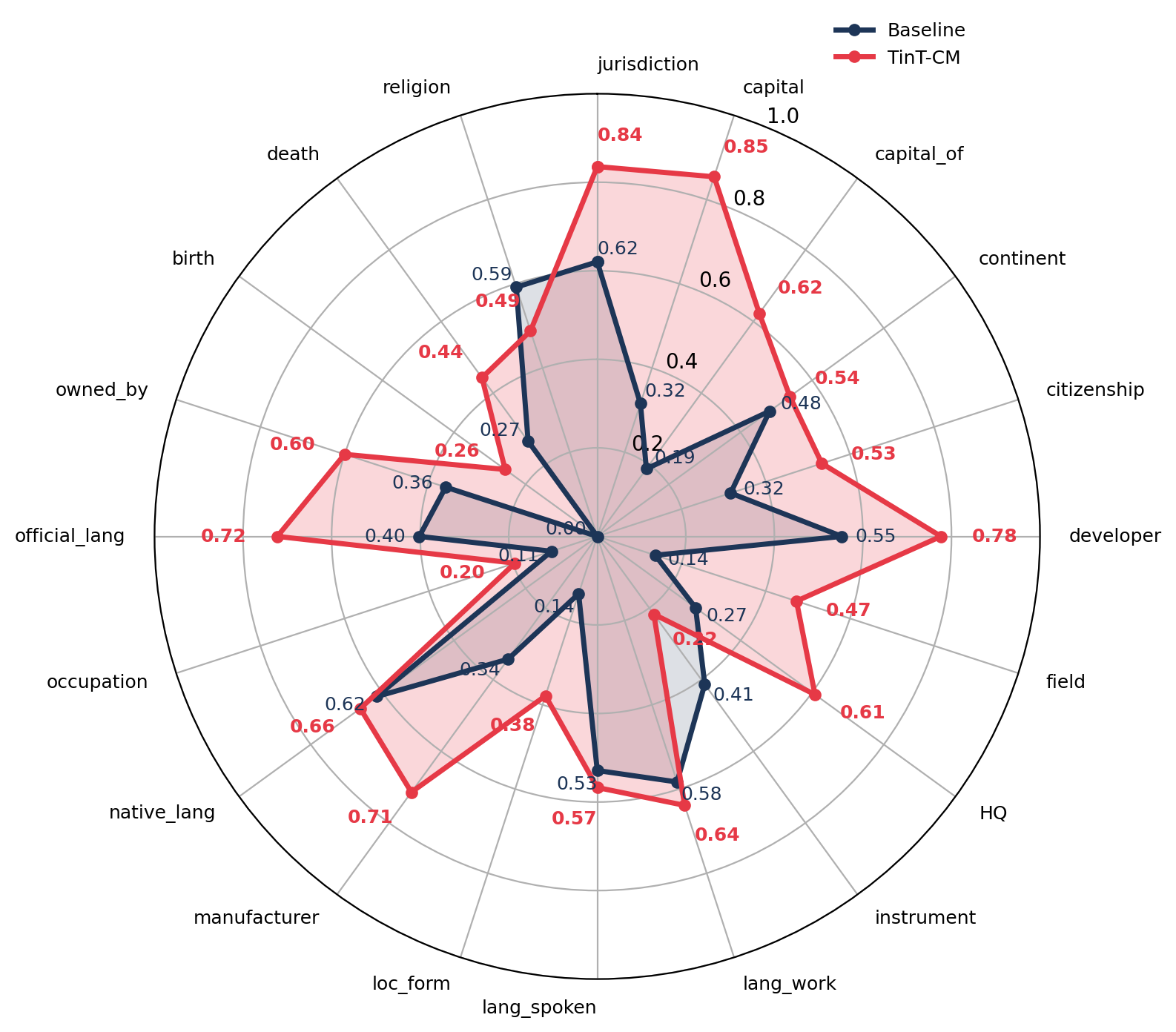}
    \caption{Sindhi}
\end{subfigure}
\begin{subfigure}[t]{0.24\textwidth}
    \centering
    \includegraphics[width=\linewidth]{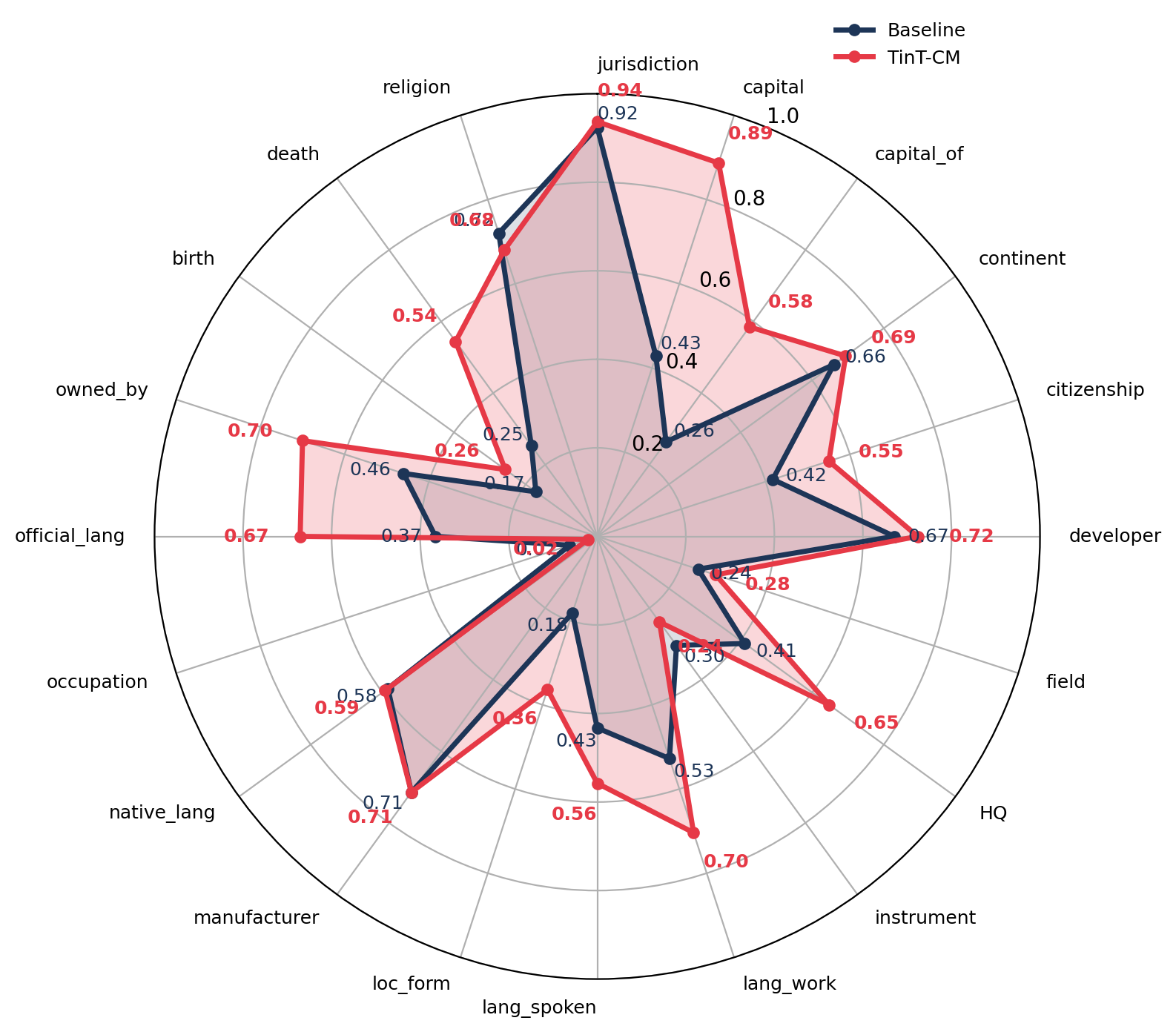}
    \caption{Tamil}
\end{subfigure}
\begin{subfigure}[t]{0.24\textwidth}
    \centering
    \includegraphics[width=\linewidth]{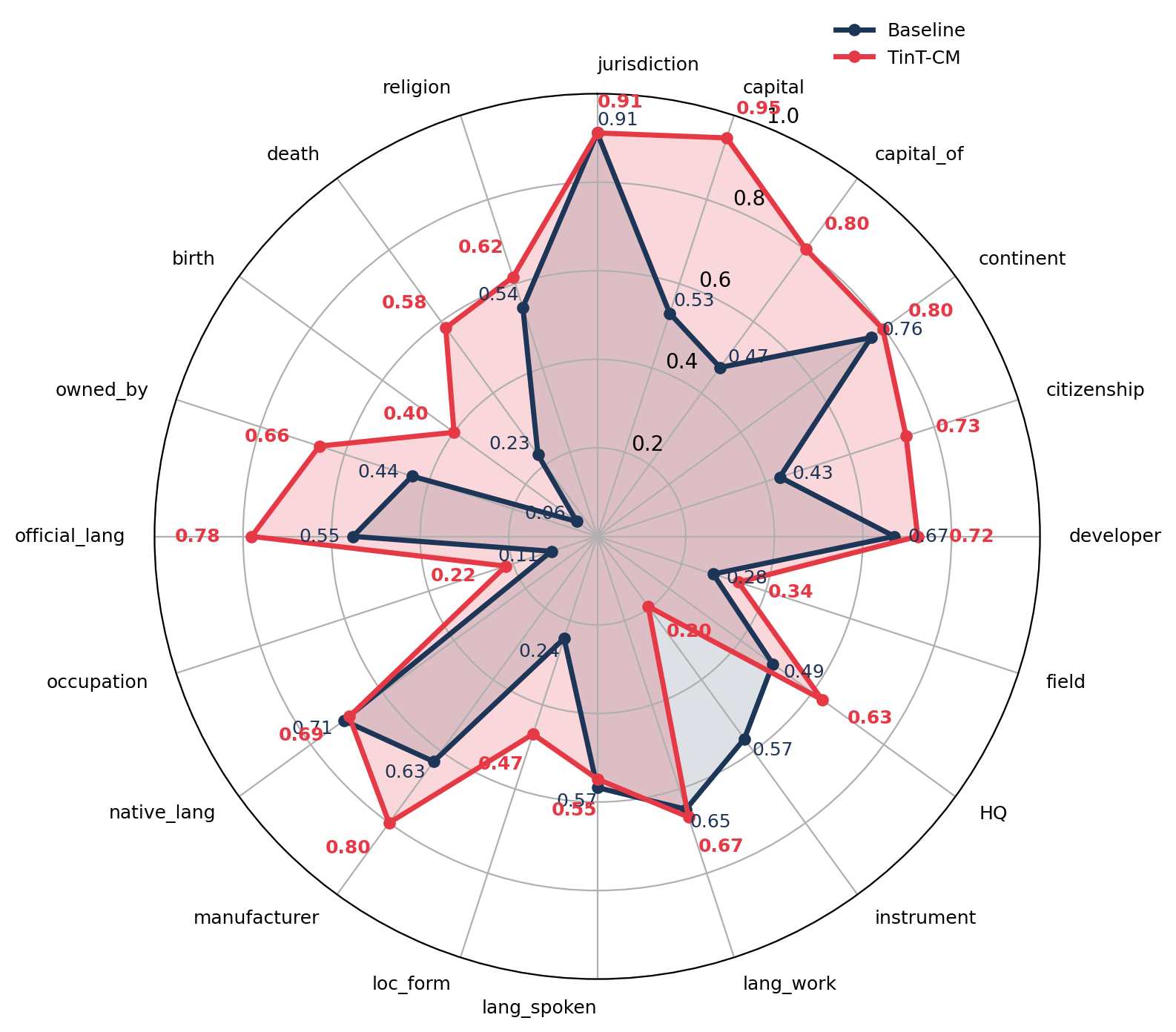}
    \caption{Urdu}
\end{subfigure}

\caption{Relation-wise performance comparison of TinT prompting and baseline prompting across the remaining 13 languages using \texttt{meta-llama/Llama-3.1-8B-Instruct}.}
\label{fig:radar_remaining_languages}
\end{figure*}

\begin{figure*}[t]
    \centering
    \includegraphics[width=\textwidth]{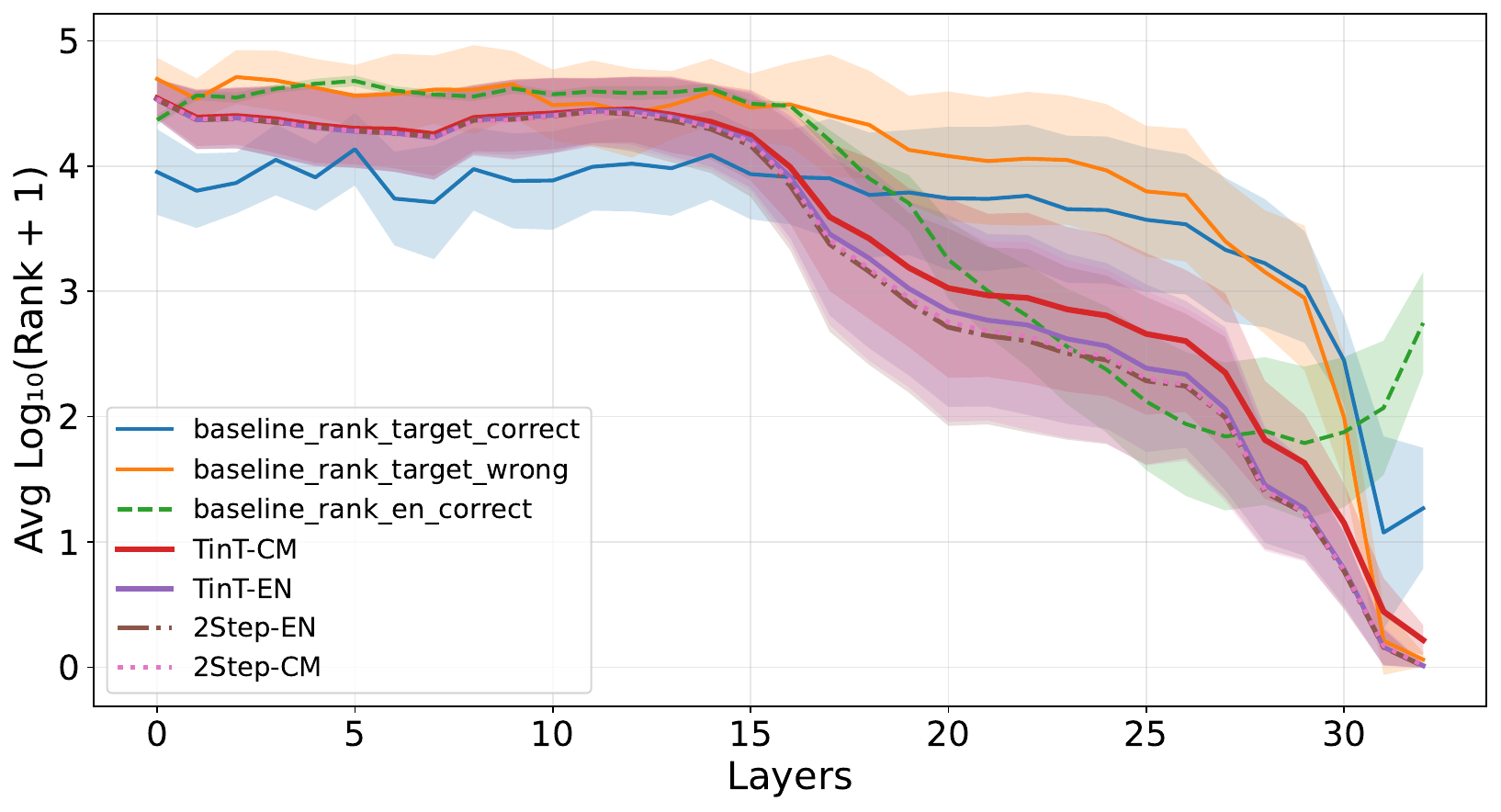}
    \caption{%
Extended layer-wise logit-lens analysis for Baseline, 2Step, TinT-CM,
and TinT-EN prompting. We track the rank of the correct answer token
across model layers, where lower rank indicates a stronger preference
for the correct answer. The trajectories provide a layer-wise
comparison of how the different prompting strategies recover the
correct answer representation.
}
    \label{fig:extended_logit_lens}
\end{figure*}

\section{Code-Mixed Dataset Generation}
\label{app:cm_generation}

The code-mixed variants in IndicKLAR were generated at the template level using GPT API. GPT was used to produce semantically equivalent code-mixed templates while preserving the underlying question, placeholders, and grammatical structure of the target Indic language. Generated templates were manually reviewed by native speakers for fluency and naturalness before inclusion in the benchmark. %

\begin{cmechpromptbox}{Code-Mixed Dataset Generation}
\begin{lstlisting}
You are generating code-mixed factual question templates for an Indic-English factual knowledge benchmark.

Given the original factual question in English and its corresponding <target-language> template, generate a natural code-mixed template that preserves the meaning and structure of the original question.

Follow these instructions strictly:

1. Preserve the original question meaning and factual relation.

2. Preserve all placeholders exactly as provided, including <subject> and <mask>.

3. Preserve the grammatical structure of the target language (<target-language>) and maintain a natural word order.

4. Preserve the original question type and relation.

5. Do not add, remove, or alter factual information.

6. Do not translate the entire question into English.

7. Do not translate or alter the <subject> placeholder.

8. Preserve the answer slot exactly as <mask>.

9. Produce a natural and concise code-mixed template suitable for English-<target-language> bilingual communication.

10. Return only the generated code-mixed template.

Original factual question:
<ENGLISH_TEMPLATE>

<target-language> template:
<NATIVE_TEMPLATE>

Code-mixed template:
\end{lstlisting}
\end{cmechpromptbox}

\section{Prompting Strategies}
\label{app:prompting_strategies}

This section presents the prompts used in our experiments.  Variables enclosed in angle-brackets (\texttt{SOURCE\_LANG},
\texttt{SOURCE\_SCRIPT}, \texttt{TARGET\_LANG}) are runtime configuration
parameters passed via command-line arguments.  Placeholders written in
\textsc{small-caps} (e.g., \texttt{QUESTION}, \texttt{ANSWER}) represent
concrete values filled in at runtime from the dataset.

\subsection{Baseline Prompting}
\label{sec:baseline}

We evaluate models under a 3-shot cloze-completion setting. Each prompt is constructed from a relation-specific template containing two placeholders: \texttt{<subject>}, filled with the entity name, and \texttt{<mask>}, which is removed to leave an open-ended sentence for the model to complete. For each test instance, three demonstrations are sampled from the same relation and language, and prepended as few-shot pairs. The test item is appended without its target, and the model generates the answer as a greedy continuation. Each demonstration line is formed by filling \texttt{<subject>} with the entity name, stripping \texttt{<mask>}, and appending the gold object with a single leading space. The test item omits the target entirely, prompting the model to continue the incomplete sentence. A prediction is accepted as correct if it is a case-insensitive prefix of the gold answer, or vice versa.

A second variant extends this setup by appending the set of candidate answers directly to the test item, followed by an \texttt{Answer:} marker, while keeping the three few-shot demonstrations in the same plain completion format. When candidate answers are available, the test item is constructed by appending the candidate list and the \texttt{Answer:} marker to the filled template. If no candidates are provided, the prompt falls back to the plain completion format described above.

\begin{cmechpromptbox}{Baseline Prompts}
\begin{lstlisting}
[template with <subject_1> filled] <answer_1>
[template with <subject_2> filled] <answer_2>
[template with <subject_3> filled] <answer_3>
[template with <subject_4> filled]
\end{lstlisting}
\hfill{\small [Baseline]}

\vspace{0.5em}

\begin{lstlisting}
[template with <subject_1> filled] <answer_1>
[template with <subject_2> filled] <answer_2>
[template with <subject_3> filled] <answer_3>
[template with <subject_4> filled]
Candidates: <candidate_1>, <candidate_2>, ..., <candidate_n>
Answer:
\end{lstlisting}
\hfill{\small [Baseline + Candidates]}
\end{cmechpromptbox}

\subsection{Two-Step Prompting Strategy}
\label{sec:two_call}

The 2 Step strategy decomposes each query into two sequential
inference passes. Stage~1 reformulates the source-language question
into an intermediate form; Stage~2 receives that reformulated question
together with three in-context demonstrations and selects the correct
answer from the candidate list. Both stages return structured JSON
output. We evaluate three reformulation strategies under this
framework.

\subsubsection{English Reformulation}
\label{sec:two_call_en}

Stage~1 translates the source-language question into English.
Stage~2 receives the English question and answers in the source
script.

\paragraph{Stage 1 — Translation.}

\begin{cmechpromptbox}{English Reformulation --- Stage 1}
\begin{lstlisting}
You are a translator.
Translate the following question from <SOURCE_LANG> (<SOURCE_SCRIPT>)
into English.

Output ONLY: {"translation": "..."}

Q: <QUESTION>
Output:
\end{lstlisting}
\end{cmechpromptbox}

\paragraph{Stage 2 — Answer.}

The English question from Stage~1 is presented alongside three source-language demonstrations, each showing the original question, its candidates, and the gold answer.

\begin{cmechpromptbox}{English Reformulation --- Stage 2}
\begin{lstlisting}
You are answering a factual question in English.
Select the correct answer from the candidates and write 
it in <SOURCE_LANG> (<SOURCE_SCRIPT>).

Output ONLY: {"answer": "..."}

Rules:
- Answer MUST be from the candidate list
- Answer MUST be in <SOURCE_SCRIPT> script

Q: <QUESTION_1>
Candidates: <CANDIDATES_1>
Output: {"answer": "<ANSWER_1>"}

Q: <QUESTION_2>
Candidates: <CANDIDATES_2>
Output: {"answer": "<ANSWER_2>"}

Q: <QUESTION_3>
Candidates: <CANDIDATES_3>
Output: {"answer": "<ANSWER_3>"}

Q: <TRANSLATED_QUESTION>
Candidates: <CANDIDATES>
Output:
\end{lstlisting}
\end{cmechpromptbox}

\subsubsection{Code-Mixed Reformulation}
\label{sec:two_call_cm}

Stage~1 converts the source-language question into a code-mixed
form, preserving the source-language grammatical structure while
replacing content words with English equivalents in Roman script.
Stage~2 answers the code-mixed question in the source script.

\paragraph{Stage 1 — Code-Mixed Conversion.}

\begin{cmechpromptbox}{Code-Mixed Reformulation --- Stage 1}
\begin{lstlisting}
You are a STRICT translator performing code-mixing.

Convert this <SOURCE_LANG> (<SOURCE_SCRIPT>) question into
<TARGET_LANG> (Roman script), preserving the original structure.

Definition of code-mixing:
- Keep <SOURCE_LANG> function words in romanized form
- Replace ONLY content words with English equivalents
- Preserve word order exactly
- Output MUST NOT be a fully fluent English sentence

Example (Hinglish):
  Input:     Bharat ki rajdhani kya hai?
  Output:    Bharat ki capital kya hai?

Output ONLY: {"translation": "<code-mixed question>"}

Q: <QUESTION>
Output:
\end{lstlisting}
\end{cmechpromptbox}

\paragraph{Stage 2 — Answer.}

Stage~2 is identical in structure to the English reformulation
Stage~2, with three source-language demonstrations and the
code-mixed question as the test item.

\begin{cmechpromptbox}{Code-Mixed Reformulation --- Stage 2}
\begin{lstlisting}
You are answering a factual question in <TARGET_LANG>.
This uses <SOURCE_LANG> grammar with English vocabulary
(Roman script).
Output ONLY: {"answer": "..."} in <SOURCE_SCRIPT> script.

Rules:
- Answer MUST be from the candidate list
- Answer MUST be in <SOURCE_SCRIPT> script

Q: <QUESTION_1>
Candidates: <CANDIDATES_1>
Output: {"answer": "<ANSWER_1>"}

Q: <QUESTION_2>
Candidates: <CANDIDATES_2>
Output: {"answer": "<ANSWER_2>"}

Q: <QUESTION_3>
Candidates: <CANDIDATES_3>
Output: {"answer": "<ANSWER_3>"}

Q: <CODE_MIXED_QUESTION>
Candidates: <CANDIDATES>
Output:
\end{lstlisting}
\end{cmechpromptbox}

\subsubsection{Transliteration-Based Reformulation}
\label{sec:two_call_translit}

Stage~1 phonetically converts the <source-language> question from
its native script into Roman script without translating any words.
Stage~2 answers the romanized question in the source script. This
variant isolates the effect of script conversion alone, keeping
all vocabulary in the source language.

\paragraph{Stage 1 — Transliteration.}

\begin{cmechpromptbox}{Transliteration --- Stage 1}
\begin{lstlisting}
You are a transliterator. Convert the following <SOURCE_LANG>
question from <SOURCE_SCRIPT> into Roman/Latin script.

- Rewrite every word phonetically in Roman script
- Do NOT translate or replace any word
- Preserve all words and their order exactly

Output ONLY: {"translation": "<romanized question>"}

Q: <QUESTION>
Output:
\end{lstlisting}
\end{cmechpromptbox}

\paragraph{Stage 2 — Answer.}

Stage~2 is identical in structure to the code-mixed Stage~2,
using three source-language demonstrations and the romanized
question as the test item.

\paragraph{One-Step Prompting Strategy}
\label{sec:one_step}

The one-step strategy performs reformulation and answering in a
\emph{single} inference pass. We evaluate four variants under
this framework, organized by reformulation type.

\subsubsection{English Setting}
\label{sec:one_step_en}

\paragraph{TinT-EN.}
The model is instructed to internally translate the source-language
question into English before answering. The translation never
appears in the output; only the answer is returned.

\begin{cmechpromptbox}{TinT-EN}
\begin{lstlisting}
You are answering factual questions written in <SOURCE_LANG>
(<SOURCE_SCRIPT> script).

You MUST internally translate the question into <TARGET_LANG>
to fully understand its meaning. This translation is ONLY
for internal reasoning --- do NOT output it.

Select the correct answer from the candidates.
Answer MUST be from the candidate list.
Answer MUST be in <SOURCE_SCRIPT> script only.

Output ONLY: {"answer": "<candidate>"}

Q: <QUESTION_1>
Candidates: <CANDIDATES_1>
Output: {"answer": "<ANSWER_1>"}

Q: <QUESTION_2>
Candidates: <CANDIDATES_2>
Output: {"answer": "<ANSWER_2>"}

Q: <QUESTION_3>
Candidates: <CANDIDATES_3>
Output: {"answer": "<ANSWER_3>"}

Q: <QUESTION>
Candidates: <CANDIDATES>
Output:
\end{lstlisting}
\end{cmechpromptbox}

\paragraph{1Step-EN+Ans.}
The model outputs both an English translation and the <source-script>
answer in a single JSON object. The translation field is logged for
analysis; only the answer field is evaluated.

\begin{cmechpromptbox}{1Step-EN+Ans}
\begin{lstlisting}
You are answering factual questions written in <SOURCE_LANG>
(<SOURCE_SCRIPT> script).

Translate the question into <TARGET_LANG>, then select the
correct answer from the candidates.

Output ONLY:
{"translation": "<English sentence>", "answer": "<candidate>"}

Rules:
- Translation MUST be a fluent <TARGET_LANG> sentence
- Translation MUST NOT contain <SOURCE_SCRIPT> text
- Answer MUST be chosen from the candidate list
- Answer MUST be in <SOURCE_SCRIPT> script only

Q: <QUESTION_1>
Candidates: <CANDIDATES_1>
Output: {"answer": "<ANSWER_1>"}

Q: <QUESTION_2>
Candidates: <CANDIDATES_2>
Output: {"answer": "<ANSWER_2>"}

Q: <QUESTION_3>
Candidates: <CANDIDATES_3>
Output: {"answer": "<ANSWER_3>"}

Q: <QUESTION>
Candidates: <CANDIDATES>
Output:
\end{lstlisting}
\end{cmechpromptbox}

\subsubsection{Code-Mixed Setting}
\label{sec:one_step_cm}

\paragraph{TinT-CM}
The model is instructed to internally perform a code-mixed
transformation before answering. This intermediate step is
never output; only the answer appears in the response.

\begin{cmechpromptbox}{TinT-CM}
\begin{lstlisting}
You are answering factual questions written in SOURCE_LANG
(SOURCE_SCRIPT script).

Before answering, internally convert the question to
TARGET_LANG by keeping SOURCE_LANG grammar and replacing
only content words with English (Roman script). Do NOT
output this step.

Example of the internal transformation:
  Input:    Bharat ki rajdhani kya hai?
  Internal: Bharat ki capital kya hai?

Select the correct answer from the candidates.
Answer MUST be from the candidate list.
Answer MUST be in SOURCE_SCRIPT script only.

Output ONLY: {"answer": "<candidate>"}

Q: <QUESTION_1>
Candidates: <CANDIDATES_1>
Output: {"answer": "<ANSWER_1>"}

Q: <QUESTION_2>
Candidates: <CANDIDATES_2>
Output: {"answer": "<ANSWER_2>"}

Q: <QUESTION_3>
Candidates: <CANDIDATES_3>
Output: {"answer": "<ANSWER_3>"}

Q: <QUESTION>
Candidates: <CANDIDATES>
Output:
\end{lstlisting}
\end{cmechpromptbox}

\paragraph{1Step-CM+Ans.}
The model outputs both a code-mixed transformation and the
source-script answer in a single JSON object. The code-mixed
field is logged for analysis; evaluation uses only the answer field.

\vspace{0.5em}

\begin{cmechpromptbox}{1Step-CM+Ans}
\begin{lstlisting}
You are answering factual questions written in <SOURCE_LANG>
(<SOURCE_SCRIPT> script).

Convert the question to code-mixed <TARGET_LANG> (<SOURCE_LANG>
grammar + English content words in Roman script), then
select the correct answer from the candidates.

Example:
  Input:      Bharat ki rajdhani kya hai?
  Code-mixed: Bharat ki capital kya hai?

Output ONLY:
{"translation": "<code-mixed question>", "answer": "<candidate>"}

Rules:
- Translation MUST preserve <SOURCE_LANG> word order
- Translation MUST NOT be fully fluent English
- Answer MUST be chosen from the candidate list
- Answer MUST be in <SOURCE_SCRIPT> script only

Q: <QUESTION_1>
Candidates: <CANDIDATES_1>
Output: {"answer": "<ANSWER_1>"}

Q: <QUESTION_2>
Candidates: <CANDIDATES_2>
Output: {"answer": "<ANSWER_2>"}

Q: <QUESTION_3>
Candidates: <CANDIDATES_3>
Output: {"answer": "<ANSWER_3>"}

Q: <QUESTION>
Candidates: <CANDIDATES>
Output:
\end{lstlisting}
\end{cmechpromptbox}

\begin{cmechpromptbox}{CoT Prompts}
\begin{lstlisting}
Answer the following question. 
Output JSON with 'reasoning' (1-2 sentences) and 'answer'.

Q: [template with <subject_1> filled]
Output: {"reasoning": "...", "answer": "<answer_1>"}

Q: [template with <subject_2> filled]
Output: {"reasoning": "...", "answer": "<answer_2>"}

Q: [template with <subject_3> filled]
Output: {"reasoning": "...", "answer": "<answer_3>"}

Q: [template with <subject_4> filled]
Output:
\end{lstlisting}
\hfill{\small [Baseline + CoT]}

\vspace{0.5em}

\begin{lstlisting}
Answer the following question. 
Output JSON with 'reasoning' (1-2 sentences) and 'answer'.

Q: [template with <subject_1> filled]
Output: {"reasoning": "...", "answer": "<answer_1>"}

Q: [template with <subject_2> filled]
Output: {"reasoning": "...", "answer": "<answer_2>"}

Q: [template with <subject_3> filled]
Output: {"reasoning": "...", "answer": "<answer_3>"}

Q: [template with <subject_4> filled]
Candidates: <candidate_1>, <candidate_2>, ..., <candidate_n>
Output:
\end{lstlisting}
\hfill{\small [Baseline + Candidates + CoT]}
\end{cmechpromptbox}

\end{document}